\documentclass{bmvc2k}

\usepackage{graphicx}
\usepackage{booktabs}
\usepackage{xcolor}
\usepackage{amsmath}
\usepackage{amssymb}
\usepackage{array}
\usepackage{multirow}
\usepackage{color}
\usepackage{colortbl}
\usepackage{framed}
\usepackage{bm}
\usepackage{bbm}
\usepackage{xspace}
\usepackage{enumitem}
\usepackage{xparse}
\usepackage{algorithm}
\usepackage{listings}
\usepackage{url}
\usepackage{wrapfig}

\title{Revisiting Image Captioning Training Paradigm via Direct CLIP-based Optimization}

\definecolor{Gray}{gray}{0.6}
\definecolor{LightCyan}{rgb}{0.88,0.95,1}
\definecolor{blond}{rgb}{0.98, 0.94, 0.75}

\def \ie {\emph{i.e.}}
\def \eg {\emph{e.g.}}
\def \etal {\emph{et al.}}

\def \argmax {\text{argmax}}

\newcommand{\tit}[1]{\smallbreak\noindent\textbf{#1.}}
\newcommand{\tinytit}[1]{\noindent\textbf{#1.}}

\newcommand{\ours}{DiCO\xspace}

\addauthor{Nicholas Moratelli}{nicholas.moratelli@unimore.it}{$\ast$1}
\addauthor{Davide Caffagni}{davide.caffagni@unimore.it}{$\ast$1}
\addauthor{Marcella Cornia}{marcella.cornia@unimore.it}{1}
\addauthor{Lorenzo Baraldi}{lorenzo.baraldi@unimore.it}{1}
\addauthor{Rita Cucchiara}{rita.cucchiara@unimore.it}{1,2}

\addinstitution{
 University of Modena and Reggio Emilia\\
 Modena, Italy
}
\addinstitution{
 IIT-CNR\\
 Pisa, Italy
}

\runninghead{N. Moratelli et al.}{Revisiting Image Captioning Training Paradigm}

\begin{document}

\maketitle

\begin{abstract}
The conventional training approach for image captioning involves pre-training a network using teacher forcing and subsequent fine-tuning with Self-Critical Sequence Training to maximize hand-crafted captioning metrics. However, when attempting to optimize modern and higher-quality metrics like CLIP-Score and PAC-Score, this training method often encounters instability and fails to acquire the genuine descriptive capabilities needed to produce fluent and informative captions. In this paper, we propose a new training paradigm termed \textit{Direct CLIP-Based Optimization} (\ours). Our approach jointly learns and optimizes a reward model that is distilled from a learnable captioning evaluator with high human correlation. This is done by solving a weighted classification problem directly inside the captioner. At the same time, \ours prevents divergence from the original model, ensuring that fluency is maintained. \ours not only exhibits improved stability and enhanced quality in the generated captions but also aligns more closely with human preferences compared to existing methods, especially in modern metrics. Additionally, it maintains competitive performance in traditional metrics. Our source code and trained models are publicly available at \url{https://github.com/aimagelab/DiCO}.
\end{abstract}

\section{Introduction}
\label{sec:intro}

The task of image captioning~\cite{stefanini2021show,vinyals2015show,xu2015show,karpathy2015deep} requires an algorithm to describe a visual input in natural language. As a captioner should ideally match the level of detail and precision desired by the user, over time there has been an increasing interest in developing training strategies for aligning the behavior of a captioner to mimic a desired style and quality level.

Traditionally, the quality of captions has been measured with textual similarity metrics, so captioners have been trained to maximize a non-differentiable metric like CIDEr~\cite{vedantam2015cider} during a fine-tuning stage based on reinforcement learning, \ie~Self-Critical Sequence Training (SCST)~\cite{rennie2017self,ranzato2015sequence,liu2017improved}. As this strategy requires the availability of multiple reference captions and tends to produce less distinctive descriptions that ignore the fine detailed aspects of an image, recently there have been preliminary attempts to optimize higher-quality image captioning metrics based on embedding spaces that do not require human references~\cite{cho2022fine,kornblith2023guiding,zhang2022distinctive}, like CLIP-Score~\cite{hessel2021clipscore} and PAC-Score~\cite{sarto2023positive}.
\begin{figure}[t]
    \centering
    \footnotesize
    \resizebox{\linewidth}{!}{
    \setlength{\tabcolsep}{.18em}
    \begin{tabular}{cc}
    \includegraphics[height=0.44\linewidth]{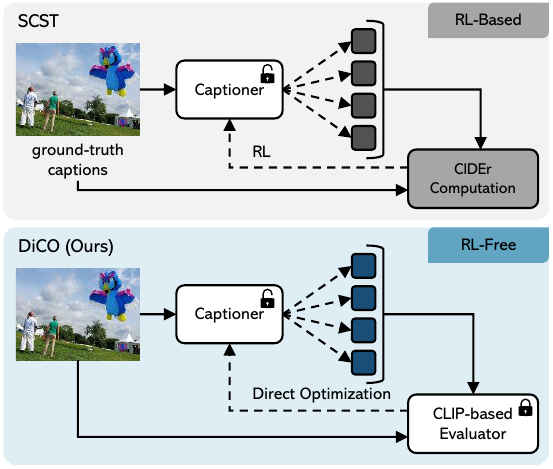} & 
    \includegraphics[height=0.44\linewidth]{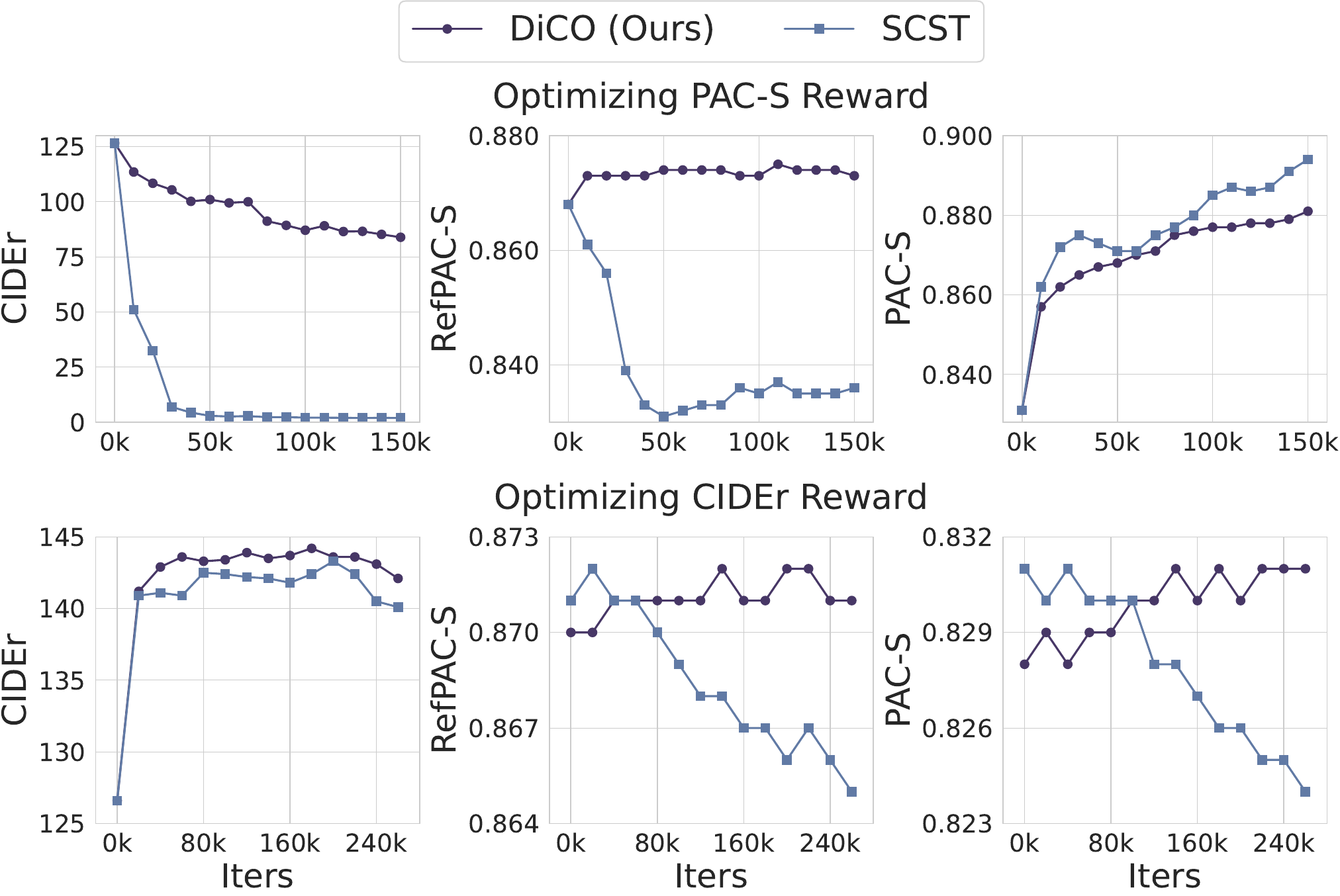} \\
    \end{tabular}
    }
    \vspace{-.3cm}
    \caption{Comparison between SCST~\cite{rennie2017self} and our \textit{Direct CLIP-Based Optimization} (\ours). \ours distills a reward model from a learnable CLIP-based captioning evaluator, without requiring reinforcement learning and preventing reward hacking and divergence.}
    \label{fig:first_page}
    \vspace{-.3cm}
\end{figure}
Besides, these metrics also consider the actual multi-modal alignment between the generated text and the visual content of the input image rather than just comparing texts. Most importantly, they also showcase a superior alignment with human judgment, making them ideal candidates for tuning the behavior of captioners towards a higher quality of generation.

Unfortunately, optimizing modern metrics with pre-existing strategies like SCST results in instability and model collapse~\cite{cho2022fine}. We showcase this in Fig.~\ref{fig:first_page}, where we employ SCST for optimizing either PAC-S or CIDEr (light blue lines). When we try to optimize PAC-S, the fine-tuned captioner hacks the metric and deviates from a fluent and high-quality generation, resulting in a rapid decrease according to all other metrics and leading to repetitions and grammatical errors. To solve these issues, we propose \ours, a novel training methodology that can align a captioner towards better quality captions by distilling from an external contrastive-based evaluator like CLIP-S or PAC-S, without incurring model collapse and without employing a reinforcement learning objective. Our approach achieves this goal by learning a reward model directly into the captioner and mimicking pairwise quality relations expressed by the external evaluator. This ensures a high degree of alignment with human preferences while avoiding reward hacking. This is visually represented in Fig.~\ref{fig:first_page} (dark blue lines): \ours can optimize both a modern metric like PAC-S and a traditional one like CIDEr by maintaining good scores across all metrics.

We assess the quality of the proposed training methodology by conducting extensive experiments on the COCO dataset~\cite{lin2014microsoft}. Furthermore, in the supplementary materials, we prove the generalization capabilities of \ours over other six image captioning benchmarks. Our experimental results demonstrate that \ours features state-of-the-art quality in the generated captions and improved training stability. This also results in a better performance in terms of modern captioning metrics, while also balancing with competitive performances on traditional handcrafted metrics. On the other hand, when adopted to maximize standard captioning metrics like CIDEr~\cite{vedantam2015cider}, \ours achieves state-of-the-art results also in this setting. Going beyond automatic image captioning metrics, we confirm the effectiveness of our approach by also employing human-based evaluation.

To sum up, our proposal markedly differs from all fine-tuning strategies in the current image captioning literature. Presently, this field remains closely tied to traditional techniques, employing the classic SCST algorithm with rewards based on ground-truth captions, while overlooking a concerted emphasis on semantic and syntactic richness, as well as alignment with human cognition. Extensive experiments on standard image captioning datasets demonstrate the effectiveness of the proposal.
\section{Related Work}
\label{sec:related}

\tinytit{Standard image captioning}
Early attempts in the field of image captioning were based on an encoder-decoder architecture, wherein the visual input content is encoded through a CNN, while the textual output is aptly generated by an RNN conditioned on the visual encoding~\cite{vinyals2015show,rennie2017self,karpathy2015deep,2018IPAS_QRNN}. Subsequently, this approach witnessed refinement through the integration of different attention-based strategies~\cite{xu2015show}, eventually applied to image regions~\cite{anderson2018bottom} and enhanced with spatial and semantic graphs~\cite{yao2018exploring,yang2019auto}. More recently, an alternative trend encompasses Transformer-based architectures, where numerous works have been developed exploring varied directions~\cite{huang2019attention,cornia2020meshed,li2022comprehending,barraco2023little}. While the aforementioned approaches exploited the same fine-tuning strategy usually composed of a pre-training with cross-entropy loss followed by reinforcement learning, we explore a different perspective. Along this line, Cho~\etal~\cite{cho2022fine} stands out as the method that is closely related to our proposal, as it defines a CLIP-based fine-tuning scheme that, however, relies on reinforcement learning. Concurrently, large-scale vision-and-language pre-training has been used to perform several tasks requiring multimodal capabilities, such as image captioning. These models~\cite{li2020oscar, zhang2021vinvl,hu2022scaling,wang2022ofa,wang2022git} are pre-trained on millions or even billions of image-text pairs, usually collected from the web, and fine-tuned for a target task. 

\tit{LLM-based image captioning} 
To leverage the power of LLMs demonstrated in different contexts, many attempts have emerged to bestow vision capabilities to a pre-trained LLM~\cite{zhu2023chatgpt,rotstein2023fusecap,koh2023grounding,koh2023generating,gao2023llama}, resulting in impressive performance over various vision-and-language tasks like image captioning. In this context, ZeroCap~\cite{tewel2022zerocap} runs a few optimization steps for each new token, to align the text produced by GPT-2~\cite{radford2019language} to the input image, using CLIP~\cite{radford2021learning} as guidance. Other works~\cite{ramos2023smallcap,mokady2021clipcap}, instead, start from a pre-trained LLM and only learn cross-attention layers to mimic the interaction between textual and visual domains. Research efforts have also been dedicated to developing large-scale multimodal models~\cite{caffagni2024r}, usually based on LLMs and trained on huge amounts of multimodal data~\cite{li2022blip,li2023blip,chen2023minigpt,caffagni2024wiki}. In this context, image captioning is employed as a pre-training task to help vision-and-language alignment, and eventually in the instruction-tuning stage~\cite{liu2023visual,liu2024improved}. Thanks to the underlying LLM, all these solutions usually lead to image captioners with greater descriptive capabilities. In this work, we show how to increase the quality and descriptiveness of generated captions without relying on any pre-trained LLM.

\tit{Training strategies for LLMs}
Aligning models with human judgment constitutes a well-known issue in both NLP and captioning literature. In this context, several strategies for fine-tuning LLMs have been explored. For example, a common research direction is to guide the model through a combination of input-output pairs and explicit instructions~\cite{wang2022self,wang2022super,iyer2022opt,vicuna2023}. However, LLMs often exhibit a tendency to generate biased and potentially harmful text. To solve this issue, some works have attempted to align models with human judgment through reinforcement learning~\cite{ziegler2019fine,ouyang2022training,touvron2023llama,sun2023principle}, also designing methodologies for efficient fine-tuning to tackle the substantial memory requirements inherent in training LLMs~\cite{lester2021power,liu2023gpt}.
\section{Proposed Method}

\tinytit{Preliminaries}
Self-critical sequence training (SCST~\cite{rennie2017self}) is a traditional training paradigm for image captioning. It consists of a two-step training methodology which pre-trains a captioner using a time-wise cross-entropy loss with respect to ground-truth sequences, and fine-tunes the same network by maximizing the CIDEr score~\cite{vedantam2015cider} using an reinforcement learning approach. Recently, it has been applied also with learnable metrics~\cite{hessel2021clipscore,sarto2023positive,sarto2024bridge} such as CLIP-S~\cite{hessel2021clipscore}, which employs a CLIP~\cite{radford2021learning} embedding space trained to align the embeddings of 400M images and caption pairs. Consequently, a high similarity between a pair of visual-textual CLIP embeddings means that the image-caption pair is highly correlated as well. On the other hand, reinforcement learning from human feedback (RLHF~\cite{ouyang2022training}) has been shown to be effective in making LLMs behave more like humans. It starts with a self-supervised pre-trained LLM, then goes through a supervised training phase, and finally, a fine-tuning stage using reinforcement learning. This last step is focused on improving the quality of generated responses by maximizing the score given by a reward model trained to imitate human judgment when comparing two candidate answers. We refer to Appendix~\ref{sec:preliminaries} for more details about SCST, RLHF, and image captioning metrics based on contrastive embedding spaces,~\ie~CLIP-S~\cite{hessel2021clipscore} and PAC-S~\cite{sarto2023positive}.

\tit{Motivation}
While adopting significantly different technical choices, there are striking conceptual similarities between the modern RLHF paradigm employed in LLMs and the traditional SCST approach employed in image captioning. Both approaches, indeed, employ reinforcement learning to optimize a reward function, which nevertheless in SCST is a hand-crafted metric, while in RLHF is a learned function from human data. While using RLHF in captioning is impracticable due to the insufficient amount 
of human preference data to train the reward model (see the comparison with RLHF in Appendix~\ref{sec:rlhf_comparison}), contrastive-based learnable metrics offer a compelling alternative to it, as they show a significant alignment with human judgment~\cite{sarto2023positive}. Our proposal solves this issue by distilling a reward model from a pre-trained captioning evaluator, considering pairwise relationships from candidate captions. In addition, it also avoids model collapse which is frequent in SCST (cf. Fig.~\ref{fig:first_page}).

\begin{figure*}[t]
    \centering
    \includegraphics[width=\linewidth]{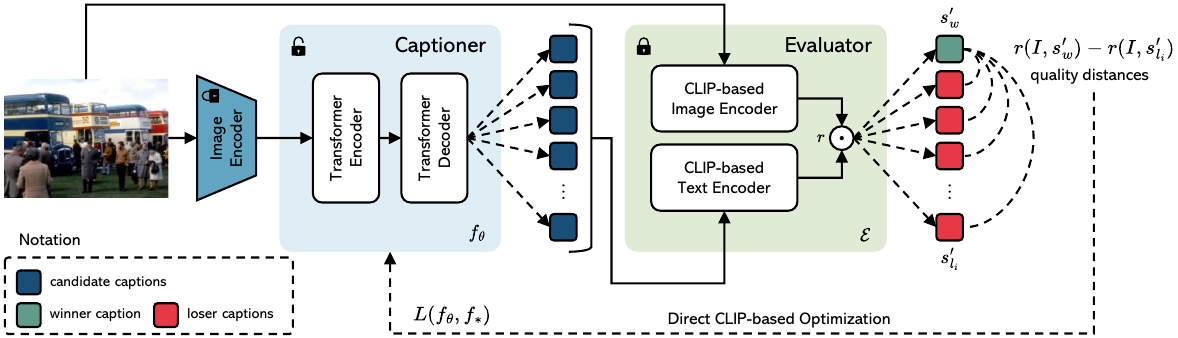}
    \vspace{-0.5cm}
    \caption{Overview of our approach. Given an image and candidate generations, the figure shows the process for captioner fine-tuning by distilling from a CLIP-based evaluator.}
    \label{fig:method}
    \vspace{-.3cm}
\end{figure*}

\tit{Deriving the fine-tuning objective}
Following recent works on LLM alignment~\cite{ouyang2022training}, we aim at fine-tuning a captioner $f_\theta$ with a  Proximal Policy Optimization (PPO) objective~\cite{schulman2017proximal}, where given an image $I$ and a caption $s'$ sampled from the model, the environment produces a reward $r(s', I)$ through a reward model. In addition, we add a per-token KL penalty with the output of the pre-trained model to mitigate overoptimization of the fine-tuned captioner to the reward model. Our objective is therefore defined as
\begin{equation}
    \max_{f_\theta} \mathbb{E}_{I \sim \mathcal{D}, s' \sim f_\theta(\cdot|I)} \left[ r(s', I) \right] - \beta \mathbb{D}_{\text{KL}} \left[ f_\theta(s' | I) || f_*(s'|I)
    \right],
    \label{eq:ppo}
\end{equation}
where $\beta$ controls the deviation from the pre-trained model, termed as $f_*$. As it can be seen, the second term has a crucial role, as it prevents the fine-tuned model $f_\theta$ from deviating from the distribution on which the reward model is accurate, and prevents the captioner from hacking it, \ie~collapsing to high-rewarded answers.

Under this objective, it can be shown~\cite{rafailov2024direct} that the optimal solution to the fine-tuning problem is given by a model $f_r$ defined as
\begin{equation}
    \label{eq:f_r}
    f_r(s'|I) = \frac{1}{Z(I)} f_* (s'|I) \exp \left( \frac{1}{\beta} r(s', I)\right),
\end{equation}
where $Z(I) = \sum_s f_*(s|I)\exp\left(\frac{1}{\beta} r(s,I)\right)$ is the partition function over possible captions. Although the partition function is difficult to estimate, we can still manipulate Eq.~\ref{eq:f_r} to express the reward function in terms of the optimal captioner, the pre-trained captioner, and the partition function, as follows:
\begin{equation}
    \label{eq:r}
    r(s',I) = \beta \log \frac{f_r(s'|I)}{f_*(s'|I)} + \beta \log Z(I).
\end{equation}

\tit{Defining a distilled reward model}
Since we do not have access to sufficiently large human preference data, defining the reward model in a purely data-driven way would be cumbersome. Instead, we learn our reward model by \textit{distilling} it from a contrastive-based captioning evaluator $\mathcal{E}$. We assume that, given an image and a candidate sentence $(I, s')$, the evaluator returns a matching score $\mathcal{E}(s', I)$ proportional to the similarity between $s'$ and $I$.

Given a dataset $\mathcal{D}$ comprising images, we let the captioner generate $k+1$ candidate captions (\eg~through beam search). Then, for each image, we select the caption with the highest score according to $\mathcal{E}$ and denote it as $s'_w$ (\ie~``winner''). The others, instead, are denoted as $\{s'_{l_i}\}_{i=1}^{k}$ (\ie~``losers''). Based on the evaluator, we define a reward model which distinguishes between the winner caption $s'_w$ and the loser captions $\{s'_{l_i}\}_i$. To make the reward model more robust and accurate, we also impose that it can predict the \textit{relative quality distances} between the winner and the loser captions.
Formally, we define our reward model through the following objective:
\begin{align}\label{eq:differences}
    \mathcal{L}_R(r) = - \mathbb{E} \left[ \log \sigma
    \left( \sum_{i=1}^k \gamma_i (r(I, s'_w) - r(I, s'_{l_i})) \right) \right],
\end{align}
where the expectation is taken over images in the dataset and winner and loser captions. Also, $\gamma_i$ weights the relative distance between the winner caption $s'_w$ and the $i$-th loser caption $s'_{l_i}$ according to the evaluator $\mathcal{E}$. Specifically, it is computed as a normalized probability distribution between score distances, as follows:
\begin{equation}\label{eq:softmax}
    \gamma_i = \text{softmax}_{s'_{l_1}, ..., s'_{l_k}} \left( \frac{\mathcal{E}(I, s'_w) - \mathcal{E}(I, s'_{l_i})}{\tau} \right),
\end{equation}
where $\tau$ is a temperature parameter.
Clearly, considering that $\gamma_i$ sum up to 1, the reward model objective can be rewritten as
\begin{align}
    \label{eq:reward_model}
    \mathcal{L}_R(r) = - \mathbb{E} \left[ \log \sigma
    \left( r(I, s'_w) - \sum_{i=1}^k \gamma_i r(I, s'_{l_i}) \right) \right].
\end{align}

\tit{Overall loss function}
Following~\cite{rafailov2024direct}, we learn the reward model directly into the captioner. Recalling that the Bradley-Terry model depends only on the difference in rewards between two
completions and that $\gamma_i$ are a valid probability distribution, we replace the definition of $r(s',I)$ as a function of the optimal fine-tuned and pre-trained captioner (Eq.~\ref{eq:r}) into the reward model objective (Eq.~\ref{eq:reward_model}), and obtain the final fine-tuning loss of \ours as
\begin{align}
    L(f_\theta, f_*) = & -\mathbb{E} 
        \left[ 
            \log \sigma  
                \left(\beta \log \frac{f_\theta(s'_w|I)}{f_*(s'_w | I)} 
                - \beta \sum_{i=1}^k \gamma_i \log \frac{f_\theta(s'_{l_i}|I)}{f_*(s'_{l_i} | I)}
                \right)\right],
\end{align}
where, noticeably, the unknown partition function $Z(I)$ has been cancelled out. Furthermore, the obtained fine-tuning loss, while being derived from the optimal solution to a PPO objective, can be directly optimized through gradient descent, without the need of employing reinforcement learning techniques.

\tit{Comparing \ours with SCST and RLHF} \ours fine-tunes a captioning model by aligning it to a contrastive-based evaluator while avoiding over-parametrization and model collapse. In comparison with SCST and RLHF, its unique feature is that of \textit{distilling a reward model from an external evaluator} by \textit{learning it directly inside of the captioner}. Further, this is done by avoiding the usage of reinforcement learning at fine-tuning time, which is common to both SCST and RLHF. Differently from RLHF, also, caption candidates are directly sampled from the model, so that a dataset of human-annotated preferences can be avoided. Finally, differently from SCST, \ours embeds a regularizer to prevent the fine-tuned model from deviating too much from the pre-trained captioner. 
\section{Experiments}
\label{sec:experiments}

\subsection{Experimental Setting}
\tinytit{Datasets} All experiments are performed on the COCO dataset~\cite{lin2014microsoft}, using the standard splits defined in~\cite{karpathy2015deep} with 5,000 images for both test and validation and the rest for training. We report our experimental results on the test set of COCO. Further, we refer the reader to Appendix~\ref{sec:rlhf_comparison} for results on six additional datasets, namely nocaps~\cite{agrawal2019nocaps}, VizWiz~\cite{gurari2020captioning}, TextCaps~\cite{sidorov2020textcaps}, Conceptual Captions 3M (CC3M)~\cite{sharma2018conceptual}, FineCapEval~\cite{cho2022fine}, and Flickr30k~\cite{young2014image}.

\tit{Evaluation metrics} In addition to the standard image captioning metrics like BLEU~\cite{papineni2002bleu}, METEOR~\cite{banerjee2005meteor}, and CIDEr~\cite{vedantam2015cider}, we employ two CLIP-based scores, namely CLIP-S~\cite{hessel2021clipscore} and PAC-S~\cite{sarto2023positive}, in both their reference-free and reference-based versions, using the ViT-B/32 backbone for both metrics (also see Appendix~\ref{sec:preliminaries}). Moreover, following recent works~\cite{kornblith2023guiding,chan2023ic}, we measure the quality of generated captions in distinguishing images in a dataset and compute the percentage of the times the image corresponding to each generated caption is retrieved among the first $K$ retrieved items. This is done by ranking the images in terms of CLIP similarity between visual and textual embeddings, using the CLIP ViT-B/32 model, and computing recall at $K$ with $K=1,5,10$. We also compute the mean reciprocal rank (MRR) for each generated caption: higher MRR scores indicate that captions are more discriminative and therefore usually more detailed.

\tit{Implementation and training details} Our baseline architecture is a standard Transformer model with 3 layers in both encoder and decoder, a hidden dimensionality equal to 512, and 8 attention heads. To extract visual features, we use either RN50, ViT-B/32, or ViT-L/14 pre-trained with a CLIP-based objective~\cite{radford2021learning}. Our code is based on the popular Hugging Face Transformers\footnote{\url{https://huggingface.co/docs/transformers}} library. All experiments are performed using the Adam optimizer, initially pre-training all the models with cross-entropy. During fine-tuning, we use a batch size of 16, a fixed learning rate equal to $1 \cdot 10^{-6}$, and a beam size of 5 (\ie~the number $k$ of looser captions is set to 4). For efficiency, we train with ZeRo memory offloading and mixed-precision~\cite{micikevicius2017mixed}. Unless otherwise specified, the $\beta$ parameter is set to 0.2 and the ViT-L/14 backbone is used to extract visual features. The temperature parameter $\tau$ defined in Eq.~\ref{eq:softmax} is set to $1/(3\cdot10^2)$. Early stopping is performed according to the reference-based version of the CLIP metrics used as reward. Ablation studies and analyses with different hyperparameters are reported in Appendix~\ref{sec:ablations}.

\begin{table*}[t]
  \setlength{\tabcolsep}{.18em}
  \resizebox{\linewidth}{!}{
  \begin{tabular}{lcc c ccccc c cccccc}
    \toprule
    & & & \multicolumn{6}{c}{\textbf{Reference-based Metrics}} & & \multicolumn{6}{c}{\textbf{Reference-free Metrics}} \\ 
    \cmidrule{5-9} \cmidrule{11-16}
    \textbf{Model} & & & & B-4 & M & C & RefCLIP-S & RefPAC-S & & CLIP-S & PAC-S & R@1 & R@5 & R@10 & MRR \\ 
    \midrule
    \textit{Standard Captioners} & & \textbf{Backbone} \\
    CLIP-VL~\cite{shen2022much} & & RN50$\times$4 & &  40.2 & 29.7 & 134.2 & 0.820 & 0.862 & & 0.770 & 0.826 & 24.0 & 48.9 & 61.5 & 34.8 \\ 
    COS-Net~\cite{li2022comprehending} & & RN101 & & 42.0 & 30.6 & 141.1 & 0.814 & 0.870 & & 0.758 & 0.832 & 25.8 & 52.3 & 64.9 & 37.1 \\
    PMA-Net~\cite{barraco2023little} & & ViT-L/14 & & 43.0 & 30.6 & 144.1 & 0.814 & 0.869 & & 0.755 & 0.821 & - & - & - & - \\ 
    \midrule
    \textit{LLM-based Captioners} & & \textbf{Backbone} \\
    ZeroCap~\cite{tewel2022zerocap} & & ViT-B/32 & & 2.3 & 10.1 & 15.1 & 0.771 & 0.800 & & 0.810 & 0.816 & - & - & - & - \\ 
    ClipCap~\cite{mokady2021clipcap} & & ViT-B/32 & & 32.3 & 28.1 & 108.5 & 0.809 & 0.862 & & 0.766 & 0.833 & 27.1 & 53.3 & 65.5 & 38.3 \\ 
    SmallCap~\cite{ramos2023smallcap} & & ViT-B/32 & & 37.0 & 27.9 & 119.7 & 0.804 & 0.863 & & 0.748 & 0.826 & 23.1 & 48.2 & 60.0 & 33.7 \\ 
    MiniGPT-v2~\cite{chen2023minigpt} & &  ViT-g/14 & & 18.8 & 24.6 & 80.4 & 0.795 & 0.848 & & 0.752 & 0.818 & 27.4 & 52.0 & 63.0 & 37.9 \\ 
    BLIP-2~\cite{li2023blip} & & ViT-g/14 & & \underline{43.7} & \underline{32.0} & \underline{145.8} & 0.823 & \underline{0.877} & & 0.767 & 0.837 & 31.4 & 57.5 & 69.1 & 42.7 \\ 
    \midrule
    \textit{CLIP-based Optimization} & \textbf{Reward} & \textbf{Backbone} \\
    Cho~\etal~(SCST)~\cite{cho2022fine} & CLIP-S & RN50 & & 6.3 & 19.7 & 11.2 & 0.786 & 0.823 & & \textbf{0.843} & 0.837 & 43.2 & 71.9 & 82.3 & 55.5 \\ 
    Cho~\etal~(SCST)~\cite{cho2022fine} & CLIP-S+Gr. & RN50 & & 16.9 & 24.9 & 71.0 & 0.792 & 0.849 & & 0.779 & 0.839 & 35.3 & 63.4 & 75.2 & 47.4 \\ 
    SCST & CLIP-S & RN50 & & 14.3 & 24.7 & 3.1 & 0.765 & 0.830 & & 0.804 & 0.837 & 36.9 &	64.9 & 75.9 & 48.7 \\
    SCST & PAC-S & RN50 & & 18.5 & 26.5 & 32.2 & 0.785 & 0.849 & & 0.799 & 0.860 & \textbf{44.3} & 73.2 & 83.4 & 56.5 \\
    \rowcolor{LightCyan}
    \textbf{\ours (Ours)} & CLIP-S & RN50 & & 20.7 & 25.7 & 78.9 & \textbf{0.811} & 0.852 & & 0.815 & 0.842 & 37.5 & 66.6 & 78.1 & 49.8 \\
    \rowcolor{LightCyan}
    \textbf{\ours (Ours)} & PAC-S & RN50 & & \textbf{22.7} & \textbf{27.0} & \textbf{79.8} & 0.801 & \textbf{0.865} & & 0.797 & \textbf{0.869} & \textbf{44.3} & \textbf{73.9} & \textbf{84.2} & \textbf{56.8} \\
    \midrule
    SCST & CLIP-S & ViT-B/32 & & 11.4 & 23.1 & 1.1 & 0.778	& 0.830 & & \textbf{0.851} & 0.846 & 43.4 &	70.8 & 81.1 & 55.1 \\ 
    SCST & PAC-S & ViT-B/32 & & 20.3 & 27.1 & 40.7 & 0.796 & 0.858 & & 0.810 & 0.870 & 50.0 & 77.6 & 87.0 & 61.8 \\ 
    \rowcolor{LightCyan}
    \textbf{\ours (Ours)} & CLIP-S & ViT-B/32 & & 22.6 & 27.0	& 81.7 & \textbf{0.817} & 0.861 & & 0.825 & 0.858 & 46.3 & 74.0 & 83.7 & 58.0 \\ 
    \rowcolor{LightCyan}
    \textbf{\ours (Ours)} & PAC-S & ViT-B/32 & & \textbf{23.7} & \textbf{27.3} & \textbf{84.8} & 0.810 & \textbf{0.872} & & 0.814 & \underline{\textbf{0.882}} & \underline{\textbf{52.9}} & \underline{\textbf{80.8}} & \underline{\textbf{89.5}} & \underline{\textbf{64.8}} \\
    \midrule
    SCST & CLIP-S & ViT-L/14 & & 10.2 & 23.0 & 1.1 & 0.793 & 0.827 & & \underline{\textbf{0.865}} & 0.834 & 43.3 & 70.7 & 80.5 & 55.0 \\ 
    SCST & PAC-S & ViT-L/14 & & 22.3 & \textbf{28.4} & 51.1 & 0.801 & 0.861 & & 0.805 & 0.862 & 46.7 & 74.7 & 84.8 & 58.8 \\
    \rowcolor{LightCyan}
    \textbf{\ours (Ours)} & CLIP-S & ViT-L/14 & & 21.4 & 27.1 & 82.6 & \underline{\textbf{0.824}} & 0.863 & & 0.837 & 0.856 & 46.5 & 74.7 & 84.7 & 58.4 \\
    \rowcolor{LightCyan}
    \textbf{\ours (Ours)} & PAC-S & ViT-L/14 & & \textbf{25.2} & \textbf{28.4} & \textbf{89.1} & 0.815 & \textbf{0.875} & & 0.812 & \textbf{0.877} & \textbf{50.9} & \textbf{78.7} & \textbf{87.6} & \textbf{62.9} \\
    \bottomrule
  \end{tabular}
}
\vspace{-0.15cm}
  \centering
    \caption{Comparison with state-of-the-art models on the COCO test set. Bold font indicates the best results among captioners optimized via CLIP-based rewards with comparable backbones, while underlined indicates the overall best results.} 
    \label{tab:comparison}
\vspace{-0.3cm}
\end{table*}

\subsection{Comparison with the State of the Art}
\tinytit{Results on COCO test set} We compare our model trained with the proposed \ours strategy with other state-of-the-art solutions. We restrain the comparison by only considering captioning models that use CLIP-based visual features to encode images, which have proven to be the most widely employed choice in recent works. In particular, we include some recent standard image captioning models exclusively trained on the COCO dataset with a standard XE+SCST training paradigm like CLIP-VL~\cite{shen2022much}, COS-Net~\cite{li2022comprehending}, and PMA-Net~\cite{barraco2023little}. Moreover, we compare with LLM-based captioning models focused on zero-shot generation capabilities such as ZeroCap~\cite{tewel2022zerocap}, lightweight architectures like ClipCap~\cite{mokady2021clipcap} and SmallCap~\cite{ramos2023smallcap}, or large-scale training paradigms such as the recently proposed MiniGPT-v2~\cite{chen2023minigpt} and BLIP-2~\cite{li2023blip} models. To directly compare our solution with other CLIP-based optimization strategies, we also report the results of our base model trained with SCST using CLIP-S or PAC-S as reward and those of the model proposed in~\cite{cho2022fine} in which a standard Transformer is optimized via SCST with a CLIP-based reward, eventually regularizing the training with an additional score that considers the grammatical correctness of generated sentences.

Results are shown in Table~\ref{tab:comparison}, including our model trained with \ours using RN50, ViT-B/32, and ViT-L/14 as visual backbones. Notably, all versions of our model achieve better results than other methods optimized with CLIP-based rewards on almost all evaluation metrics. For example, when comparing our solution optimized via PAC-S reward with SCST and the model proposed in~\cite{cho2022fine}, we can notice how not only \ours improves the performance in terms of standard metrics (\eg~79.8 CIDEr points using RN50 features vs. 32.2 and 71.0 respectively obtained by SCST and~\cite{cho2022fine}), but also obtains increased retrieval-based scores indicating that captions generated by our model are more discriminative and detailed than those generated by competitors. 
\begin{figure}[t]
    \begin{minipage}{0.165\linewidth}
        \includegraphics[width=0.97\linewidth]{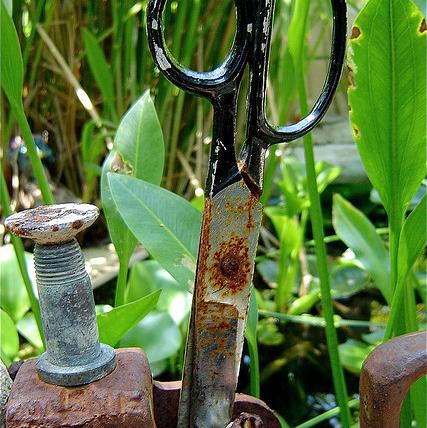}
        \end{minipage}
        \begin{minipage}{0.32\linewidth}
        \scriptsize{
        \textbf{SCST:} A rusted rusty rusty rusty rusty rusty rusty scissors in a garden with plants in the background.\\
        \textbf{DiCO (Ours):} A rusted scissors sticking out of a metal fence with plants in the background.
        }
    \end{minipage}
    \hspace{0.02cm}
    \begin{minipage}{0.165\linewidth}
        \includegraphics[width=0.97\linewidth]{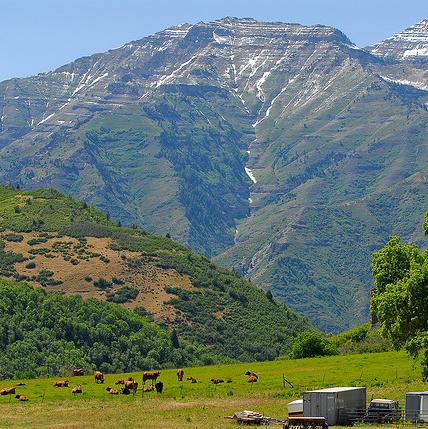}
        \end{minipage}
        \begin{minipage}{0.32\linewidth}
        \scriptsize{
        \textbf{SCST:} Several cows grazing in a field on a mountain range with a mountain in distance under mountain range in background.\\
        \textbf{DiCO (Ours):} A herd of cattle grazing on a lush green hill next to a large mountain range.
        }
    \end{minipage}
    
    \vspace{0.1cm}

    \begin{minipage}{0.165\linewidth}
        \includegraphics[width=0.97\linewidth]{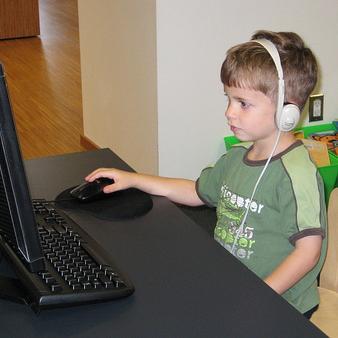}
        \end{minipage}
        \begin{minipage}{0.32\linewidth}
        \scriptsize{
        \textbf{SCST:} A little boy with headphones sitting in front of a computer with headphones on a desk in corner with headphones on background.\\
        \textbf{DiCO (Ours):} A little boy with headphones sitting at a desk using a computer.
        }
    \end{minipage}
    \hspace{0.02cm}
    \begin{minipage}{0.165\linewidth}
        \includegraphics[width=0.97\linewidth]{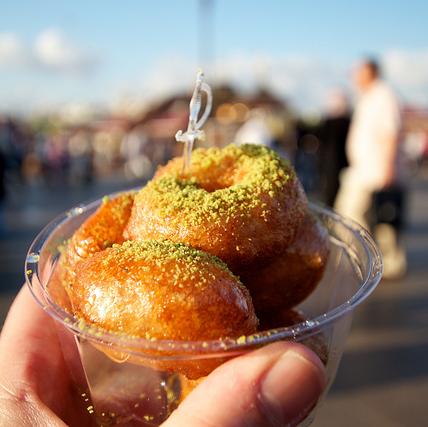}
        \end{minipage}
        \begin{minipage}{0.32\linewidth}
        \scriptsize{
        \textbf{SCST:} A person holding up a clear plastic plastic container filled with sugar covered donuts with people in background in background on background.\\
        \textbf{DiCO (Ours):} A hand holding a plastic container filled with sugar covered donuts.
        }
    \end{minipage}
\vspace{-.1cm}
\caption{Qualitative results on COCO sample images, using PAC-S as reward.}
\label{fig:qualitatives}
\vspace{-.3cm}
\end{figure}
Additionally, \ours leads to the overall best results on reference-free metrics also surpassing huge models trained on millions or even billions of data like MiniGPT-v2 and BLIP-2, further confirming the effectiveness of our training strategy.
To validate the quality of generated captions, we report in Fig.~\ref{fig:qualitatives} some qualitative results on sample images from the COCO dataset. \ours can generate more descriptive and detailed captions while reducing repetitions and grammatical errors typically generated using SCST. We refer to Appendix~\ref{sec:qualitatives} for additional qualitative results.

\tit{Human-based and LLM-based evaluations}
As a complement of standard metrics, we also perform a user study and an evaluation based on a widely used LLM (\ie~GPT-3.5). To perform the user study, we present the users with an image and a pair of captions, one generated by our model and the other generated by a competitor, and ask them to select the preferred caption judging in terms of (1) \textit{helpfulness} (\ie~which caption is most helpful to someone who can not see the image), and (2) \textit{correctness} (\ie~which caption is more correct both in terms of grammar and consistency with the image). Users could also state that captions are equivalent on one or both evaluation axes. In this case, 0.5 points are given to both captions. To perform LLM-based evaluation, instead, we leverage the Turbo version of GPT-3.5 and directly ask it to evaluate a pair of captions taking into account the 
\begin{wraptable}{r}{0.65\textwidth}
  \vspace{-0.1cm}
  \centering
  \setlength{\tabcolsep}{.18em}
  \resizebox{0.99\linewidth}{!}{
  \begin{tabular}{lc cc c c}
    \toprule
    & & \multicolumn{2}{c}{\textbf{Humans}} \\
    \cmidrule{3-4} 
    & & Helpfulness & Correctness & & \textbf{GPT-3.5} \\
    \midrule
    ZeroCap~\cite{tewel2022zerocap} & & 20.3 & 27.8 & & 20.8 \\
    SmallCap~\cite{ramos2023smallcap} & & 27.8 & 36.1 & & 50.0 \\
    MiniGPT-v2~\cite{chen2023minigpt} & & 33.3 & 42.9 & & 44.8 \\
    BLIP-2~\cite{li2023blip} & & 49.1 & 48.6 & & \textcolor{Gray}{51.2} \\
    \midrule
    Cho~\etal~(CLIP-S Reward)~\cite{cho2022fine} & & 11.2 & 17.9 & & 21.5 \\
    Cho~\etal~(CLIP-S+Gr. Reward)~\cite{cho2022fine} & & 41.3 & 36.7 & & 43.0 \\
    SCST (PAC-S Reward) & & 44.6 & 40.6 & & 48.5 \\
    \bottomrule
  \end{tabular}
}
\vspace{0.2cm}
\caption{Percentage of times a caption from a competitor is preferred against that generated by our proposal, using either human-based evaluations or GPT-3.5. Our solution is preferred more than 50\% of the time in almost all cases.}
\label{tab:user_study}
\vspace{-0.15cm}
\end{wraptable}
corresponding reference sentences. In particular, we ask the LLM to return a score between 0 and 100 for each caption between the two in the prompt, where one is generated by our model and the other by a competitor, and use this score to compute the number of times GPT-3.5 prefers our solution against a competitor and vice versa. If the score is the same for both captions, we give 0.5 points to both of them. To force the model to produce a more accurate evaluation, we also ask it to produce a reason for each score, which has been shown to lead to ratings that correlate well with human judgment~\cite{chan2023clair}. 

Table~\ref{tab:user_study} shows one-to-one
comparisons between our model and one of the considered competitors in terms of both human-based and LLM-based evaluations. Results are reported on a subset of 1,000 images randomly taken from the COCO test set. For each comparison, we report the percentage of times a caption generated by one of the competitors is preferred against the one generated by our solution with PAC-S reward. As it can be seen, \ours is almost always preferred more than 50\% of the time, having a comparable number of preferences only when compared with BLIP-2. When instead considering other CLIP-based optimized models, captions generated by our solution are selected in a considerable number of cases from both human evaluators and GPT-3.5 (\eg~more than 55-60\% compared to~\cite{cho2022fine} with CLIP-S+Grammar reward). Additional details are reported in Appendix~\ref{sec:details}.

\begin{table}[b]
\vspace{-0.2cm}
  \centering
  \setlength{\tabcolsep}{.35em}
  \resizebox{0.88\linewidth}{!}{
  \begin{tabular}{lcc c ccc c cccccc}
  \toprule
  & & & & \multicolumn{3}{c}{\textbf{Semantic}} & & \multicolumn{6}{c}{\textbf{Grammar}} \\
  \cmidrule{5-7} \cmidrule{9-14}
  \textbf{Model} & \textbf{Reward} & \textbf{Backbone} & & C & CLIP-S & PAC-S & & $\textit{n}_{1}$ & $\textit{n}_{2}$ & $\textit{n}_{3}$ & $\textit{n}_{4}$ & RE & \%Correct \\
  \midrule
  SCST & CLIP-S & RN50 & & 3.1 & 0.804 & 0.837 & & 11.762 & 5.168 & 2.809 & 1.518 & 6.0 & 24.7 \\
  SCST & PAC-S & RN50 & & 32.2 & 0.799 & 0.860 & & 5.453 & 1.588 & 0.645 & 0.288 & 1.6 & 71.6 \\
  \rowcolor{LightCyan}
  \textbf{\ours} & CLIP-S & RN50 & & 78.9 & \textbf{0.815} & 0.842 & & \textbf{1.583} & \textbf{0.143} & \textbf{0.039} & \textbf{0.015} & \textbf{0.1} & \textbf{96.1} \\
  \rowcolor{LightCyan}
  \textbf{\ours} & PAC-S & RN50 & & \textbf{79.8} & 0.797 & \textbf{0.869} & & 2.051 & 0.219 & 0.055 & 0.017 & \textbf{0.1} & 94.4 \\
  \midrule
  SCST & CLIP-S & ViT-B/32 & & 1.1 & \textbf{0.851} & 0.846 & & 11.166 & 3.566 & 1.232 & 0.395 & 1.5 & 3.9 \\
  SCST & PAC-S & ViT-B/32 & & 40.7 & 0.810 & 0.870 & & 5.078 & 1.443 & 0.584 & 0.260 & 1.6 & 73.3 \\
  \rowcolor{LightCyan}
  \textbf{\ours} & CLIP-S & ViT-B/32 & & 81.7 & 0.825 & 0.858 & & \textbf{1.938} & 0.230 & 0.071 & 0.026 & 0.2 & 94.8 \\
  \rowcolor{LightCyan}
  \textbf{\ours} & PAC-S & ViT-B/32 & & \textbf{84.8} & 0.814 & \textbf{0.882} & & 1.939 & \textbf{0.190} & \textbf{0.048} & \textbf{0.014} & \textbf{0.1} & \textbf{96.4} \\
  \midrule
  SCST & CLIP-S & ViT-L/14 & & 1.1 & \textbf{0.865} & 0.834 & & 8.788 & 2.114 & 0.716 & 0.248 & 1.0 & 2.6 \\
  SCST & PAC-S & ViT-L/14 & & 51.1 & 0.805 & 0.862 & & 8.788 & 4.611 & 1.200 & 0.479 & 1.3 & 72.6 \\
  \rowcolor{LightCyan}
  \textbf{\ours} & CLIP-S & ViT-L/14 & & 82.6 & 0.837 & 0.856 & & \textbf{1.710} & \textbf{0.142} & \textbf{0.039} & \textbf{0.014} & \textbf{0.1} & \textbf{95.4} \\
  \rowcolor{LightCyan}
  \textbf{\ours} & PAC-S & ViT-L/14 & & \textbf{89.1} & 0.812 & \textbf{0.877} & & 2.107 & 0.218 & 0.056 & 0.017 & \textbf{0.1} & 94.3 \\
  \bottomrule
  \end{tabular}
}
\vspace{0.25cm}
  \caption{Comparison on semantic and grammar metrics. $\textit{n}_{i}$ means \textit{i}-gram repetitions. Results are reported on the COCO test set.}
  \label{tab:grammar}
\end{table}

\tit{Additional results on grammar metrics}
Besides the semantic coherence between images and their descriptions, we compare our method against SCST from the point of view of the fluency and grammatical correctness of the generated captions. To this end, in Table~\ref{tab:grammar}
we report the average number of $n$-gram repetitions per caption (\ie~$n_i$ with $i=1,2,3,4$), computed using the \texttt{nltk} language toolkit\footnote{\url{https://www.nltk.org/}}. We also include the Repetition Evaluation (RE) proposed in~\citep{xiong2018move}, which measures the redundancy of $n$-grams inside a caption (where $n=4$ as in the original paper). Additionally, we employ the text encoder from~\cite{cho2022fine} and present the percentage of captions classified as grammatically correct (\ie~\%Correct). Experiments across different backbones confirm that SCST reaches high scores on the optimized metrics, but collapses to predictions that exhibit many repetitions, undermining the fluency of the generated text. \ours does not suffer from the same problem, keeping low values for repetitions while showcasing state-of-the-art performance over the reward metrics.

\tit{CIDEr-based optimization}
Finally, we assess whether our training paradigm can also be applied using the CIDEr score as reward, as usually done in standard image captioning approaches. Results are reported in Table~\ref{tab:cider}, showing the performance of a standard Transformer model fine-tuned with the classical SCST procedure and that of other captioners. For completeness, we also include the results in terms of ROUGE~\cite{lin2004rouge} and SPICE~\cite{spice2016} which are typically used in standard image captioning evaluation. In this case, we apply \ours with different $\beta$ values on the same baseline architecture used in previous experiments (\ie~a vanilla Transformer with 3 layers in both encoder and decoder). Interestingly, our solution achieves better results than SCST also in this setting, with 144.2 CIDEr points vs. 143.3 obtained by SCST. As an additional result, \ours reaches better or comparable performance to that obtained by recent captioning models based on more complex architectures and optimized via SCST, thus proving to be a valid alternative also in a standard CIDEr-based setting.

\begin{table}[t]
  \centering
  \setlength{\tabcolsep}{.25em}
  \resizebox{0.95\linewidth}{!}{
  \begin{tabular}{lc ccccccc c cc}
    \toprule
    & & \multicolumn{7}{c}{\textbf{Reference-based}} & & \multicolumn{2}{c}{\textbf{Reference-free}} \\
    \cmidrule{3-9} \cmidrule{11-12}
    \textbf{Model} & & B-4 & M & R & C & S & RefCLIP-S & RefPAC-S & & CLIP-S & PAC-S \\
    \midrule
    Up-Down~\cite{anderson2018bottom} & & 36.3 & 27.7 & 56.9 & 120.1 & 21.4 & 0.787 & 0.848 & & 0.723 & 0.803 \\
    SGAE~\cite{yang2019auto} & & 39.0 & 28.4 & 58.9 & 129.1 & 22.2 & 0.796 & 0.855 & & 0.734 & 0.812 \\
    AoANet~\cite{huang2019attention} & & 38.9 & 29.2 & 58.8 & 129.8 & 22.4 & 0.797 & 0.857 & & 0.737 & 0.815 \\
    $\mathcal{M}^2$ Transformer~\cite{cornia2020meshed} & & 39.1 & 29.8 & 58.3 & 131.3 & 22.6 & 0.793 & 0.852 & & 0.734 & 0.813 \\
    COS-Net~\cite{li2022comprehending} & & 42.0 & 30.6 & 60.6 & 141.1 & \textbf{24.6} & 0.814 & 0.870 & & \textbf{0.758} & \textbf{0.832} \\
    PMA-Net~\cite{barraco2023little} & & 43.0 & 30.6 & 61.1 & 144.1 & 24.0 & 0.814 & 0.869 & & 0.755 & 0.821 \\
    \midrule
    Transformer (SCST) & & 43.6 & 30.8 & 61.0 & 143.3 & 23.2 & 0.809 & 0.866 & & 0.750 & 0.826 \\
    \rowcolor{LightCyan}
    \textbf{Transformer (\ours} w/ $\beta=0.05$) & & 43.2 & 31.2 & 61.1 & \textbf{144.2} & 24.4 & 0.815 & 0.871 & & 0.756 & 0.831 \\
    \rowcolor{LightCyan}
    \textbf{Transformer (\ours} w/ $\beta=0.1$) & & \textbf{43.7} & 31.2 & 61.2 & 143.8 & 24.5 & \textbf{0.817} & \textbf{0.872} & & 0.757 & \textbf{0.832} \\
    \rowcolor{LightCyan}
    \textbf{Transformer (\ours} w/ $\beta=0.2$) & & \textbf{43.7} & \textbf{31.3} & \textbf{61.3} & 143.5 & 24.4 & 0.816 & \textbf{0.872} & & 0.756 & 0.831 \\
    \bottomrule
  \end{tabular}
}
  \vspace{0.25cm}
  \caption{Comparison with standard captioners using CIDEr-based optimization.}
  \label{tab:cider}
    \vspace{-0.3cm}
\end{table}

\section{Conclusion}
\label{sec:conclusion}
We presented \ours, a novel fine-tuning strategy for image captioning which aligns a model to a learnable evaluator with high human correlation. Our approach optimizes a distilled reward model by solving a weighted classification problem directly inside the captioner, which allows it to capture fine-grained differences between multiple candidate captions. Experimental results on several datasets, conducted through automatic metrics and human evaluations, validate the effectiveness of our approach, which can generate more descriptive and detailed captions than competitors. At the same time, it achieves state-of-the-art results when trained to optimize traditional reference-based metrics.

\subsection*{Acknowledgments}
We acknowledge the CINECA award under the ISCRA initiative, for the availability of high-performance computing resources. This work has been conducted under a research grant co-funded by Altilia s.r.l. and supported by the PNRR-M4C2 (PE00000013) project ``FAIR - Future Artificial Intelligence Research'', by the PNRR project ``ITSERR - Italian Strengthening of Esfri RI Resilience'' (CUP B53C22001770006), and by the PRIN project ``MUSMA'' (CUP G53D23002930006 - M4C2 I1.1), all funded by EU - Next-Generation EU.

\bibliography{bibliography}

\begin{thebibliography}{74}
\providecommand{\natexlab}[1]{#1}
\providecommand{\url}[1]{\texttt{#1}}
\expandafter\ifx\csname urlstyle\endcsname\relax
  \providecommand{\doi}[1]{doi: #1}\else
  \providecommand{\doi}{doi: \begingroup \urlstyle{rm}\Url}\fi

\bibitem[Aditya et~al.(2015)Aditya, Yang, Baral, Fermuller, and Aloimonos]{aditya2015images}
Somak Aditya, Yezhou Yang, Chitta Baral, Cornelia Fermuller, and Yiannis Aloimonos.
\newblock {From Images to Sentences through Scene Description Graphs using Commonsense Reasoning and Knowledge}.
\newblock \emph{arXiv preprint arXiv:1511.03292}, 2015.

\bibitem[Agrawal et~al.(2019)Agrawal, Desai, Chen, Jain, Batra, Parikh, Lee, and Anderson]{agrawal2019nocaps}
Harsh Agrawal, Karan Desai, Xinlei Chen, Rishabh Jain, Dhruv Batra, Devi Parikh, Stefan Lee, and Peter Anderson.
\newblock nocaps: novel object captioning at scale.
\newblock In \emph{ICCV}, 2019.

\bibitem[Anderson et~al.(2016)Anderson, Fernando, Johnson, and Gould]{spice2016}
Peter Anderson, Basura Fernando, Mark Johnson, and Stephen Gould.
\newblock {SPICE: Semantic Propositional Image Caption Evaluation}.
\newblock In \emph{ECCV}, 2016.

\bibitem[Anderson et~al.(2018)Anderson, He, Buehler, Teney, Johnson, Gould, and Zhang]{anderson2018bottom}
Peter Anderson, Xiaodong He, Chris Buehler, Damien Teney, Mark Johnson, Stephen Gould, and Lei Zhang.
\newblock Bottom-up and top-down attention for image captioning and visual question answering.
\newblock In \emph{CVPR}, 2018.

\bibitem[Banerjee and Lavie(2005)]{banerjee2005meteor}
Satanjeev Banerjee and Alon Lavie.
\newblock {METEOR: An automatic metric for MT evaluation with improved correlation with human judgments}.
\newblock In \emph{ACL Workshops}, 2005.

\bibitem[Barraco et~al.(2023)Barraco, Sarto, Cornia, Baraldi, and Cucchiara]{barraco2023little}
Manuele Barraco, Sara Sarto, Marcella Cornia, Lorenzo Baraldi, and Rita Cucchiara.
\newblock {With a Little Help from your own Past: Prototypical Memory Networks for Image Captioning}.
\newblock In \emph{ICCV}, 2023.

\bibitem[Bolelli et~al.(2018)Bolelli, Baraldi, and Grana]{2018IPAS_QRNN}
Federico Bolelli, Lorenzo Baraldi, and Costantino Grana.
\newblock {A Hierarchical Quasi-Recurrent approach to Video Captioning}.
\newblock In \emph{IPAS}, 2018.

\bibitem[Caffagni et~al.(2024{\natexlab{a}})Caffagni, Cocchi, Barsellotti, Moratelli, Sarto, Baraldi, Baraldi, Cornia, and Cucchiara]{caffagni2024r}
Davide Caffagni, Federico Cocchi, Luca Barsellotti, Nicholas Moratelli, Sara Sarto, Lorenzo Baraldi, Lorenzo Baraldi, Marcella Cornia, and Rita Cucchiara.
\newblock {The Revolution of Multimodal Large Language Models: A Survey}.
\newblock In \emph{ACL Findings}, 2024{\natexlab{a}}.

\bibitem[Caffagni et~al.(2024{\natexlab{b}})Caffagni, Cocchi, Moratelli, Sarto, Cornia, Baraldi, and Cucchiara]{caffagni2024wiki}
Davide Caffagni, Federico Cocchi, Nicholas Moratelli, Sara Sarto, Marcella Cornia, Lorenzo Baraldi, and Rita Cucchiara.
\newblock {Wiki-LLaVA: Hierarchical Retrieval-Augmented Generation for Multimodal LLMs}.
\newblock In \emph{CVPR Workshops}, 2024{\natexlab{b}}.

\bibitem[Chan et~al.(2023{\natexlab{a}})Chan, Petryk, Gonzalez, Darrell, and Canny]{chan2023clair}
David Chan, Suzanne Petryk, Joseph~E Gonzalez, Trevor Darrell, and John Canny.
\newblock {CLAIR: Evaluating Image Captions with Large Language Models}.
\newblock In \emph{EMNLP}, 2023{\natexlab{a}}.

\bibitem[Chan et~al.(2023{\natexlab{b}})Chan, Myers, Vijayanarasimhan, Ross, and Canny]{chan2023ic}
David~M Chan, Austin Myers, Sudheendra Vijayanarasimhan, David~A Ross, and John Canny.
\newblock {IC$^{3}$: Image Captioning by Committee Consensus}.
\newblock In \emph{EMNLP}, 2023{\natexlab{b}}.

\bibitem[Chen et~al.(2023)Chen, Li, Zhang, Xiong, and Elhoseiny]{chen2023minigpt}
Jun Chen, Deyao Zhu1 Xiaoqian Shen1~Xiang Li, Zechun Liu2~Pengchuan Zhang, Raghuraman Krishnamoorthi2 Vikas Chandra2~Yunyang Xiong, and Mohamed Elhoseiny.
\newblock {MiniGPT-v2: Large Language Model as a Unified Interface for Vision-Language Multi-task Learning}.
\newblock \emph{arXiv preprint arXiv:2310.09478}, 2023.

\bibitem[Chiang et~al.(2023)Chiang, Li, Lin, Sheng, Wu, Zhang, Zheng, Zhuang, Zhuang, Gonzalez, Stoica, and Xing]{vicuna2023}
Wei-Lin Chiang, Zhuohan Li, Zi~Lin, Ying Sheng, Zhanghao Wu, Hao Zhang, Lianmin Zheng, Siyuan Zhuang, Yonghao Zhuang, Joseph~E. Gonzalez, Ion Stoica, and Eric~P. Xing.
\newblock {Vicuna: An Open-Source Chatbot Impressing GPT-4 with 90\%* ChatGPT Quality}, 2023.

\bibitem[Cho et~al.(2022)Cho, Yoon, Kale, Dernoncourt, Bui, and Bansal]{cho2022fine}
Jaemin Cho, Seunghyun Yoon, Ajinkya Kale, Franck Dernoncourt, Trung Bui, and Mohit Bansal.
\newblock {Fine-grained Image Captioning with CLIP Reward}.
\newblock In \emph{NAACL}, 2022.

\bibitem[Christiano et~al.(2017)Christiano, Leike, Brown, Martic, Legg, and Amodei]{christiano2017deep}
Paul~F Christiano, Jan Leike, Tom Brown, Miljan Martic, Shane Legg, and Dario Amodei.
\newblock Deep reinforcement learning from human preferences.
\newblock In \emph{NeurIPS}, 2017.

\bibitem[Cornia et~al.(2020)Cornia, Stefanini, Baraldi, and Cucchiara]{cornia2020meshed}
Marcella Cornia, Matteo Stefanini, Lorenzo Baraldi, and Rita Cucchiara.
\newblock {Meshed-Memory Transformer for Image Captioning}.
\newblock In \emph{CVPR}, 2020.

\bibitem[Gao et~al.(2023)Gao, Han, Zhang, Lin, Geng, Zhou, Zhang, Lu, He, Yue, et~al.]{gao2023llama}
Peng Gao, Jiaming Han, Renrui Zhang, Ziyi Lin, Shijie Geng, Aojun Zhou, Wei Zhang, Pan Lu, Conghui He, Xiangyu Yue, et~al.
\newblock {LLaMA-Adapter V2: Parameter-Efficient Visual Instruction Model}.
\newblock \emph{arXiv preprint arXiv:2304.15010}, 2023.

\bibitem[Gurari et~al.(2020)Gurari, Zhao, Zhang, and Bhattacharya]{gurari2020captioning}
Danna Gurari, Yinan Zhao, Meng Zhang, and Nilavra Bhattacharya.
\newblock {Captioning Images Taken by People Who Are Blind}.
\newblock In \emph{ECCV}, 2020.

\bibitem[Hessel et~al.(2021)Hessel, Holtzman, Forbes, Bras, and Choi]{hessel2021clipscore}
Jack Hessel, Ari Holtzman, Maxwell Forbes, Ronan~Le Bras, and Yejin Choi.
\newblock {CLIPScore: A Reference-free Evaluation Metric for Image Captioning}.
\newblock In \emph{EMNLP}, 2021.

\bibitem[Hodosh et~al.(2013)Hodosh, Young, and Hockenmaier]{hodosh2013framing}
Micah Hodosh, Peter Young, and Julia Hockenmaier.
\newblock Framing image description as a ranking task: Data, models and evaluation metrics.
\newblock \emph{JAIR}, 47:\penalty0 853--899, 2013.

\bibitem[Hu et~al.(2022)Hu, Gan, Wang, Yang, Liu, Lu, and Wang]{hu2022scaling}
Xiaowei Hu, Zhe Gan, Jianfeng Wang, Zhengyuan Yang, Zicheng Liu, Yumao Lu, and Lijuan Wang.
\newblock {Scaling Up Vision-Language Pre-training for Image Captioning}.
\newblock In \emph{CVPR}, 2022.

\bibitem[Huang et~al.(2019)Huang, Wang, Chen, and Wei]{huang2019attention}
Lun Huang, Wenmin Wang, Jie Chen, and Xiao-Yong Wei.
\newblock {Attention on Attention for Image Captioning}.
\newblock In \emph{ICCV}, 2019.

\bibitem[Iyer et~al.(2022)Iyer, Lin, Pasunuru, Mihaylov, Simig, Yu, Shuster, Wang, et~al.]{iyer2022opt}
Srinivasan Iyer, Xi~Victoria Lin, Ramakanth Pasunuru, Todor Mihaylov, Daniel Simig, Ping Yu, Kurt Shuster, Tianlu Wang, et~al.
\newblock {OPT-IML: Scaling Language Model Instruction Meta Learning through the Lens of Generalization}.
\newblock \emph{arXiv preprint arXiv:2212.12017}, 2022.

\bibitem[Karpathy and Fei-Fei(2015)]{karpathy2015deep}
Andrej Karpathy and Li~Fei-Fei.
\newblock Deep visual-semantic alignments for generating image descriptions.
\newblock In \emph{CVPR}, 2015.

\bibitem[Koh et~al.(2023{\natexlab{a}})Koh, Fried, and Salakhutdinov]{koh2023generating}
Jing~Yu Koh, Daniel Fried, and Ruslan Salakhutdinov.
\newblock {Generating Images with Multimodal Language Models}.
\newblock In \emph{NeurIPS}, 2023{\natexlab{a}}.

\bibitem[Koh et~al.(2023{\natexlab{b}})Koh, Salakhutdinov, and Fried]{koh2023grounding}
Jing~Yu Koh, Ruslan Salakhutdinov, and Daniel Fried.
\newblock {Grounding Language Models to Images for Multimodal Inputs and Outputs}.
\newblock In \emph{ICML}, 2023{\natexlab{b}}.

\bibitem[Kornblith et~al.(2023)Kornblith, Li, Wang, and Nguyen]{kornblith2023guiding}
Simon Kornblith, Lala Li, Zirui Wang, and Thao Nguyen.
\newblock {Guiding Image Captioning Models Toward More Specific Captions}.
\newblock In \emph{ICCV}, 2023.

\bibitem[Lester et~al.(2021)Lester, Al-Rfou, and Constant]{lester2021power}
Brian Lester, Rami Al-Rfou, and Noah Constant.
\newblock The power of scale for parameter-efficient prompt tuning.
\newblock In \emph{EMNLP}, 2021.

\bibitem[Li et~al.(2022{\natexlab{a}})Li, Li, Xiong, and Hoi]{li2022blip}
Junnan Li, Dongxu Li, Caiming Xiong, and Steven Hoi.
\newblock {BLIP: Bootstrapping Language-Image Pre-training for Unified Vision-Language Understanding and Generation}.
\newblock In \emph{ICML}, 2022{\natexlab{a}}.

\bibitem[Li et~al.(2023)Li, Li, Savarese, and Hoi]{li2023blip}
Junnan Li, Dongxu Li, Silvio Savarese, and Steven Hoi.
\newblock {BLIP-2: Bootstrapping Language-Image Pre-training with Frozen Image Encoders and Large Language Models}.
\newblock In \emph{ICML}, 2023.

\bibitem[Li et~al.(2020)Li, Yin, Li, Zhang, Hu, Zhang, Wang, Hu, Dong, Wei, et~al.]{li2020oscar}
Xiujun Li, Xi~Yin, Chunyuan Li, Pengchuan Zhang, Xiaowei Hu, Lei Zhang, Lijuan Wang, Houdong Hu, Li~Dong, Furu Wei, et~al.
\newblock {Oscar: Object-Semantics Aligned Pre-training for Vision-Language Tasks}.
\newblock In \emph{ECCV}, 2020.

\bibitem[Li et~al.(2022{\natexlab{b}})Li, Pan, Yao, and Mei]{li2022comprehending}
Yehao Li, Yingwei Pan, Ting Yao, and Tao Mei.
\newblock Comprehending and ordering semantics for image captioning.
\newblock In \emph{CVPR}, 2022{\natexlab{b}}.

\bibitem[Lin(2004)]{lin2004rouge}
Chin-Yew Lin.
\newblock Rouge: A package for automatic evaluation of summaries.
\newblock In \emph{ACL Workshops}, 2004.

\bibitem[Lin et~al.(2014)Lin, Maire, Belongie, Hays, Perona, Ramanan, Doll{\'a}r, and Zitnick]{lin2014microsoft}
Tsung-Yi Lin, Michael Maire, Serge Belongie, James Hays, Pietro Perona, Deva Ramanan, Piotr Doll{\'a}r, and C~Lawrence Zitnick.
\newblock {Microsoft COCO: Common Objects in Context}.
\newblock In \emph{ECCV}, 2014.

\bibitem[Liu et~al.(2023{\natexlab{a}})Liu, Li, Wu, and Lee]{liu2023visual}
Haotian Liu, Chunyuan Li, Qingyang Wu, and Yong~Jae Lee.
\newblock {Visual Instruction Tuning}.
\newblock In \emph{NeurIPS}, 2023{\natexlab{a}}.

\bibitem[Liu et~al.(2024)Liu, Li, Li, and Lee]{liu2024improved}
Haotian Liu, Chunyuan Li, Yuheng Li, and Yong~Jae Lee.
\newblock Improved baselines with visual instruction tuning.
\newblock In \emph{CVPR}, 2024.

\bibitem[Liu et~al.(2017)Liu, Zhu, Ye, Guadarrama, and Murphy]{liu2017improved}
Siqi Liu, Zhenhai Zhu, Ning Ye, Sergio Guadarrama, and Kevin Murphy.
\newblock {Improved Image Captioning via Policy Gradient Optimization of SPIDEr}.
\newblock In \emph{ICCV}, 2017.

\bibitem[Liu et~al.(2023{\natexlab{b}})Liu, Zheng, Du, Ding, Qian, Yang, and Tang]{liu2023gpt}
Xiao Liu, Yanan Zheng, Zhengxiao Du, Ming Ding, Yujie Qian, Zhilin Yang, and Jie Tang.
\newblock {GPT understands, too}.
\newblock \emph{AI Open}, 2023{\natexlab{b}}.

\bibitem[Micikevicius et~al.(2018)Micikevicius, Narang, Alben, Diamos, Elsen, Garcia, Ginsburg, Houston, Kuchaiev, Venkatesh, and Wu]{micikevicius2017mixed}
Paulius Micikevicius, Sharan Narang, Jonah Alben, Gregory Diamos, Erich Elsen, David Garcia, Boris Ginsburg, Michael Houston, Oleksii Kuchaiev, Ganesh Venkatesh, and Hao Wu.
\newblock {Mixed Precision Training}.
\newblock In \emph{ICLR}, 2018.

\bibitem[Mokady et~al.(2021)Mokady, Hertz, and Bermano]{mokady2021clipcap}
Ron Mokady, Amir Hertz, and Amit~H Bermano.
\newblock {ClipCap: CLIP Prefix for Image Captioning}.
\newblock \emph{arXiv preprint arXiv:2111.09734}, 2021.

\bibitem[Ouyang et~al.(2022)Ouyang, Wu, Jiang, Almeida, Wainwright, Mishkin, Zhang, Agarwal, Slama, Ray, et~al.]{ouyang2022training}
Long Ouyang, Jeffrey Wu, Xu~Jiang, Diogo Almeida, Carroll Wainwright, Pamela Mishkin, Chong Zhang, Sandhini Agarwal, Katarina Slama, Alex Ray, et~al.
\newblock Training language models to follow instructions with human feedback.
\newblock In \emph{NeurIPS}, 2022.

\bibitem[Papineni et~al.(2002)Papineni, Roukos, Ward, and Zhu]{papineni2002bleu}
Kishore Papineni, Salim Roukos, Todd Ward, and Wei-Jing Zhu.
\newblock {BLEU: a method for automatic evaluation of machine translation}.
\newblock In \emph{ACL}, 2002.

\bibitem[Radford et~al.(2019)Radford, Wu, Child, Luan, Amodei, Sutskever, et~al.]{radford2019language}
Alec Radford, Jeffrey Wu, Rewon Child, David Luan, Dario Amodei, Ilya Sutskever, et~al.
\newblock Language models are unsupervised multitask learners.
\newblock \emph{OpenAI blog}, 1\penalty0 (8):\penalty0 9, 2019.

\bibitem[Radford et~al.(2021)Radford, Kim, Hallacy, Ramesh, Goh, Agarwal, Sastry, Askell, Mishkin, Clark, et~al.]{radford2021learning}
Alec Radford, Jong~Wook Kim, Chris Hallacy, Aditya Ramesh, Gabriel Goh, Sandhini Agarwal, Girish Sastry, Amanda Askell, Pamela Mishkin, Jack Clark, et~al.
\newblock {Learning Transferable Visual Models From Natural Language Supervision}.
\newblock In \emph{ICML}, 2021.

\bibitem[Rafailov et~al.(2024)Rafailov, Sharma, Mitchell, Manning, Ermon, and Finn]{rafailov2024direct}
Rafael Rafailov, Archit Sharma, Eric Mitchell, Christopher~D Manning, Stefano Ermon, and Chelsea Finn.
\newblock Direct preference optimization: Your language model is secretly a reward model.
\newblock In \emph{NeurIPS}, 2024.

\bibitem[Ramos et~al.(2023)Ramos, Martins, Elliott, and Kementchedjhieva]{ramos2023smallcap}
Rita Ramos, Bruno Martins, Desmond Elliott, and Yova Kementchedjhieva.
\newblock {SmallCap: Lightweight Image Captioning Prompted With Retrieval Augmentation}.
\newblock In \emph{CVPR}, 2023.

\bibitem[Ranzato et~al.(2015)Ranzato, Chopra, Auli, and Zaremba]{ranzato2015sequence}
Marc'Aurelio Ranzato, Sumit Chopra, Michael Auli, and Wojciech Zaremba.
\newblock {Sequence Level Training with Recurrent Neural Networks}.
\newblock In \emph{ICLR}, 2015.

\bibitem[Rennie et~al.(2017)Rennie, Marcheret, Mroueh, Ross, and Goel]{rennie2017self}
Steven~J Rennie, Etienne Marcheret, Youssef Mroueh, Jarret Ross, and Vaibhava Goel.
\newblock {Self-Critical Sequence Training for Image Captioning}.
\newblock In \emph{CVPR}, 2017.

\bibitem[Rotstein et~al.(2024)Rotstein, Bensaid, Brody, Ganz, and Kimmel]{rotstein2023fusecap}
Noam Rotstein, David Bensaid, Shaked Brody, Roy Ganz, and Ron Kimmel.
\newblock {FuseCap: Leveraging Large Language Models to Fuse Visual Data into Enriched Image Captions}.
\newblock In \emph{WACV}, 2024.

\bibitem[Sarto et~al.(2023)Sarto, Barraco, Cornia, Baraldi, and Cucchiara]{sarto2023positive}
Sara Sarto, Manuele Barraco, Marcella Cornia, Lorenzo Baraldi, and Rita Cucchiara.
\newblock {Positive-Augmented Contrastive Learning for Image and Video Captioning Evaluation}.
\newblock In \emph{CVPR}, 2023.

\bibitem[Sarto et~al.(2024)Sarto, Cornia, Baraldi, and Cucchiara]{sarto2024bridge}
Sara Sarto, Marcella Cornia, Lorenzo Baraldi, and Rita Cucchiara.
\newblock {BRIDGE: Bridging Gaps in Image Captioning Evaluation with Stronger Visual Cues}.
\newblock In \emph{ECCV}, 2024.

\bibitem[Schulman et~al.(2017)Schulman, Wolski, Dhariwal, Radford, and Klimov]{schulman2017proximal}
John Schulman, Filip Wolski, Prafulla Dhariwal, Alec Radford, and Oleg Klimov.
\newblock {Proximal Policy Optimization Algorithms}.
\newblock \emph{arXiv preprint arXiv:1707.06347}, 2017.

\bibitem[Sharma et~al.(2018)Sharma, Ding, Goodman, and Soricut]{sharma2018conceptual}
Piyush Sharma, Nan Ding, Sebastian Goodman, and Radu Soricut.
\newblock {Conceptual Captions: A Cleaned, Hypernymed, Image Alt-text Dataset For Automatic Image Captioning}.
\newblock In \emph{ACL}, 2018.

\bibitem[Shen et~al.(2022)Shen, Li, Tan, Bansal, Rohrbach, Chang, Yao, and Keutzer]{shen2022much}
Sheng Shen, Liunian~Harold Li, Hao Tan, Mohit Bansal, Anna Rohrbach, Kai-Wei Chang, Zhewei Yao, and Kurt Keutzer.
\newblock {How Much Can CLIP Benefit Vision-and-Language Tasks?}
\newblock In \emph{ICLR}, 2022.

\bibitem[Sidorov et~al.(2020)Sidorov, Hu, Rohrbach, and Singh]{sidorov2020textcaps}
Oleksii Sidorov, Ronghang Hu, Marcus Rohrbach, and Amanpreet Singh.
\newblock {TextCaps: A Dataset for Image Captioning with Reading Comprehension}.
\newblock In \emph{ECCV}, 2020.

\bibitem[Stefanini et~al.(2022)Stefanini, Cornia, Baraldi, Cascianelli, Fiameni, and Cucchiara]{stefanini2021show}
Matteo Stefanini, Marcella Cornia, Lorenzo Baraldi, Silvia Cascianelli, Giuseppe Fiameni, and Rita Cucchiara.
\newblock {From Show to Tell: A Survey on Deep Learning-based Image Captioning}.
\newblock \emph{IEEE Trans. PAMI}, 45\penalty0 (1):\penalty0 539--559, 2022.

\bibitem[Sun et~al.(2023)Sun, Shen, Zhou, Zhang, Chen, Cox, Yang, and Gan]{sun2023principle}
Zhiqing Sun, Yikang Shen, Qinhong Zhou, Hongxin Zhang, Zhenfang Chen, David Cox, Yiming Yang, and Chuang Gan.
\newblock Principle-driven self-alignment of language models from scratch with minimal human supervision.
\newblock In \emph{NeurIPS}, 2023.

\bibitem[Tewel et~al.(2022)Tewel, Shalev, Schwartz, and Wolf]{tewel2022zerocap}
Yoad Tewel, Yoav Shalev, Idan Schwartz, and Lior Wolf.
\newblock {ZeroCap: Zero-Shot Image-to-Text Generation for Visual-Semantic Arithmetic}.
\newblock In \emph{CVPR}, 2022.

\bibitem[Touvron et~al.(2023)Touvron, Martin, Stone, Albert, Almahairi, Babaei, Bashlykov, Batra, Bhargava, Bhosale, et~al.]{touvron2023llama}
Hugo Touvron, Louis Martin, Kevin Stone, Peter Albert, Amjad Almahairi, Yasmine Babaei, Nikolay Bashlykov, Soumya Batra, Prajjwal Bhargava, Shruti Bhosale, et~al.
\newblock Llama 2: Open foundation and fine-tuned chat models.
\newblock \emph{arXiv preprint arXiv:2307.09288}, 2023.

\bibitem[Vedantam et~al.(2015)Vedantam, Lawrence~Zitnick, and Parikh]{vedantam2015cider}
Ramakrishna Vedantam, C~Lawrence~Zitnick, and Devi Parikh.
\newblock {CIDEr: Consensus-based Image Description Evaluation}.
\newblock In \emph{CVPR}, 2015.

\bibitem[Vinyals et~al.(2015)Vinyals, Toshev, Bengio, and Erhan]{vinyals2015show}
Oriol Vinyals, Alexander Toshev, Samy Bengio, and Dumitru Erhan.
\newblock Show and tell: A neural image caption generator.
\newblock In \emph{CVPR}, 2015.

\bibitem[Wang et~al.(2022{\natexlab{a}})Wang, Yang, Hu, Li, Lin, Gan, Liu, Liu, and Wang]{wang2022git}
Jianfeng Wang, Zhengyuan Yang, Xiaowei Hu, Linjie Li, Kevin Lin, Zhe Gan, Zicheng Liu, Ce~Liu, and Lijuan Wang.
\newblock {GIT: A Generative Image-to-text Transformer for Vision and Language}.
\newblock \emph{arXiv preprint arXiv:2205.14100}, 2022{\natexlab{a}}.

\bibitem[Wang et~al.(2022{\natexlab{b}})Wang, Yang, Men, Lin, Bai, Li, Ma, Zhou, Zhou, and Yang]{wang2022ofa}
Peng Wang, An~Yang, Rui Men, Junyang Lin, Shuai Bai, Zhikang Li, Jianxin Ma, Chang Zhou, Jingren Zhou, and Hongxia Yang.
\newblock {OFA: Unifying Architectures, Tasks, and Modalities Through a Simple Sequence-to-Sequence Learning Framework}.
\newblock In \emph{ICML}, 2022{\natexlab{b}}.

\bibitem[Wang et~al.(2022{\natexlab{c}})Wang, Mishra, Alipoormolabashi, Kordi, Mirzaei, Arunkumar, Ashok, Dhanasekaran, Naik, Stap, et~al.]{wang2022super}
Yizhong Wang, Swaroop Mishra, Pegah Alipoormolabashi, Yeganeh Kordi, Amirreza Mirzaei, Anjana Arunkumar, Arjun Ashok, Arut~Selvan Dhanasekaran, Atharva Naik, David Stap, et~al.
\newblock {Super-NaturalInstructions: Generalization via Declarative Instructions on 1600+ NLP Tasks}.
\newblock In \emph{EMNLP}, 2022{\natexlab{c}}.

\bibitem[Wang et~al.(2023)Wang, Kordi, Mishra, Liu, Smith, Khashabi, and Hajishirzi]{wang2022self}
Yizhong Wang, Yeganeh Kordi, Swaroop Mishra, Alisa Liu, Noah~A Smith, Daniel Khashabi, and Hannaneh Hajishirzi.
\newblock {Self-Instruct: Aligning Language Models with Self-Generated Instructions}.
\newblock In \emph{ACL}, 2023.

\bibitem[Xiong et~al.(2018)Xiong, Dai, and Lin]{xiong2018move}
Yilei Xiong, Bo~Dai, and Dahua Lin.
\newblock {Move Forward and Tell: A Progressive Generator of Video Descriptions}.
\newblock In \emph{ECCV}, 2018.

\bibitem[Xu et~al.(2015)Xu, Ba, Kiros, Cho, Courville, Salakhutdinov, Zemel, and Bengio]{xu2015show}
Kelvin Xu, Jimmy Ba, Ryan Kiros, Kyunghyun Cho, Aaron Courville, Ruslan Salakhutdinov, Richard~S Zemel, and Yoshua Bengio.
\newblock {Show, Attend and Tell: Neural Image Caption Generation with Visual Attention}.
\newblock In \emph{ICML}, 2015.

\bibitem[Yang et~al.(2019)Yang, Tang, Zhang, and Cai]{yang2019auto}
Xu~Yang, Kaihua Tang, Hanwang Zhang, and Jianfei Cai.
\newblock {Auto-Encoding Scene Graphs for Image Captioning}.
\newblock In \emph{CVPR}, 2019.

\bibitem[Yao et~al.(2018)Yao, Pan, Li, and Mei]{yao2018exploring}
Ting Yao, Yingwei Pan, Yehao Li, and Tao Mei.
\newblock {Exploring Visual Relationship for Image Captioning}.
\newblock In \emph{ECCV}, 2018.

\bibitem[Young et~al.(2014)Young, Lai, Hodosh, and Hockenmaier]{young2014image}
Peter Young, Alice Lai, Micah Hodosh, and Julia Hockenmaier.
\newblock {From image descriptions to visual denotations: New similarity metrics for semantic inference over event descriptions}.
\newblock \emph{TACL}, 2:\penalty0 67--78, 2014.

\bibitem[Zhang et~al.(2021)Zhang, Li, Hu, Yang, Zhang, Wang, Choi, and Gao]{zhang2021vinvl}
Pengchuan Zhang, Xiujun Li, Xiaowei Hu, Jianwei Yang, Lei Zhang, Lijuan Wang, Yejin Choi, and Jianfeng Gao.
\newblock {VinVL: Revisiting visual representations in vision-language models}.
\newblock In \emph{CVPR}, 2021.

\bibitem[Zhang et~al.(2022)Zhang, Wang, Wu, and Xu]{zhang2022distinctive}
Youyuan Zhang, Jiuniu Wang, Hao Wu, and Wenjia Xu.
\newblock {Distinctive Image Captioning via CLIP Guided Group Optimization}.
\newblock In \emph{ECCV}, 2022.

\bibitem[Zhu et~al.(2023)Zhu, Chen, Haydarov, Shen, Zhang, and Elhoseiny]{zhu2023chatgpt}
Deyao Zhu, Jun Chen, Kilichbek Haydarov, Xiaoqian Shen, Wenxuan Zhang, and Mohamed Elhoseiny.
\newblock {ChatGPT Asks, BLIP-2 Answers: Automatic Questioning Towards Enriched Visual Descriptions}.
\newblock \emph{arXiv preprint arXiv:2303.06594}, 2023.

\bibitem[Ziegler et~al.(2019)Ziegler, Stiennon, Wu, Brown, Radford, Amodei, Christiano, and Irving]{ziegler2019fine}
Daniel~M Ziegler, Nisan Stiennon, Jeffrey Wu, Tom~B Brown, Alec Radford, Dario Amodei, Paul Christiano, and Geoffrey Irving.
\newblock {Fine-Tuning Language Models from Human Preferences}.
\newblock \emph{arXiv preprint arXiv:1909.08593}, 2019.

\end{thebibliography}

\vspace{1cm}
\appendix

\noindent In the following, we present additional materials about \ours. In particular, we provide additional analyses and ablation studies, comparing \ours with the standard SCST training paradigm. Moreover, we report further implementation details and qualitative results on all considered datasets and settings. 

\section{Preliminaries}
\label{sec:preliminaries}
In this section, we first recap the definition of the SCST and Reinforcement Learning from Human Feedback (RLHF) training protocols~\cite{rennie2017self,ouyang2022training}. Then, we introduce captioning metrics based on contrastive embedding spaces~\cite{radford2021learning}.

\tit{Self-critical sequence training} SCST~\cite{rennie2017self} is a two-step training methodology which (1) pre-trains a captioner $f_\theta$ using a time-wise cross-entropy loss with respect to ground-truth sequences, and (2) fine-tunes the same network by maximizing the CIDEr score~\cite{vedantam2015cider} using a reinforcement learning (RL) approach. We assume that the captioner takes as input an image $I$ described with a sequence of visual features $(v_1, v_2, ..., v_R)$, and a ground-truth sequence $s = (w_1, w_2, ..., w_T)$, where $w_i$ is a token belonging to a pre-defined vocabulary. Noticeably, depending on the dataset there might be multiple ground-truth sequences associated with each image. During the first training stage, the network is conditioned on visual features and all ground-truth tokens up to the current prediction step $t$, and $f_\theta$ is optimized using the cross-entropy loss (teacher forcing). In the second training stage, instead, the network is only conditioned on the input image and generates an entire caption $s' = (w'_1, w'_2, ..., w'_{T'})$ by sampling input tokens from the output probability distribution generated at the previous time step. For instance, $w'_t$ might be chosen as  $w'_t = \argmax f_\theta(w_t|w'_{t-1}, ..., w'_1, v_1, ..., v_R)$, or multiple sentences can be sampled via beam search. The generated sentences are then employed to compute the CIDEr metric, which is later used as a reward to guide a policy-gradient RL update step (see~\cite{rennie2017self} for details).

\tit{Reinforcement learning from human feedback}
Recent NLP literature has employed techniques based on RLHF~\cite{ouyang2022training} to align the behavior of a large language model to human preferences. This approach is usually based on the collection of large-scale datasets of human preferences: the language model $f_\theta$\footnote{With a slight abuse of notation, in this paragraph we use $f_\theta$ to refer to a single-modality language model.} is prompted with a prompt $x$ to produce pairs of answers $(s'_1, s'_2) \sim f_\theta$, which are then presented to human labelers who express preference for one answer, \ie~$s'_w \succ s'_l$, where $s'_w$ and $s'_l$ indicate, respectively, the preferred and dispreferred completion. The resulting dataset of human preferences $\mathcal{D} = \{ x_i, s'_{w,i}, s'_{l,i} \}_{i=1}^N$ is then employed to train a reward model on top of it~\cite{christiano2017deep}, for subsequent optimization with reinforcement learning. In image captioning, due to the lack in size of existing human preference datasets~\cite{hodosh2013framing,aditya2015images,vedantam2015cider}, \textit{training a learnable reward model to follow the RLHF approach is impracticable} (see also Sec.~\ref{sec:rlhf_comparison}). 

\tit{Learnable contrastive captioning metrics}
As pointed out by recent literature on captioning evaluation, a model learned with language-image pre-training~\cite{radford2021learning} can be straightforwardly employed as a captioning metric. Given a caption $s'$ generated from $I$, indeed, its correctness score can be defined as a function of the similarity predicted by the image-text model, \ie~$\text{sim}(I, s')$. A popular choice~\cite{hessel2021clipscore} is to define the score to be proportional to the ReLU of the predicted similarity and to employ a scalar multiplier $w$ to stretch the resulting score within the range of $\left[0,1\right]$: 
\begin{equation}
    \text{CLIP-S}(I, s') = w \cdot \text{ReLU}(\text{sim}(I, s')).
\end{equation}
In the original formulation of~\cite{hessel2021clipscore} (termed CLIP-S), the backbone employed for computing similarities was pre-trained on 400M noisy (image, text) pairs collected from the internet. While CLIP-S shows a significantly higher alignment with human judgments compared to traditional metrics (\eg~BLEU, METEOR, CIDEr), the noisy nature of the training data limits the CLIP-S capability to distinguish fluent human-generated captions. To overcome this issue, a recent choice~\cite{sarto2023positive} is that of fine-tuning the backbone on cleaned data, which further boosts the correlation with human judgments. Specifically, the PAC-S score~\cite{sarto2023positive} trains on the basis of a similarity matrix built with human-collected captions and machine-generated ones, where the latter are obtained from a captioner trained to mimic the same distribution of human captions.
In case a set of reference captions $R=\{r_i\}^{N}_{i=1}$ is given, there exists a version of the CLIP-based metrics accounting for them~\cite{hessel2021clipscore}, which is defined as follows:
\begin{equation}
    \text{RefCLIP-S}(I, s', R) = \text{H-Mean} (\text{CLIP-S}(I, s'), \text{ReLU}(
    \max_{r \in R}
    cos(s',r))).
\end{equation}
Following~\cite{sarto2023positive}, the same formula can be applied to compute the reference-based version of PAC-S (RefPAC-S).

\section{Ablation Studies}
\label{sec:ablations}
\tinytit{Early stopping condition}
When comparing multiple training strategies, we always employ an early stopping condition based on the validation value of the reference-based version of the metric used as a reward. In practice, when optimizing for CLIP-S, we early stop the training according to the validation RefCLIP-S, while when optimizing for PAC-S we early stop based on the validation RefPAC-S. We then take the model state corresponding to the epoch with the highest validation score and report its evaluation metrics. While this provides a reasonable evaluation strategy that equally promotes all compared approaches, evaluating a single model state does not capture the full training behavior of different fine-tuning strategies.

\begin{figure}[t]
\centering
\includegraphics[width=0.95\linewidth]{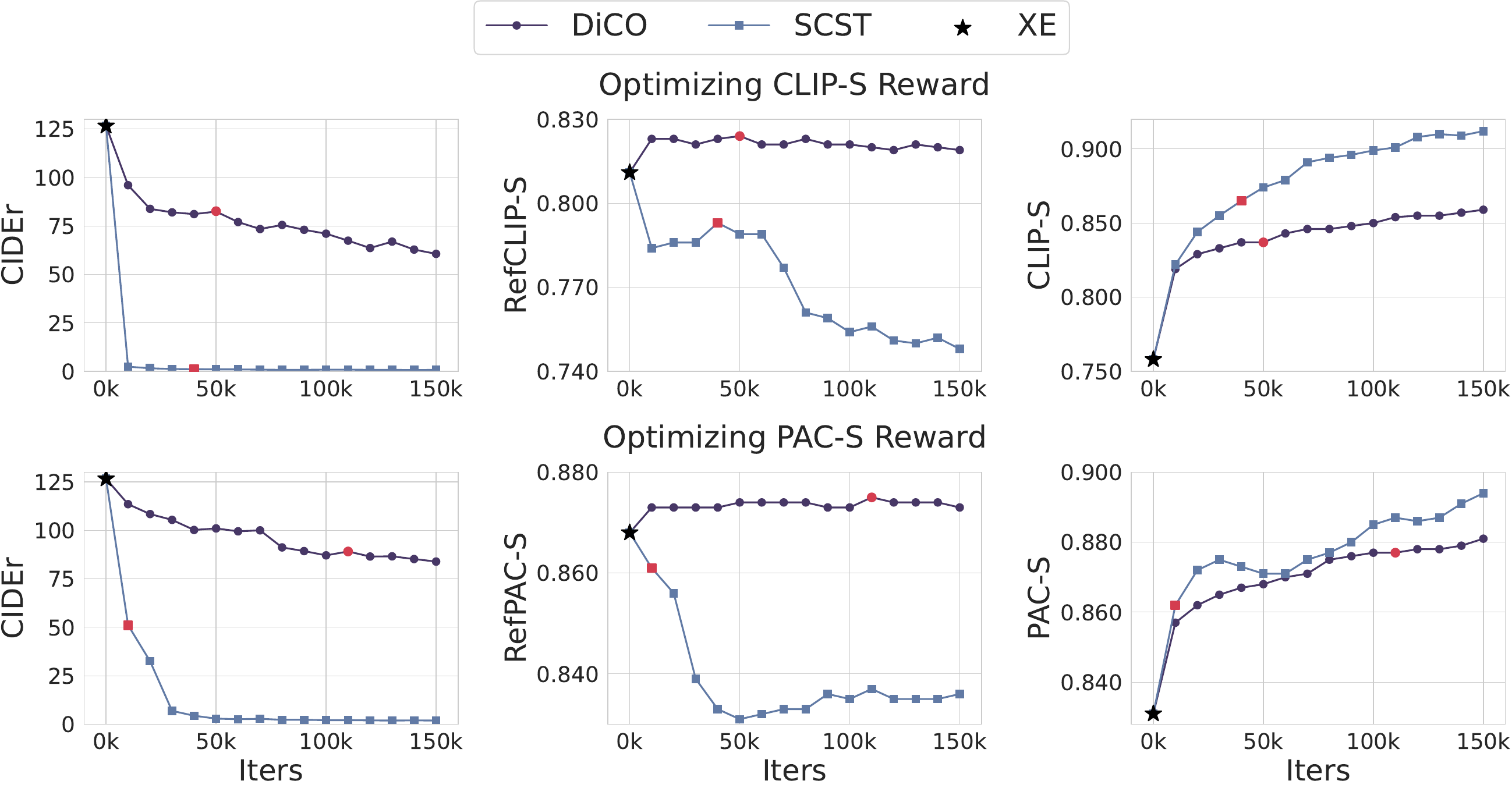}
\vspace{-0.15cm}
\caption{Metric curves when optimizing CLIP-S (top) and PAC-S (bottom) scores with \ours and SCST. The red dot indicates the early stopping point we employ.}
\label{fig:early_stopping}
\vspace{-0.2cm}
\end{figure}

To complement Fig.~\ref{fig:first_page} of the main paper, in Fig.~\ref{fig:early_stopping} we report the test curves of CIDEr, RefCLIP-S, and CLIP-S obtained when optimizing the CLIP-S score and again those of CIDEr, RefPAC-S, and PAC-S obtained when optimizing the PAC-S score. For both cases, we compare the results using \ours and SCST. With a red marker, we indicate the model state chosen by the early stopping condition, while a star marker indicates the model state after XE pre-training. As it can be seen, SCST hacks the reward metric immediately after the start of the fine-tuning phase, at the expense of CIDEr, RefCLIP-S, and RefPAC-S. Correspondingly, when optimizing using PAC-S as reward, the early stopping condition is forced to select the model state corresponding to the first fine-tuning epoch, which indeed showcases the highest RefPAC-S. Continuing the fine-tuning, though, would let SCST hack the reward metric even further and provide lower-quality captions. 

\begin{figure*}[t]
\centering
\vspace{-0.5cm}
\begin{minipage}{0.24\linewidth}
    \includegraphics[width=0.98\linewidth]{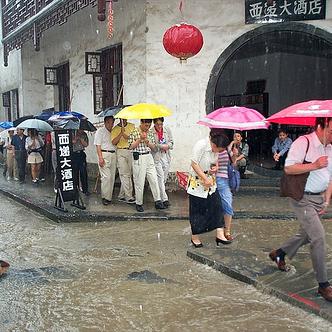}
    \end{minipage}
\begin{minipage}{0.75\linewidth}
    \scriptsize{
    \textbf{SCST (after 10k iters):} A group of people with umbrellas walking down a street with people walking down a wet sidewalk holding pink umbrellas in rain.\\
    \textbf{SCST (after 50k iters):} Many people crossing wet wet alley with people walking with colorful umbrellas outside a building with wet alley with people walking under umbrellas outside rainy surface.\\
    \textbf{SCST (after 100k iters):} Pedestrians walking down wet wet road with pedestrians carrying pink umbrellas outside a building on wet sidewalk outside rainy wall with buildings outside surface poles surface poles.\\
    \textbf{\ours (Ours):} A group of people walking down a wet sidewalk with umbrellas in the rain.
    }
\end{minipage}
\vspace{0.15cm}

\begin{minipage}{0.24\linewidth}
    \includegraphics[width=0.98\linewidth]{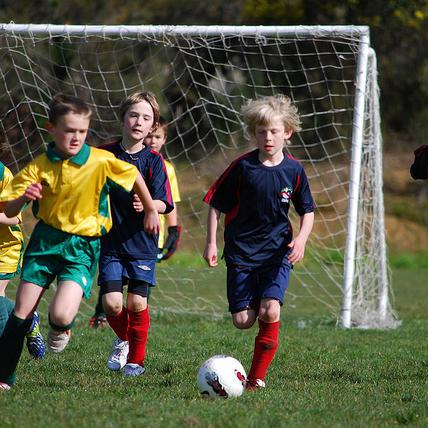}
    \end{minipage}
\begin{minipage}{0.75\linewidth}
    \scriptsize{
    \textbf{SCST (after 10k iters):} A group of young boys kicking around a soccer ball on a soccer field with other young boys running around with net in background.\\
    \textbf{SCST (after 50k iters):} Young boys kicking soccer ball around soccer goal kicking grass underneath a goal on grass behind background behind surface surface with trees in background behind surface surface.\\
    \textbf{SCST (after 100k iters):} Young boys soccer teams chasing after after soccer soccer goalie in background with green leaves on grass behind background surface court with young boy.\\
    \textbf{\ours (Ours):} A group of young children kicking a soccer ball in a field.
    }
\end{minipage}   
\vspace{0.15cm}

\begin{minipage}{0.24\linewidth}
    \includegraphics[width=0.98\linewidth]{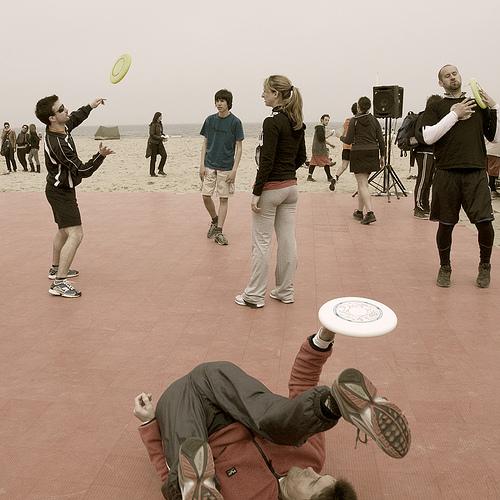}
    \end{minipage}
\begin{minipage}{0.75\linewidth}
    \scriptsize{
    \textbf{SCST (after 10k iters):} A group of people playing frisbee with a man laying on ground with a person laying on ground with other people in background.\\
    \textbf{SCST (after 50k iters):} Group of kids playing ultimate frisbee with man laying on ground with people on sand floor with frisbees while people gather around background behind surface surface.\\
    \textbf{SCST (after 100k iters):} A group of kids playground with man laying on cement floor playing frisbee game with man laying outside a crowd in background surface outside surface poles leg.\\
    \textbf{\ours (Ours):} A group of people playing with a frisbee on a beach with other people in the background.
    }
\end{minipage}
\vspace{0.15cm}

\begin{minipage}{0.24\linewidth}
    \includegraphics[width=0.98\linewidth]{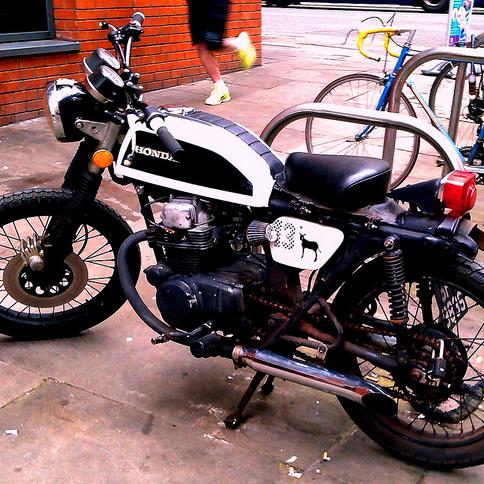}
    \end{minipage}
\begin{minipage}{0.75\linewidth}
    \scriptsize{
    \textbf{SCST (after 10k iters):}  A black motorcycle parked on a sidewalk next to a parked motorcycle on a sidewalk next to a rack with bicycles in background.\\
    \textbf{SCST (after 50k iters):} An old motorcycle parked on sidewalk with parked bicycle outside a brick background with other bikes on sidewalk outside clear background behind surface background behind surface surface.\\
    \textbf{SCST (after 100k iters):} Antique motorcycle parked outside a brick building with a silver seat outside a bike on a sidewalk with other bikes outside background surface outside surface poles top.\\
    \textbf{\ours (Ours):} A small black motorcycle parked on a sidewalk next to other bikes.
    }
\end{minipage}   
\vspace{0.15cm}

\begin{minipage}{0.24\linewidth}
    \includegraphics[width=0.98\linewidth]{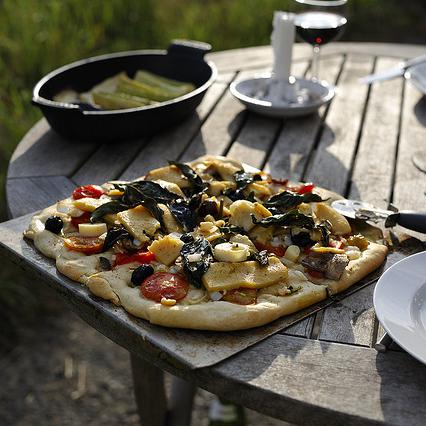}
    \end{minipage}
\begin{minipage}{0.75\linewidth}
    \scriptsize{
    \textbf{SCST (after 10k iters):} A small pizza with vegetables on a wooden picnic table with a pizza on a picnic table with silverware and wine in background.\\
    \textbf{SCST (after 50k iters):} Small vegetable pizza with vegetable vegetable on wooden picnic table with serving dish with other foods on grass outside clear surface behind background behind surface outside surface.\\
    \textbf{SCST (after 100k iters):} Cooked vegetable vegetable vegetable vegetable pizza served outside outside table with fork on picnic table outside a wine holder on sun surface outside background surface poles hand.\\
    \textbf{\ours (Ours):} A small pizza on a wooden picnic table with silverware and a wine glass in the background.
    }
\end{minipage}
\vspace{-0.15cm}
\caption{Qualitative results on sample images from the COCO Karpathy test split~\cite{lin2014microsoft} using SCST optimization with PAC-S reward at different fine-tuning states, in comparison with \ours.}
\label{fig:early_stopping_qualitatives}
\end{figure*}

On the contrary, \ours showcases a more robust training behavior. While CLIP-S and PAC-S values increase during fine-tuning as a result of the optimization process, the decrease in CIDEr is well restrained, while RefCLIP-S and RefPAC-S even increase with respect to the XE state. This highlights that \ours can optimize modern captioning metrics without incurring reward hacking and without deviating from a fluent and high-quality generation. Finally, in Fig.~\ref{fig:early_stopping_qualitatives} we also report sample captions from the COCO Karpathy test split when optimizing the PAC-S score with SCST at different training stages, in comparison with \ours. While SCST optimization tends to produce degraded and repetitive captions over time, \ours maintains fluency and generation quality.

\begin{table}[t]
  \centering
  \setlength{\tabcolsep}{.3em}
  \resizebox{0.98\linewidth}{!}{
  \begin{tabular}{lc c ccccc c cccc}
    \toprule
    & & & \multicolumn{5}{c}{\textbf{Reference-based}} & & \multicolumn{4}{c}{\textbf{Reference-free}} \\
    \cmidrule{4-8} \cmidrule{10-13}
    \textbf{Training} & \textbf{Reward} & & B-4 & M & C & RefCLIP-S & RefPAC-S & & CLIP-S & PAC-S & R@1 & MRR \\
    \midrule
    \ours (w/o quality distances) & CLIP-S & & 19.3 & 25.6 & 79.5 & 0.820 & 0.858 & &  0.836 & 0.851 & 45.7 & 57.2 \\
    \rowcolor{LightCyan}
    \textbf{\ours} & CLIP-S & & \textbf{21.4} & \textbf{27.1} & \textbf{82.6} & \textbf{0.824} & \textbf{0.863} & & \textbf{0.837} & \textbf{0.856} & \textbf{46.5} & \textbf{58.4} \\
    \midrule
    \ours (w/o quality distances) & PAC-S & & 24.9 & 27.6 & \textbf{91.7} & 0.812 & 0.873 & & 0.809 & 0.875 & 50.6 & 62.4 \\
    \rowcolor{LightCyan}
    \textbf{\ours} & PAC-S & & \textbf{25.2} & \textbf{28.4} & 89.1 & \textbf{0.815} & \textbf{0.875} & & \textbf{0.812} & \textbf{0.877} & \textbf{50.9} & \textbf{62.9} \\
    \bottomrule
  \end{tabular} 
}
\vspace{0.2cm}
  \caption{Effectiveness analysis of using quality distances to weight rewards. Results are reported on the COCO test set using ViT-L/14 as backbone.}
  \label{tab:ablation}
  \vspace{-0.45cm}
\end{table}

\tit{Effectiveness of using quality distances}
We also evaluate the effectiveness of weighting rewards with quality distances (cf. Eq.~\ref{eq:differences}) and train a different version of our \ours approach setting $\gamma_i=\frac{1}{k}$. Table~\ref{tab:ablation} reports the results of this analysis, using both CLIP-S and PAC-S as rewards. Notably, using quality distances to weight rewards improves the performance on both reference-based and reference-free metrics, thus demonstrating the usefulness of our strategy.

\begin{wrapfigure}{r}{0.6\textwidth}
\vspace{-0.2cm}
\centering
\includegraphics[width=\linewidth]{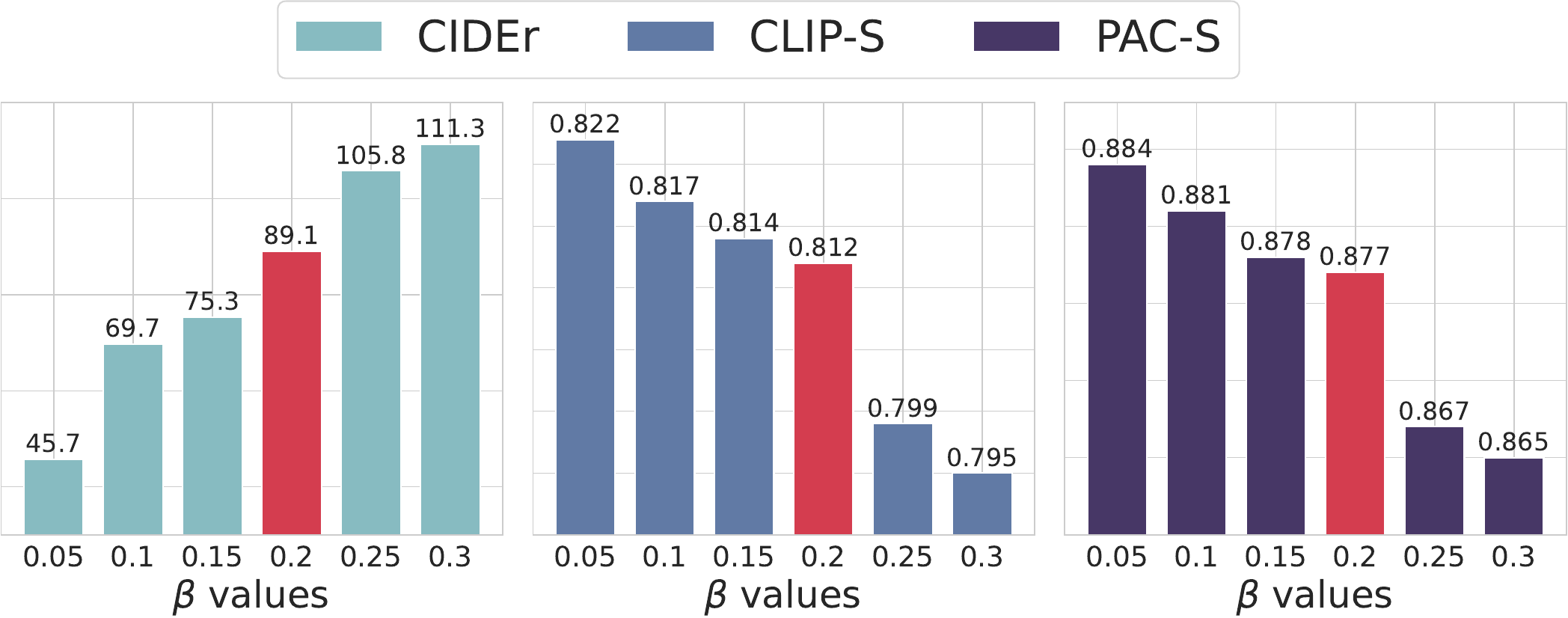}
\vspace{-0.5cm}
\caption{CIDEr, CLIP-S, and PAC-S scores when changing the $\beta$ parameter using ViT-L/14 as backbone. Higher $\beta$ values prevent the model from deviating from the pre-trained captioner, while penalizing reference-free metrics. The best trade-off is given by $\beta=0.2$.}
\label{fig:graphs}
\vspace{-0.3cm}
\end{wrapfigure}

\tit{Effect of varying the $\beta$ parameter} Fig.~\ref{fig:graphs} shows how evaluation metrics vary when changing the $\beta$ parameter, which regularizes the deviation from the pre-trained model. In particular, we report CIDEr, CLIP-S, and PAC-S scores using six different $\beta$ values (\ie~from 0.05 
to 0.3). As it can be seen, a higher $\beta$ value prevents the model from deviating from the original pre-trained captioner (trained with XE loss), with CIDEr scores greater than 100 and, as a consequence, lower CLIP-S and PAC-S. On the contrary, when using a lower $\beta$ value, reference-based metrics like CIDEr are penalized as the model is more inclined to deviate from the original version, thus boosting CLIP-S and PAC-S metrics. Overall, we find that $\beta=0.2$ represents a good compromise between reference-based and reference-free metrics, and therefore we employ this value for all experiments.

\tit{Number of loser captions}
\ours requires generating $k + 1$ captions at each training step, of which the $k$ worst are selected as losers according to the metric employed as reward. Table~\ref{tab:ablation_k} shows the results as we vary the parameter $k$. In our experiments, we select $k = 4$ as it achieves the highest scores on reference-free metrics while keeping competitive performance on reference-based metrics.

\begin{table}[t]
  \centering
  \setlength{\tabcolsep}{.3em}
  \resizebox{0.9\linewidth}{!}{
  \begin{tabular}{lcc c ccccc c cccc}
    \toprule
    & & & & \multicolumn{5}{c}{\textbf{Reference-based}} & & \multicolumn{4}{c}{\textbf{Reference-free}} \\
    \cmidrule{5-9} \cmidrule{11-14}
    \textbf{Training} & \textbf{Reward} & $k$ & & B-4 & M & C & RefCLIP-S & RefPAC-S & & CLIP-S & PAC-S & R@1 & MRR \\
    \midrule
    SCST & PAC-S & - & & 22.3 & 28.4 & 51.1 & 0.801 & 0.861 & & 0.805 & 0.862 & 46.7 & 58.8 \\
    \midrule
    \textbf{\ours (Ours)} & PAC-S & 1 & & 30.4 & 28.2 & 109.5 & 0.819 & 0.876 & & 0.790 & 0.861 & 41.4 & 54.0 \\
    \textbf{\ours (Ours)} & PAC-S & 2 & & 27.1 & 28.3 & 99.9 & 0.818 & 0.876 & & 0.802 & 0.871 & 46.2 & 58.9 \\
    \rowcolor{LightCyan}
    \textbf{\ours (Ours)} & PAC-S & 4 & & 25.2 & 28.4 & 89.1 & 0.815 & 0.875 & & \textbf{0.812} & \textbf{0.877} & \textbf{50.9} & \textbf{62.9} \\
    \textbf{\ours (Ours)} & PAC-S & 6 & & 24.9 & 28.6 & 86.9 & 0.813 & 0.874 & & 0.811 & 0.876 & \textbf{50.9} & 62.5 \\
    \textbf{\ours (Ours)} & PAC-S & 7 & & 24.8 & 28.5 & 86.5 & 0.812 & 0.873 & & 0.811 & 0.876 & 50.5 & 62.4 \\
    \bottomrule
  \end{tabular}
}
    \vspace{0.2cm}
  \caption{Performance varying the number of ``loser'' captions $k$. Results are reported on the COCO test set using ViT-L/14 as backbone.}
  \label{tab:ablation_k}
  \vspace{-0.5cm}
\end{table}

\begin{table}[b]
\vspace{-0.5cm}
  \centering
  \setlength{\tabcolsep}{.3em}
  \resizebox{0.8\linewidth}{!}{
  \begin{tabular}{lc c ccccc c cccc}
    \toprule
    & & & \multicolumn{5}{c}{\textbf{Reference-based}} & & \multicolumn{4}{c}{\textbf{Reference-free}} \\
    \cmidrule{4-8} \cmidrule{10-13}
    \textbf{Training} & \textbf{Reward} & & B-4 & M & C & RefCLIP-S & RefPAC-S & & CLIP-S & PAC-S & R@1 & MRR \\
    \midrule
    XE & - & & 37.3 & 30.4 & 126.6 & 0.811 & 0.868 & & 0.758 & 0.831 & 27.7 & 38.5 \\
    RLHF & HF & & 21.4 & 27.8 & 57.9 & 0.776 & 0.843 & & 0.745 & 0.819 & 24.7 & 34.9 \\
    \midrule
    RLHF & CLIP-S & & 12.9 & 24.2 & 2.3 & 0.714 & 0.800 & & 0.732 & 0.794 & 19.5 & 29.0 \\
    SCST & CLIP-S & & 10.2 & 23.0 & 1.1 & 0.793 & 0.827 & & \textbf{0.865} & 0.834 & 43.3 & 55.0 \\
    \rowcolor{LightCyan}
    \textbf{\ours} & CLIP-S & & \textbf{21.4} & \textbf{27.1} & \textbf{82.6} & \textbf{0.824} & \textbf{0.863} & & 0.837 & \textbf{0.856} & \textbf{46.5} & \textbf{58.4} \\
    \midrule
    RLHF & PAC-S & & 12.4 & 23.7 & 2.0 & 0.712 & 0.798 & & 0.726 & 0.790 & 18.1 & 27.5 \\
    SCST & PAC-S & & 22.3 & \textbf{28.4} & 51.1 & 0.801 & 0.861 & & 0.805 & 0.862 & 46.7 & 58.8 \\
    \rowcolor{LightCyan}
    \textbf{\ours} & PAC-S & & \textbf{25.2} & \textbf{28.4} & 89.1 & \textbf{0.815} & \textbf{0.875} & & \textbf{0.812} & \textbf{0.877} & \textbf{50.9} & \textbf{62.9} \\
    \bottomrule
  \end{tabular} 
}
\vspace{0.2cm}
\caption{Comparison with different fine-tuning strategies. Results are reported on the COCO test set using ViT-L/14 as backbone.}
\label{tab:strategies}
\end{table}

\section{Additional Experimental Results}
\label{sec:rlhf_comparison}
\tinytit{Comparison with SCST and RLHF} To complement the analyses reported in the main paper, we compare our fine-tuning strategy with SCST~\cite{rennie2017self} and RLHF~\cite{christiano2017deep}. As we focus on the optimization of modern captioning metrics, for SCST experiments we directly apply a CLIP-based reward using either CLIP-S or PAC-S. Further, we adapt the RLHF paradigm to a captioning setting by first training a reward model based on human feedback and then optimizing the captioning model via reinforcement learning based on the PPO objective~\cite{schulman2017proximal}, using the score from the reward model as a reward. To train the reward model, we employ a combination of datasets typically used to evaluate the correlation of captioning metrics with human judgments, namely Flickr8k-Experts, Flickr8k-CF, and Composite~\cite{hodosh2013framing,aditya2015images}. All datasets contain multiple candidate captions, either human-annotated or generated by a captioning model, associated with a given image and corresponding human ratings that evaluate whether each caption correctly describes the image. Overall, we obtain around 3.5k unique images and 50k captions each associated with a normalized rating between 0 and 1. At training time, we sample a pair of candidate captions for each image and use the associated human ratings to train the reward model, using maximum likelihood estimation. The reward model is built by modifying the captioner pre-trained with XE so to have a single final output and is trained with a negative log-likelihood loss following~\cite{ouyang2022training}. In addition to this adaptation of the RLHF training strategy, we also design a variant in which the human preferences-based reward model is replaced with a CLIP-based evaluator, directly employing CLIP-S or PAC-S as reward. For completeness, we also include the results of the model trained with XE loss only, which is the starting point for all other fine-tuning strategies.

Results are reported in Table~\ref{tab:strategies} in terms of reference-based and reference-free evaluation metrics. As it can be seen, the proposed optimization strategy generally leads to better results across all metrics, surpassing both SCST and RLHF by a significant margin. Specifically, we can notice that optimizing the captioner with human feedback does not improve the final results. This is probably due to the limited size of available captioning datasets with human ratings, that prevent the effective application of standard RLHF fine-tuning to a captioning model. 
When instead using CLIP-S and PAC-S as rewards, both SCST and RLHF experience a significant drop in standard image captioning metrics. In terms of CLIP-based metrics, SCST obtains quite good results which however are not supported with robustness on all other metrics. Overall, our \ours strategy exhibits good performance in all evaluation directions, obtaining the best results in terms of CLIP-based and retrieval-based scores while maintaining competitive performance on standard metrics.

\begin{table*}[t]
  \centering
  \setlength{\tabcolsep}{.25em}
  \resizebox{0.9\linewidth}{!}{
  \begin{tabular}{lcc c ccccc c cc}
    \toprule
    & & & & \multicolumn{5}{c}{\textbf{Reference-based}} & & \multicolumn{2}{c}{\textbf{Reference-free}} \\
    \cmidrule{5-9} \cmidrule{11-12}
    \textbf{Model} & \textbf{Backbone} & \textbf{Reward} & & B-4 & M & C & RefCLIP-S & RefPAC-S & & CLIP-S & PAC-S \\
    \midrule 
    Cho~\etal~(SCST)~\cite{cho2022fine} & RN50 & CLIP-S & & 5.9 & 14.2 & 13.9 & 0.689 & 0.769 & & \textbf{0.721} & 0.803 \\
    Cho~\etal~(SCST)~\cite{cho2022fine} & RN50 & CLIP-S+Gr. & & 11.1 & 16.3 & 19.1 & 0.683 & 0.784 & & 0.684 & 0.808 \\
    \rowcolor{LightCyan}
    \textbf{\ours (Ours)} & RN50 & CLIP-S & & \textbf{14.2} & 16.1 & 17.2 & 0.688 & 0.782 & & 0.696 & 0.805 \\
    \rowcolor{LightCyan}
    \textbf{\ours (Ours)} & RN50 & PAC-S & & 13.6 & \textbf{16.4} & \textbf{19.2} & \textbf{0.695} & \textbf{0.800} & & 0.704 & \textbf{0.835} \\
    \midrule
    \rowcolor{LightCyan}
    \textbf{\ours (Ours)} & ViT-B/32 & PAC-S & & 13.8 & 16.7 & 19.3	& 0.708 & 0.811 & & 0.726 & 0.855 \\
    \rowcolor{LightCyan}
    \textbf{\ours (Ours)} & ViT-L/14 & PAC-S & & 15.0 & 17.4 & 23.2 & 0.722 & 0.822 & & 0.731 & 0.855 \\
    \bottomrule
  \end{tabular}
}
   \vspace{0.2cm}
  \caption{Fine-grained image captioning results on the FineCapEval dataset.}
  \label{tab:finecapeval}
  \vspace{-0.15cm}
\end{table*}

\tit{Fine-grained image captioning evaluation} As an additional analysis, we report in Table~\ref{tab:finecapeval} fine-grained image captioning results on the FineCapEval dataset~\cite{cho2022fine}, which contains 1,000 images from COCO and CC3M~\cite{sharma2018conceptual} annotated with 5 detailed and fine-grained captions, describing the background of the scene, the objects and their attributes, and the relations between them. Also in this setting, \ours confirms its superior performance compared to other CLIP-based optimized captioners~\cite{cho2022fine}, thus further demonstrating the effectiveness of directly optimizing a captioning model with the proposed solution. Specifically, when considering the same backbone used in~\cite{cho2022fine} (\ie~RN50), \ours achieves the best results in terms of both standard captioning metrics and CLIP-based scores.

\begin{table*}[t]
  \centering
  \setlength{\tabcolsep}{.18em}
  \resizebox{\linewidth}{!}{
  \begin{tabular}{lcc c ccc c ccc c ccc c ccc}
    \toprule
    & & & & \multicolumn{3}{c}{\textbf{nocaps}} & & \multicolumn{3}{c}{\textbf{VizWiz}} & & \multicolumn{3}{c}{\textbf{TextCaps}} & & \multicolumn{3}{c}{\textbf{CC3M}} \\ 
    \cmidrule{5-7} \cmidrule{9-11} \cmidrule{13-15} \cmidrule{17-19}
    \textbf{Model} & \textbf{Reward} & \textbf{Backbone} & & C & CLIP-S & PAC-S & & C & CLIP-S & PAC-S & & C & CLIP-S & PAC-S & & C & CLIP-S & PAC-S \\ 
    \midrule
    Cho~\etal~(SCST)~\cite{cho2022fine} & CLIP-S & RN50 & & 10.9 & \textbf{0.765} & 0.819 & & 4.7 & \textbf{0.693} & 0.784 & & 7.6& \textbf{0.731} & 0.813 & & 3.6 & \textbf{0.717} & 0.784 \\
    Cho~\etal~(SCST)~\cite{cho2022fine} & CLIP-S+Gr. & RN50 & & 54.0 & 0.712 & 0.822 & & 20.4 & 0.648 & 0.774 & & 26.8 & 0.680 & 0.814 & & 18.0 & 0.671 & 0.790 \\
    SCST & PAC-S & RN50 & & 20.9 & 0.741 & 0.850 & & 13.0 & 0.668 & 0.795 & & 22.0 & 0.683 & 0.822 & & 5.8 & 0.699 & 0.797 \\
    \rowcolor{LightCyan}
    \textbf{\ours (Ours)} & PAC-S & RN50 & & \textbf{64.6} & 0.733 & \textbf{0.851} & & \textbf{29.3} & 0.680 & \textbf{0.813} & & \textbf{30.8} & 0.696 & \textbf{0.838} & & \textbf{21.4} & 0.690 & \textbf{0.815} \\
    \midrule
    SCST & PAC-S & ViT-B/32 & & 35.7 & 0.750 & 0.854 & & 20.1 & \textbf{0.715} & 0.837 & & 21.9 & 0.699 & 0.835 & & 9.8 & \textbf{0.698} & 0.809 \\ 
    \rowcolor{LightCyan}
    \textbf{\ours (Ours)} & PAC-S & ViT-B/32 & & \textbf{66.5} & \textbf{0.754} & \textbf{0.869} & & \textbf{32.7} & 0.710 & \textbf{0.842} & & \textbf{31.8} & \textbf{0.712} & \textbf{0.853} & & \textbf{23.4} & 0.697 & \textbf{0.821} \\
    \midrule
    SCST & PAC-S & ViT-L/14 & & 44.8 & 0.746 & 0.850 & & 26.8 & 0.701 & 0.820 & & 23.6 & 0.705 & 0.836 & & 13.2 & 0.701 & 0.811 \\
    \rowcolor{LightCyan}
    \textbf{\ours (Ours)} & PAC-S & ViT-L/14  & & \textbf{74.3} & \textbf{0.755} & \textbf{0.865}  & & \textbf{40.6} & \textbf{0.706} & \textbf{0.832}  & & \textbf{33.7} & \textbf{0.717} & \textbf{0.852} & &  \textbf{26.7} & \textbf{0.704} & \textbf{0.824} \\
    \bottomrule
  \end{tabular}
}
\vspace{-0.15cm}
  \caption{Image captioning results on out-of-domain datasets like nocaps, VizWiz, TextCaps, and CC3M.}
  \label{tab:other_datasets}
  \vspace{-0.45cm}
\end{table*}

\tit{Out-of-domain evaluation} To evaluate generalization capabilities to out-of-domain images, we extend our analysis by considering diverse image captioning datasets, including nocaps~\cite{agrawal2019nocaps}, which has been introduced for novel object captioning and contains object classes that are not present in the COCO dataset, VizWiz~\cite{gurari2020captioning}, composed of images taken by blind people, TextCaps~\cite{sidorov2020textcaps}, which is instead focused on text-rich images, and CC3M~\cite{sharma2018conceptual}, composed of image-caption pairs collected from the web. In Table~\ref{tab:other_datasets} we report the results of our approach using PAC-S as reward, compared to the CLIP-based training strategy proposed in~\cite{cho2022fine} and the standard SCST with the same reward as in our approach. Even in this challenging context, \ours achieves the best results across all datasets and backbones, demonstrating better descriptive capabilities than competitors.

\tit{Additional results on Flickr30k} 
Finally, we also benchmark our method on images from the Flickr30k dataset~\cite{young2014image}. We report the results in Table~\ref{tab:flickr}, using both PAC-S and CLIP-S as reward and comparing with the approach proposed in~\cite{cho2022fine}. As it can be noticed, \ours demonstrates strong generalization capabilities, achieving the best results on almost all evaluation metrics further confirming the effectiveness of our training strategy.

\begin{table*}[t]
  \centering
  \setlength{\tabcolsep}{.25em}
  \resizebox{0.88\linewidth}{!}{
  \begin{tabular}{lcc c ccccc c cc}
    \toprule
    & & & & \multicolumn{5}{c}{\textbf{Reference-based}} & & \multicolumn{2}{c}{\textbf{Reference-free}} \\
    \cmidrule{5-9} \cmidrule{11-12}
    \textbf{Model} & \textbf{Backbone} & \textbf{Reward} & & B-4 & M & C & RefCLIP-S & RefPAC-S & & CLIP-S & PAC-S \\
    \midrule 
    Cho~\etal~(SCST)~\cite{cho2022fine} & RN50 & CLIP-S & & 4.0 & 16.5 & 10.0 & 0.751 & 0.806 & & \textbf{0.818} & 0.839 \\
    Cho~\etal~(SCST)~\cite{cho2022fine} & RN50 & CLIP-S+Gr. & & 11.0 & 20.9 & 36.8 & 0.750 & 0.826 & & 0.755 & 0.839 \\
    \rowcolor{LightCyan}
    \textbf{\ours (Ours)} & RN50 & CLIP-S & & 16.8 & 22.0 & 44.9 & 0.762 & 0.829 & & 0.774 & 0.839 \\
    \rowcolor{LightCyan}
    \textbf{\ours (Ours)} & RN50 & PAC-S & & \textbf{17.2} & \textbf{22.6} & \textbf{46.8} & \textbf{0.769} & \textbf{0.846} & & 0.786 & \textbf{0.871} \\
    \midrule
    \rowcolor{LightCyan}
    \textbf{\ours (Ours)} & ViT-B/32 & PAC-S & & 17.8 & 22.8 & 48.6 & 0.780 & 0.855 & & 0.810 & 0.890 \\
    \rowcolor{LightCyan}
    \textbf{\ours (Ours)} & ViT-L/14 & PAC-S & & 19.0 & 24.5 & 55.8 & 0.790 & 0.862 & & 0.804 & 0.883 \\
    \bottomrule
  \end{tabular}
}
\vspace{0.2cm}
\caption{Image captioning results on the Flickr30k dataset.}
\label{tab:flickr}
\vspace{-0.45cm}
\end{table*}

\section{Additional Details} 
\label{sec:details}
\tinytit{Additional implementation and training details} During cross-entropy pre-training, we accumulate gradients for 8 training steps over 2 GPUs, resulting in 1,024 samples per batch. For this training stage, the learning rate is linearly increased up to $2.5 \cdot 10^{-4}$. Each fine-tuning experiment starts from the XE checkpoint with the highest CIDEr, leveraging 2 GPUs and a global batch size of 16. Training the reward models for RLHF follows the same settings as the fine-tuning phase.

\tit{CIDEr-based optimization}
In computing quality distances with
CIDEr metric as reward
\begin{wrapfigure}{r}{0.55\textwidth}
\vspace{-0.1cm}
  \centering
  \setlength{\tabcolsep}{.18em}
  \resizebox{\linewidth}{!}{
\includegraphics[width=0.8\linewidth]{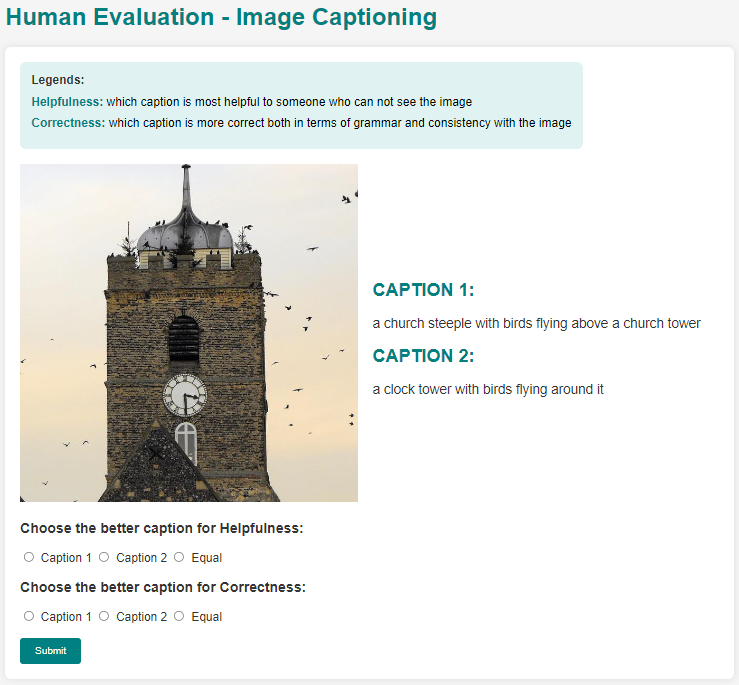}
}
\vspace{-0.6cm}
\caption{User study interface to evaluate helpfulness and correctness of given captions.}
\label{fig:user_study}
\vspace{-0.5cm}
\end{wrapfigure}
(see Table~\ref{tab:cider} of the main paper), we set the softmax temperature $\tau$ to 1, a higher value than the one used for CLIP-S and PAC-S optimization (equal to $1/(3 \cdot 10^2)$). We argue that the CIDEr score is discriminant enough to discern the goodness of similar captions sampled from a beam search. On the contrary, 
CLIP-based metrics are less sensible to small changes, thus needing a lower temperature to amplify the score differences.

\tit{Human-based evaluation}
As shown in the main paper, we conducted a user study to evaluate the quality of generated captions. 
To this end, we developed a custom web interface that presents the users with an image and two captions, one generated by \ours, and one drawn from a different model (cf. Table~2), and asks them to select the best caption based on correctness and helpfulness. We show a screenshot of the developed interface is shown in Fig.~\ref{fig:user_study}. Overall, the evaluation involved more than 50 different users, collecting approximately 3,000 evaluations for both criteria.

\tit{LLM-based evaluation} 
GPT-3.5 Turbo proves itself very compliant with our requests. However, we find about a hundred failure cases (\eg~wrong JSON format, more scores than the number of candidate captions, etc.) out of 7,000 requests. We opt for simply discarding them in the winner rate computations. For fair evaluation, we randomly swap the order in which we insert the two candidate captions in the prompt. This ensures that the descriptions generated by our competitors have on average the same probability as ours to be processed first by the LLM causal attention, which may influence the final score. Following~\cite{chan2023clair}, the prompt we used is:

\vspace{0.3cm}
\resizebox{0.96\linewidth}{!}{
\begin{minipage}{\linewidth}
\footnotesize
\noindent\texttt{You are trying to tell if each sentence in a candidate set of captions is describing the same image as a reference set of captions.}

\noindent\texttt{Candidate set: \{candidate captions\}}

\noindent\texttt{Reference set: \{target captions\}}

\noindent\texttt{You have to determine how likely is that each of the sentences in the candidate set is describing the same image as the reference set, on a scale from 0 to 100. Please output exclusively a JSON list, with an item for each candidate. Each item should be a dictionary with a key ``score'', containing a value between 0 and 100, and a key ``reason'' with a string value containing the justification of the rating. Start directly with the json.}
\end{minipage}
}

\section{Additional Qualitative Results}
\label{sec:qualitatives}
\begin{figure*}[t]
\centering
\begin{minipage}{0.22\linewidth}
    \includegraphics[width=0.98\linewidth]{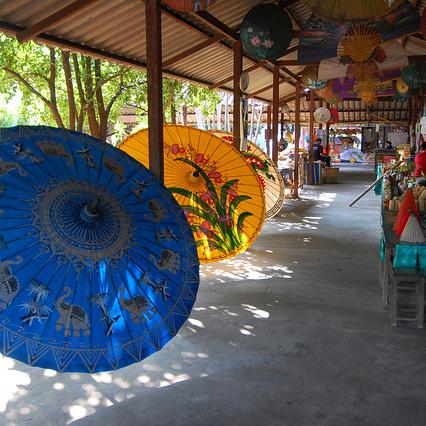}
    \end{minipage}
\begin{minipage}{0.77\linewidth}
    \scriptsize{
    \textbf{BLIP-2~\cite{li2023blip}:} A group of colorful umbrellas under a covered area.\\
    \textbf{Cho~\etal~\cite{cho2022fine}:} A large blue vase sitting on the dirt ground with colorful decorations next to a market.\\
    \textbf{SCST:} Several colorful colorful umbrellas hanging from a wooden structure under a tree tree with statues on display in outdoor market under palm trees on clear background.\\
    \textbf{\ours (Ours):} A display of colorful umbrellas in a shop with decorations.
    }
\end{minipage}
\vspace{0.05cm}

\begin{minipage}{0.22\linewidth}
    \includegraphics[width=0.98\linewidth]{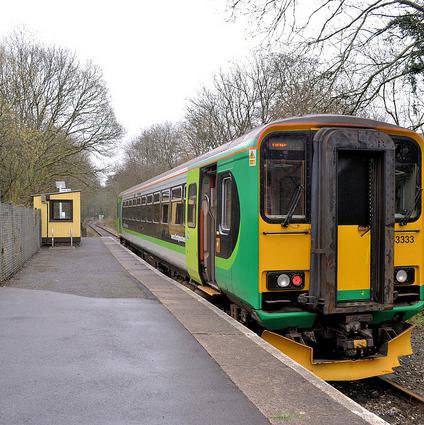}
\end{minipage}
\begin{minipage}{0.77\linewidth}
    \scriptsize{
    \textbf{BLIP-2~\cite{li2023blip}:} A green and yellow train pulling into a station.\\
    \textbf{Cho~\etal~\cite{cho2022fine}:} A green commuter train parked near a platform area with a green trees area motion stance ear stance.\\
    \textbf{SCST:} A green and yellow passenger train traveling down train tracks next to a loading platform with a green passenger on a platform with trees in background.\\
    \textbf{\ours (Ours):} A green and yellow passenger train traveling down train tracks next to a platform.
    }
\end{minipage}
\vspace{0.05cm}

\begin{minipage}{0.22\linewidth}
    \includegraphics[width=0.98\linewidth]{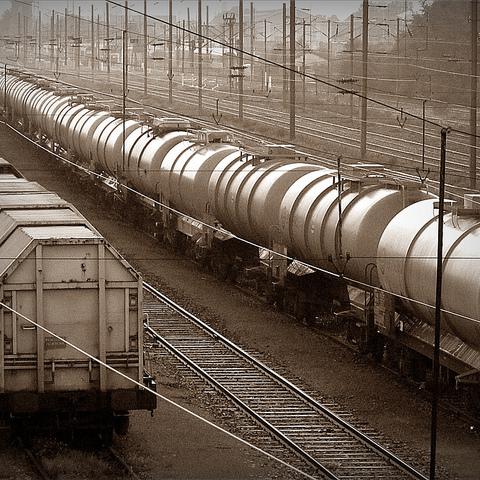}
    \end{minipage}
\begin{minipage}{0.77\linewidth}
    \scriptsize{
    \textbf{BLIP-2~\cite{li2023blip}:} A black and white photo of a train.\\
    \textbf{Cho~\etal~\cite{cho2022fine}:} A large metal train driving next to a lot of tanks on the tracks.\\
    \textbf{SCST:} Black and white photograph of freight freight freight freight cars on railroad tracks with tanker cars on track with wires in background on background.\\
    \textbf{\ours (Ours):} A black and white photo of a freight train traveling down railroad tracks next to wires.
    }
\end{minipage}
\vspace{0.05cm}

\begin{minipage}{0.22\linewidth}
    \includegraphics[width=0.98\linewidth]{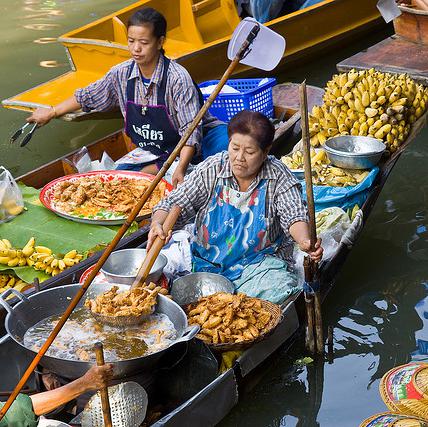}
    \end{minipage}
\begin{minipage}{0.77\linewidth}
    \scriptsize{
    \textbf{BLIP-2~\cite{li2023blip}:} A woman in a boat selling food on the water.\\
    \textbf{Cho~\etal~\cite{cho2022fine}:} A couple of women preparing a tray of food in the river with bananas.\\
    \textbf{SCST:} Two women in canoes with baskets full of bananas and other asian asian workers carrying baskets on shelves with baskets on clear surface in background.\\
    \textbf{\ours (Ours):} Two asian women in a small boat filled with food and bananas.
    }
\end{minipage} 
\vspace{0.05cm}

\begin{minipage}{0.22\linewidth}
    \includegraphics[width=0.98\linewidth]{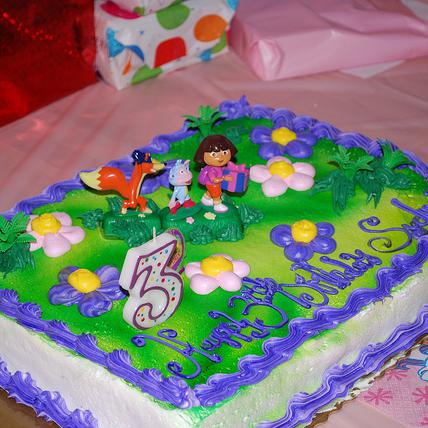}
    \end{minipage}
\begin{minipage}{0.77\linewidth}
    \scriptsize{
    \textbf{BLIP-2~\cite{li2023blip}:} A birthday cake with dora the explorer on it.\\
    \textbf{Cho~\etal~\cite{cho2022fine}:} A large blue birthday cake with toys and toys on the table.\\
    \textbf{SCST:} A colorful birthday cake decorated with purple and green flowers on top of purple birthday cake with decorations on table in background on background.\\
    \textbf{\ours (Ours):} A birthday cake with purple and green decorations on it.
    }
\end{minipage}
\vspace{0.05cm}

\begin{minipage}{0.22\linewidth}
    \includegraphics[width=0.98\linewidth]{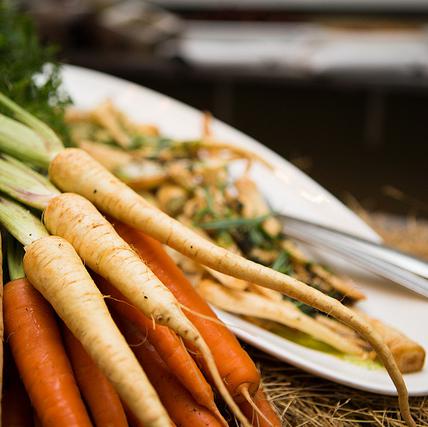}
    \end{minipage}
\begin{minipage}{0.77\linewidth}
    \scriptsize{
    \textbf{BLIP-2~\cite{li2023blip}:} A bunch of carrots next to a plate of food.\\
    \textbf{Cho~\etal~\cite{cho2022fine}:} A bunch of carrots and other carrots on a white plate with a knife behind them.\\
    \textbf{SCST:} A white plate topped with carrots and other vegetables on a clear surface with other vegetables on display in background on background in background.\\
    \textbf{\ours (Ours):} A bunch of carrots and other vegetables on a white plate.
    }
\end{minipage}
\vspace{-0.15cm}
\caption{Qualitative results on sample images from the COCO Karpathy test split using \ours with PAC-S reward. We compare our approach with SCST using PAC-S as reward, the model proposed in~\cite{cho2022fine} with CLIP+S+Grammar as reward, and the BLIP-2 model~\cite{li2023blip} which has been trained on large-scale vision-and-language datasets.}
\label{fig:coco_qualitatives_supp1}
\vspace{-.3cm}
\end{figure*}

\begin{figure*}
\begin{minipage}{0.22\linewidth}
    \includegraphics[width=0.98\linewidth]{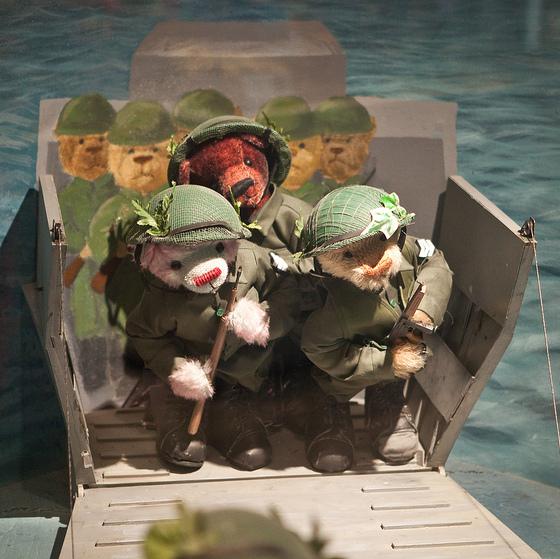}
\end{minipage}
\begin{minipage}{0.77\linewidth}
    \scriptsize{
    \textbf{BLIP-2~\cite{li2023blip}:} A group of teddy bears on a boat.\\
    \textbf{Cho~\etal~\cite{cho2022fine}:} A couple of teddy bears wearing hats sitting on a boat with a plant behind them.\\
    \textbf{SCST:} Teddy bears dressed in green costumes riding a miniature boat decorated with green hats on a blue wall in military uniform on display in background.\\
    \textbf{\ours (Ours):} Stuffed animals dressed in green costumes riding in a boat.
    }
\end{minipage}
\vspace{0.05cm}

\begin{minipage}{0.22\linewidth}
    \includegraphics[width=0.98\linewidth]{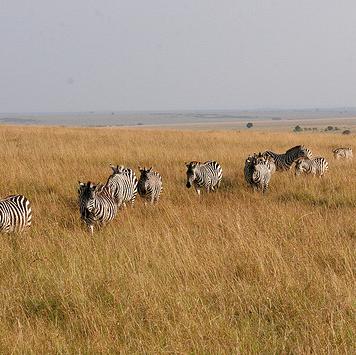}
    \end{minipage}
\begin{minipage}{0.77\linewidth}
    \scriptsize{
    \textbf{BLIP-2~\cite{li2023blip}:} A herd of zebras walking through a grassy field.\\
    \textbf{Cho~\etal~\cite{cho2022fine}:} A large herd of zebra and other animals grazing in the prairie.\\
    \textbf{SCST:} A herd of zebras running through tall brown grass in savanna with distance in distance in background on clear surface in background on background.\\
    \textbf{\ours (Ours):} A large herd of zebras walking through tall brown grass in a large field.
    }
\end{minipage} 
\vspace{0.05cm}

\begin{minipage}{0.22\linewidth}
    \includegraphics[width=0.98\linewidth]{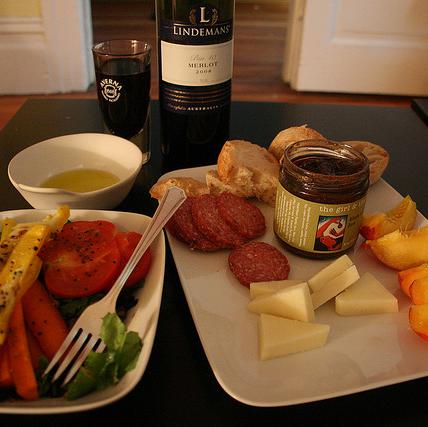}
   \end{minipage}
\begin{minipage}{0.77\linewidth}
    \scriptsize{
    \textbf{BLIP-2~\cite{li2023blip}:} A table topped with food and a bottle of wine.\\
    \textbf{Cho~\etal~\cite{cho2022fine}:} Two plates of food and a bottle of wine on the table with a bottle.\\
    \textbf{SCST:} A white plate topped with meat cheese and vegetables next to a bottle of wine and bread with cheese and tomatoes on wooden surface in background.\\
    \textbf{\ours (Ours):} A table topped with two bowls of food next to a bottle of wine and cheese.
    }
\end{minipage}
\vspace{0.05cm}

\begin{minipage}{0.22\linewidth}
    \includegraphics[width=0.98\linewidth]{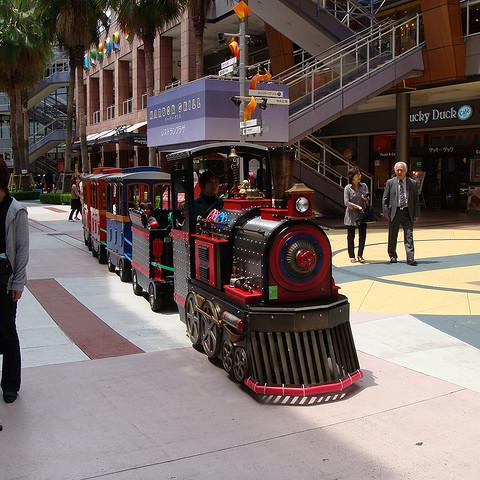}
    \end{minipage}
\begin{minipage}{0.77\linewidth}
    \scriptsize{
    \textbf{BLIP-2~\cite{li2023blip}:} A small train that is on display in a mall.\\
    \textbf{Cho~\etal~\cite{cho2022fine}:} A large red train driving on a busy street with people near it.\\
    \textbf{SCST:} A miniature miniature train with a miniature train on a sidewalk with people walking around a mall with a mall on a mall platform in background.\\
    \textbf{\ours (Ours):} People walking around a miniature train on a sidewalk in a shopping mall.
    }
\end{minipage} 
\vspace{0.05cm}

\begin{minipage}{0.22\linewidth}
    \includegraphics[width=0.98\linewidth]{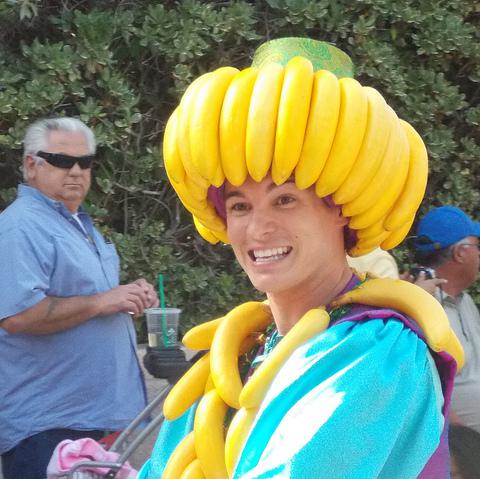}
    \end{minipage}
\begin{minipage}{0.77\linewidth}
    \scriptsize{
    \textbf{BLIP-2~\cite{li2023blip}:} A woman with a bunch of bananas on her head.\\
    \textbf{Cho~\etal~\cite{cho2022fine}:} A smiling woman wearing a colorful costume holding a bunch of bananas on the background.\\
    \textbf{SCST:} A woman dressed in colorful costume with yellow bananas on her head with a man's head dressed in colorful costume in background on background.\\
    \textbf{\ours (Ours):} A smiling woman wearing a large banana costume on her head with people in the background.
    }
\end{minipage}
\vspace{0.05cm}

\begin{minipage}{0.22\linewidth}
    \includegraphics[width=0.98\linewidth]{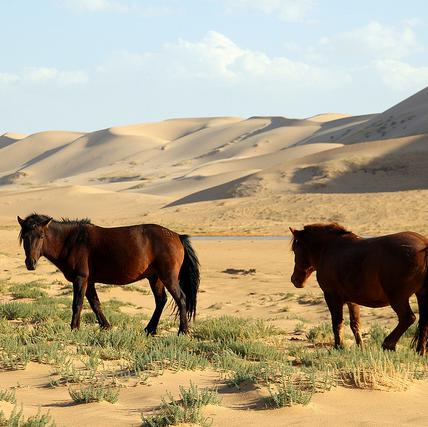}
    \end{minipage}
\begin{minipage}{0.77\linewidth}
    \scriptsize{
    \textbf{BLIP-2~\cite{li2023blip}:} Two horses walking in the desert with mountains in the background.\\
    \textbf{Cho~\etal~\cite{cho2022fine}:} A group of three brown horses walking together in the desert.\\
    \textbf{SCST:} Two brown horses walking through dry desert desert with sand on clear surface in distance with clear background on clear surface in background.\\
    \textbf{\ours (Ours):} Two brown horses walking through a desert plain with sand and bushes in background.
    }
\end{minipage} 
\vspace{-0.15cm}
\caption{Qualitative results on sample images from the COCO Karpathy test split using \ours with PAC-S reward. We compare our approach with SCST using PAC-S as reward, the model proposed in~\cite{cho2022fine} with CLIP+S+Grammar as reward, and the BLIP-2 model~\cite{li2023blip} which has been trained on large-scale vision-and-language datasets.}
\label{fig:coco_qualitatives_supp2}
\end{figure*}

\begin{figure}[t]
    \begin{minipage}{0.165\linewidth}
        \includegraphics[width=0.97\linewidth]{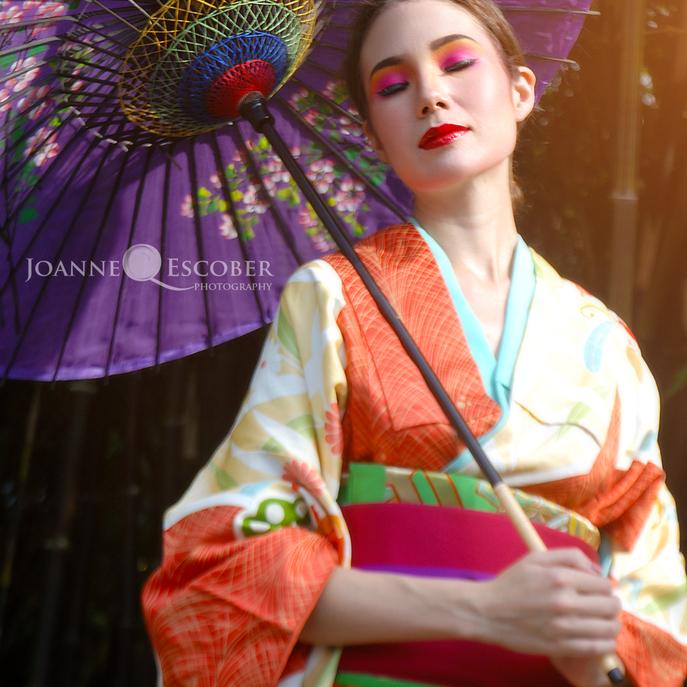}
        \end{minipage}
        \begin{minipage}{0.32\linewidth}
        \scriptsize{
        \textbf{SCST:} A woman dressed in colorful costume holding a colorful umbrella in a rainoutfit with chinese writing on her face.\\
        \textbf{DiCO (Ours):} A asian woman wearing a colorful costume holding a parasol.
        }
    \end{minipage}
    \hspace{0.02cm}
    \begin{minipage}{0.165\linewidth}
        \includegraphics[width=0.97\linewidth]{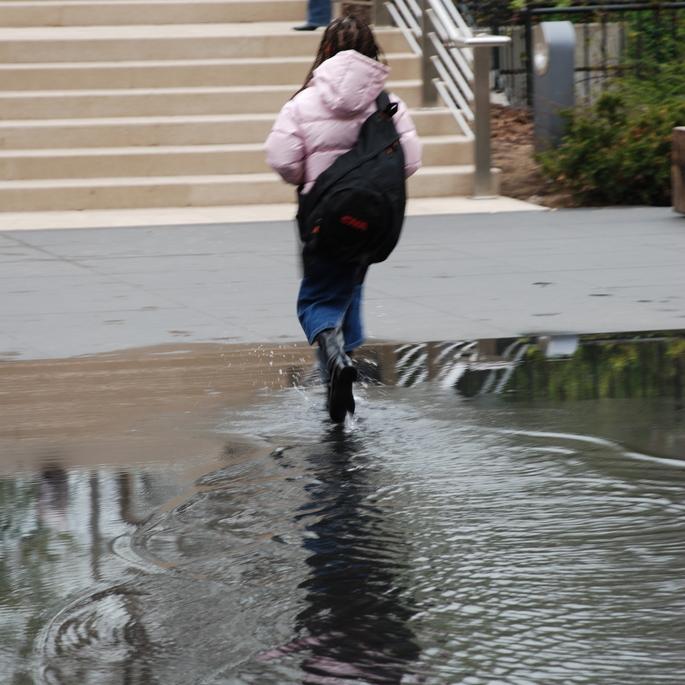}
        \end{minipage}
        \begin{minipage}{0.32\linewidth}
        \scriptsize{
        \textbf{SCST:} A little girl walking through water with a backpack walking through a flooded street with water in a girl's hand. \\
        \textbf{DiCO (Ours):} A girl walking through a flooded street with a backpack.
        }
    \end{minipage}
    
    \vspace{0.3cm}

    \begin{minipage}{0.165\linewidth}
        \includegraphics[width=0.97\linewidth]{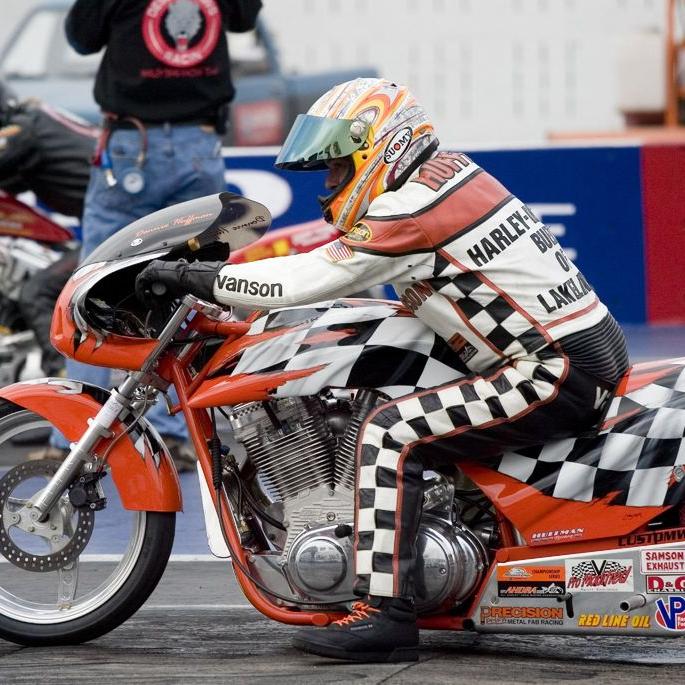}
        \end{minipage}
        \begin{minipage}{0.32\linewidth}
        \scriptsize{
        \textbf{SCST:} A person sitting on a red motorcycle with a helmet on a red motorcycle at a show.\\
        \textbf{DiCO (Ours):} A person in a red and white outfit sitting on a red motorcycle at a show.
        }
    \end{minipage}
    \hspace{0.02cm}
    \begin{minipage}{0.165\linewidth}
        \includegraphics[width=0.97\linewidth]{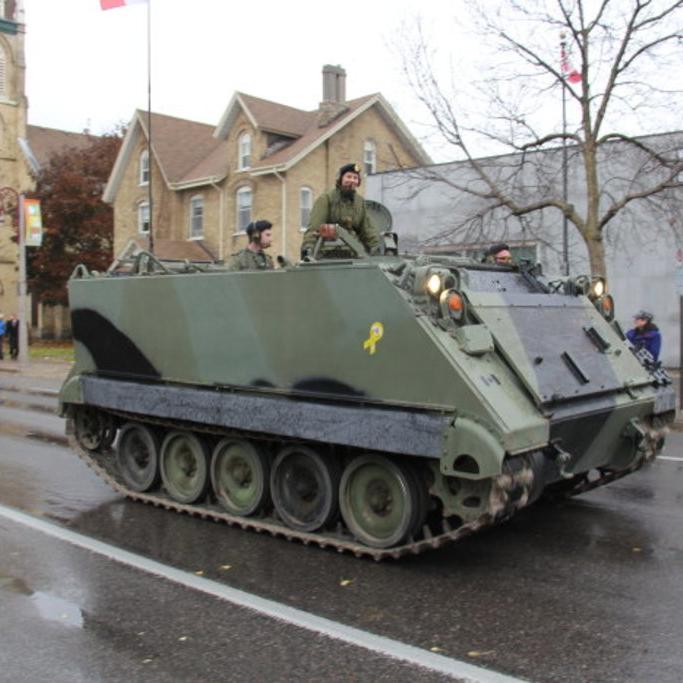}
        \end{minipage}
        \begin{minipage}{0.32\linewidth}
        \scriptsize{
        \textbf{SCST:} A military military vehicle driving down a rain soaked road with people in a military military vehicle on a rainy day. \\
        \textbf{DiCO (Ours):} A military vehicle driving down a wet road with people standing on it.
        }
    \end{minipage}
    
    \vspace{0.3cm}

        \begin{minipage}{0.165\linewidth}
        \includegraphics[width=0.97\linewidth]{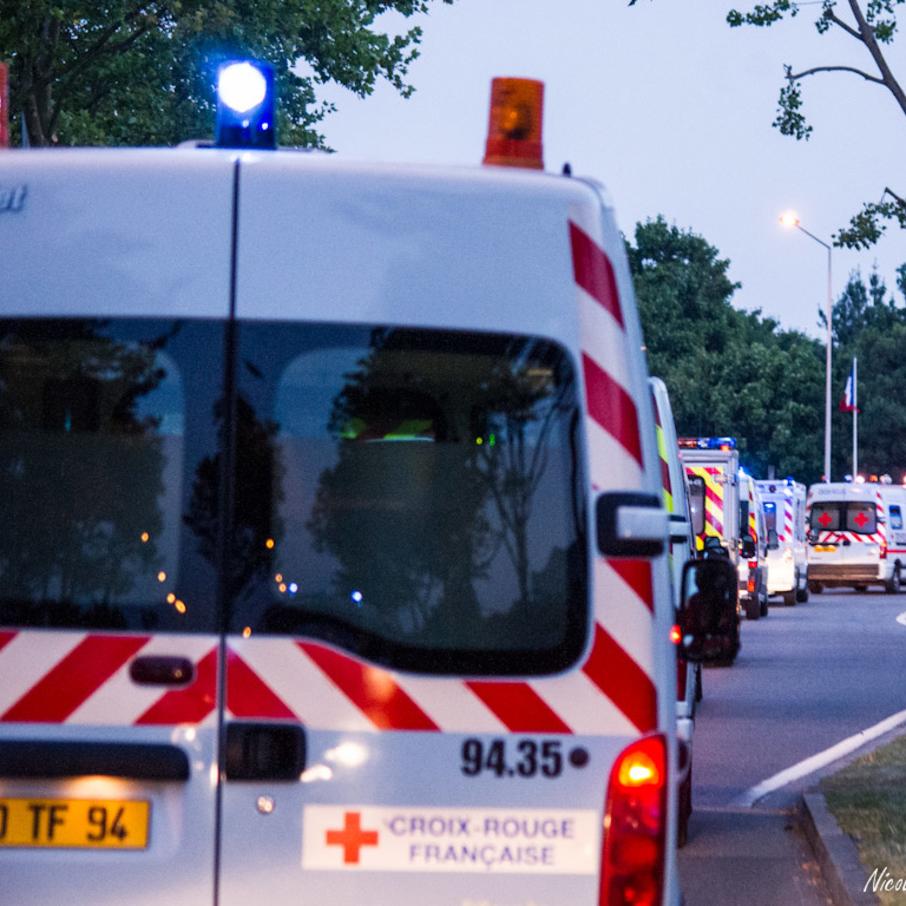}
        \end{minipage}
        \begin{minipage}{0.32\linewidth}
        \scriptsize{
        \textbf{SCST:} A police car driving down a road with lights on driving down a busy road with other vehicles on a sunny day.\\
        \textbf{DiCO (Ours):} A line of emergency vehicles driving down a road with trees in the background.
        }
    \end{minipage}
    \hspace{0.02cm}
    \begin{minipage}{0.165\linewidth}
        \includegraphics[width=0.97\linewidth]{images/supplementary/out_of_domain/nocaps/cropped_7562a43f192ef66c.jpg}
        \end{minipage}
        \begin{minipage}{0.32\linewidth}
        \scriptsize{
        \textbf{SCST:} A chocolate cake topped with strawberries and strawberries on a plate with ice cream with strawberries on a white surface. \\
        \textbf{DiCO (Ours):} A cake topped with strawberries and whipped cream.
        }
    \end{minipage}
    \vspace{-0.15cm}
    \caption{Qualitative results on sample images from nocaps.}
    \label{fig:nocaps}
\end{figure}

\begin{figure}[t]
    \begin{minipage}{0.165\linewidth}
        \includegraphics[width=0.97\linewidth]{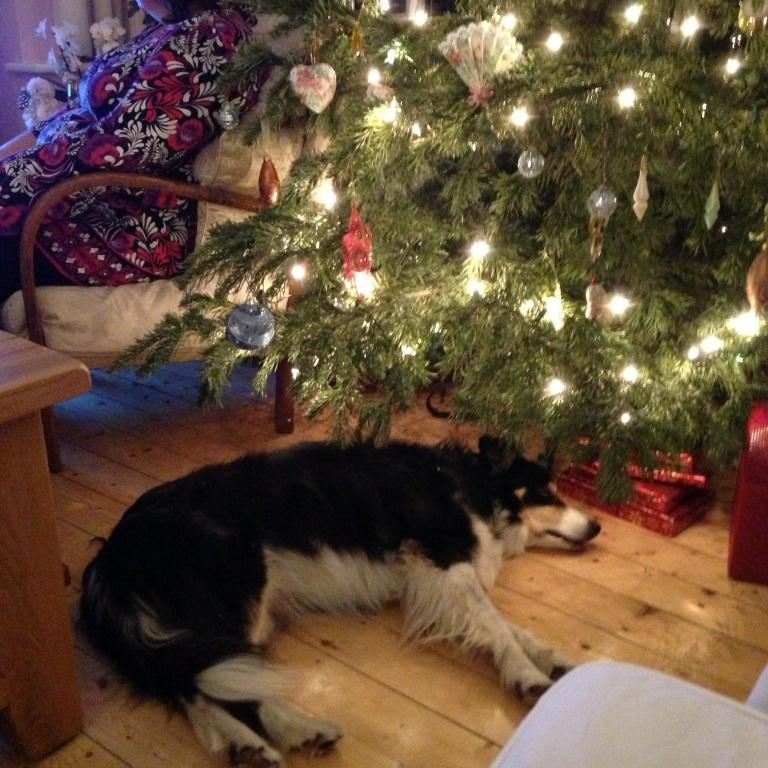}
        \end{minipage}
        \begin{minipage}{0.32\linewidth}
        \scriptsize{
        \textbf{SCST:} A black and white dog laying in front of a Christmas tree with a dog laying on the floor next to it.\\
        \textbf{DiCO (Ours):} A black and white dog laying in front of a christmas tree.
        }
    \end{minipage}
    \hspace{0.02cm}
    \begin{minipage}{0.165\linewidth}
        \includegraphics[width=0.97\linewidth]{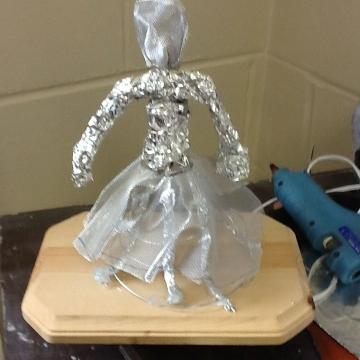}
        \end{minipage}
        \begin{minipage}{0.32\linewidth}
        \scriptsize{
        \textbf{SCST:} A plastic statue of a person wearing aluminum foil on a wooden board with aluminum foil on a counter. \\
        \textbf{DiCO (Ours):} A sculpture of a person wearing a dress standing on a wooden board.
        }
    \end{minipage}
    
    \vspace{0.3cm}

    \begin{minipage}{0.165\linewidth}
        \includegraphics[width=0.97\linewidth]{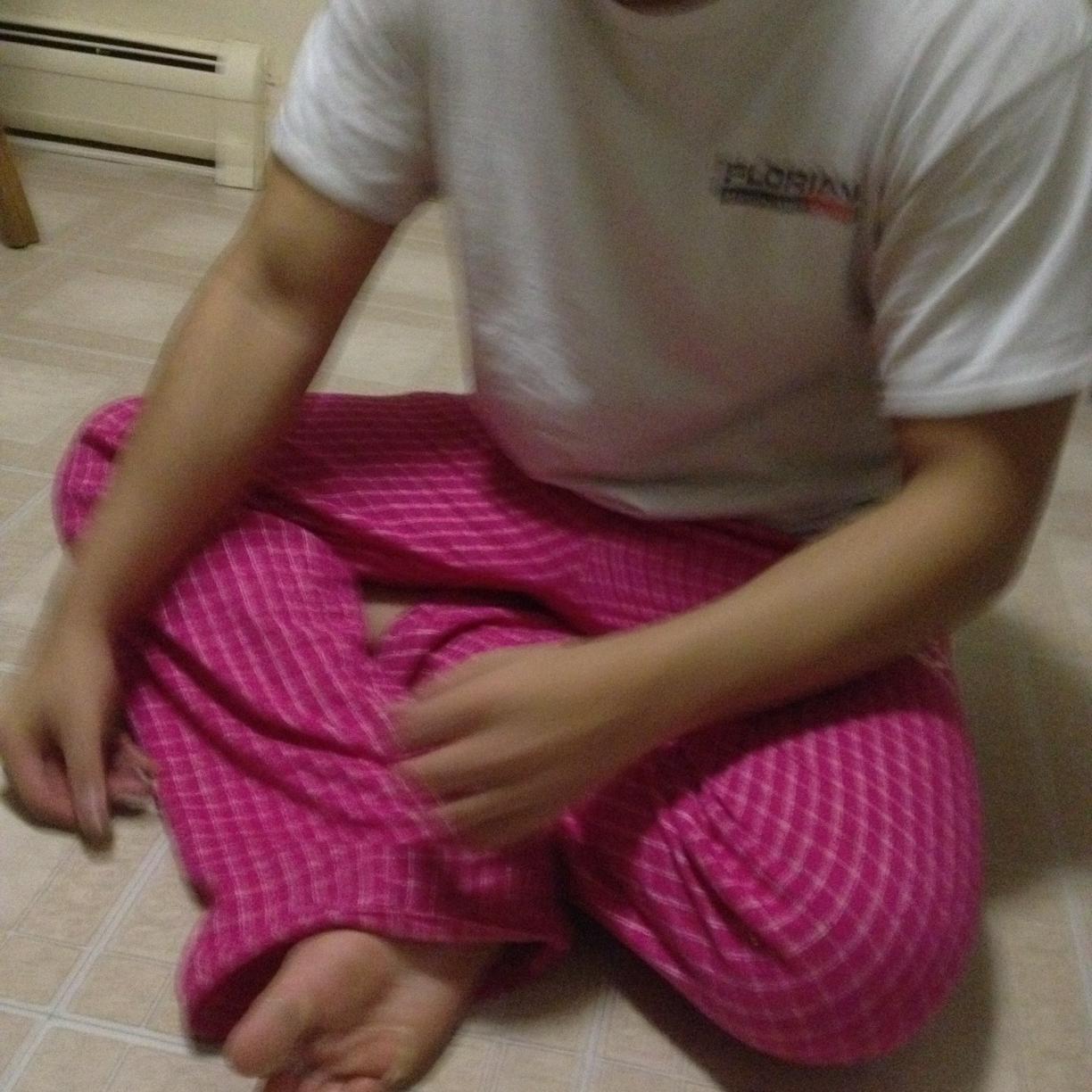}
        \end{minipage}
        \begin{minipage}{0.32\linewidth}
        \scriptsize{
        \textbf{SCST:} A person sitting on a tiled floor wearing pink pants sitting on the floor with his feet on the floor.\\
        \textbf{DiCO (Ours):} A man sitting on the floor with his legs crossed.
        }
    \end{minipage}
    \hspace{0.02cm}
    \begin{minipage}{0.165\linewidth}
        \includegraphics[width=0.97\linewidth]{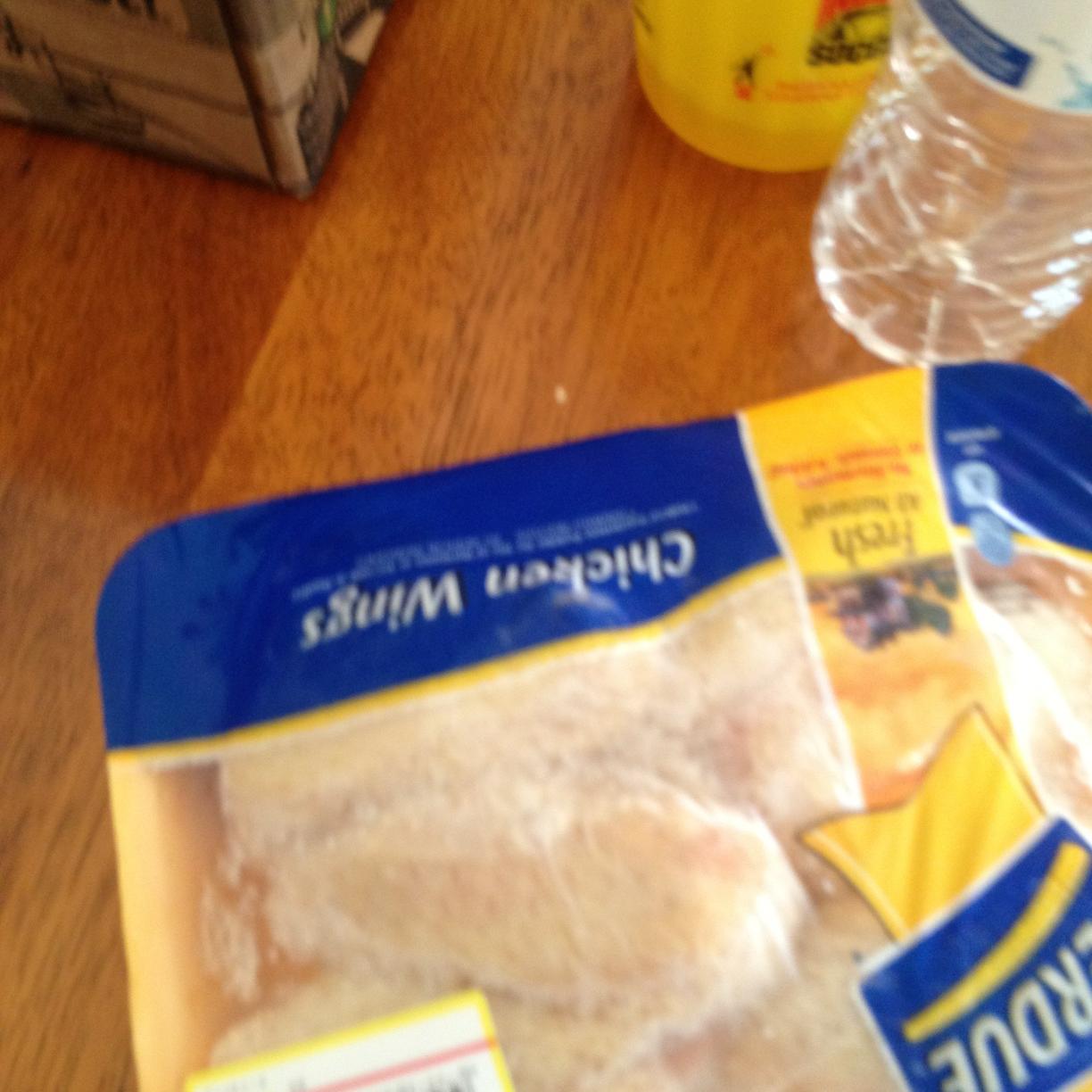}
        \end{minipage}
        \begin{minipage}{0.32\linewidth}
        \scriptsize{
        \textbf{SCST:} A plastic plastic container filled with meat on a wooden table next to a water bottle on a wood surface. \\
        \textbf{DiCO (Ours):} A plastic container of chicken on a wooden table next to a water bottle.
        }
    \end{minipage}
    
    \vspace{0.3cm}

        \begin{minipage}{0.165\linewidth}
        \includegraphics[width=0.97\linewidth]{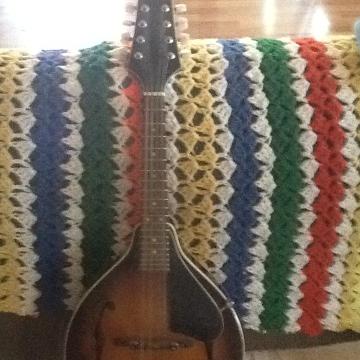}
        \end{minipage}
        \begin{minipage}{0.32\linewidth}
        \scriptsize{
        \textbf{SCST:} An electric guitar on carpet with yarn holder behind background bottom a hawk on a couch cushion behind background.\\
        \textbf{DiCO (Ours):} An old fashioned guitar sitting on a colorful rug.
        }
    \end{minipage}
    \hspace{0.02cm}
    \begin{minipage}{0.165\linewidth}
        \includegraphics[width=0.97\linewidth]{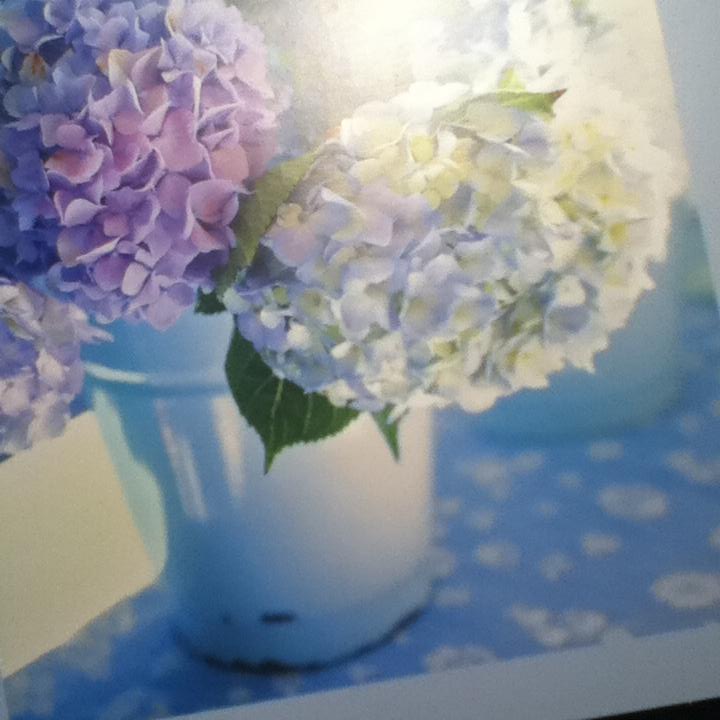}
        \end{minipage}
        \begin{minipage}{0.32\linewidth}
        \scriptsize{
        \textbf{SCST:} A blue vase filled with purple and white flowers on a blue table with other flowers on a blue background. \\
        \textbf{DiCO (Ours):} A picture of purple and white flowers on a blue table.
        }
    \end{minipage}
    \vspace{-0.15cm}
    \caption{Qualitative results on sample images from VizWiz.}
    \label{fig:vizwiz}
\end{figure}

\begin{figure}[t]
    \begin{minipage}{0.165\linewidth}
        \includegraphics[width=0.97\linewidth]{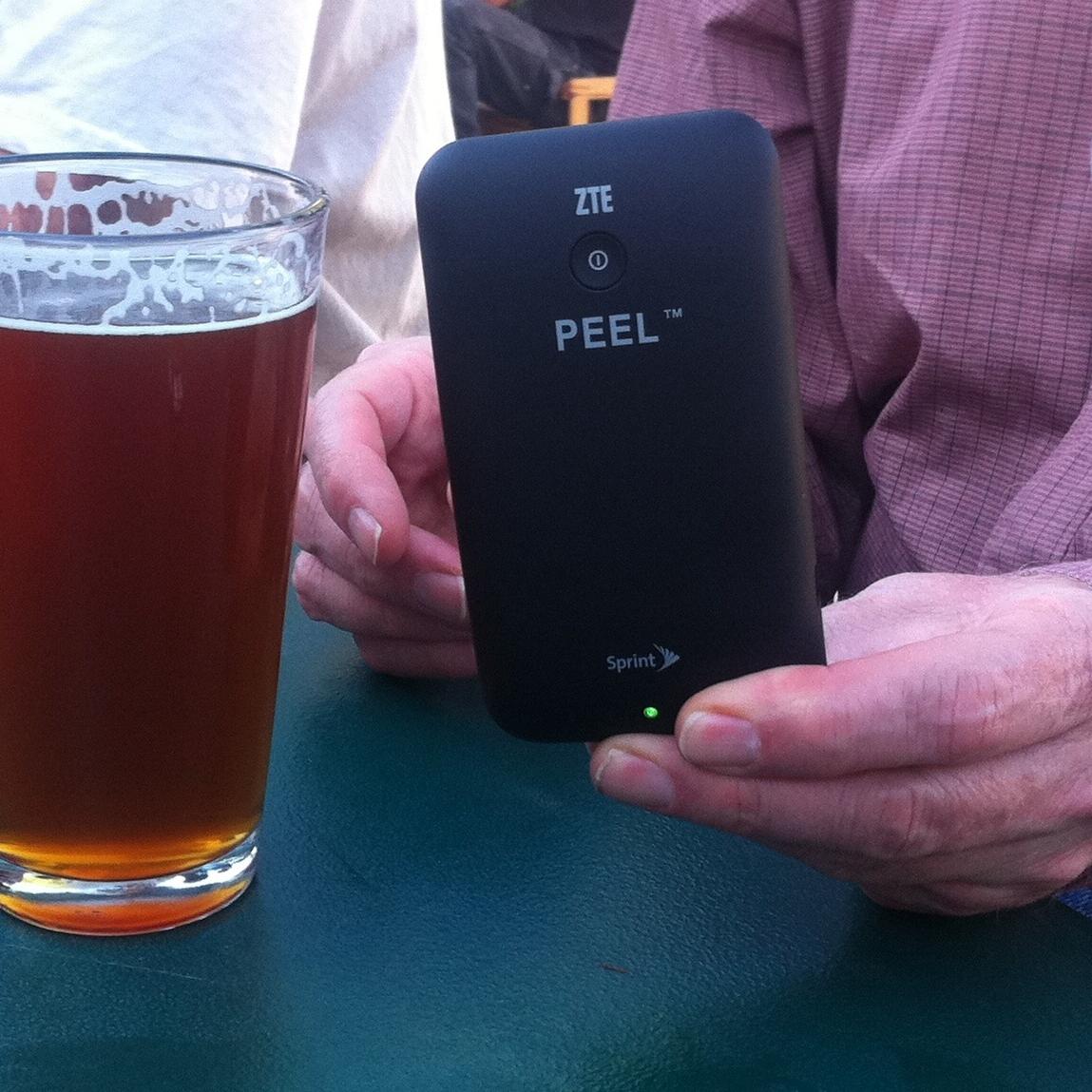}
        \end{minipage}
        \begin{minipage}{0.32\linewidth}
        \scriptsize{
        \textbf{SCST:} A person holding a cell phone with a beer in front of a hand holding a cell phone with a beer in it. \\
        \textbf{DiCO (Ours):} A person holding a smart phone next to a beer.
        }
    \end{minipage}
    \hspace{0.02cm}
    \begin{minipage}{0.165\linewidth}
        \includegraphics[width=0.97\linewidth]{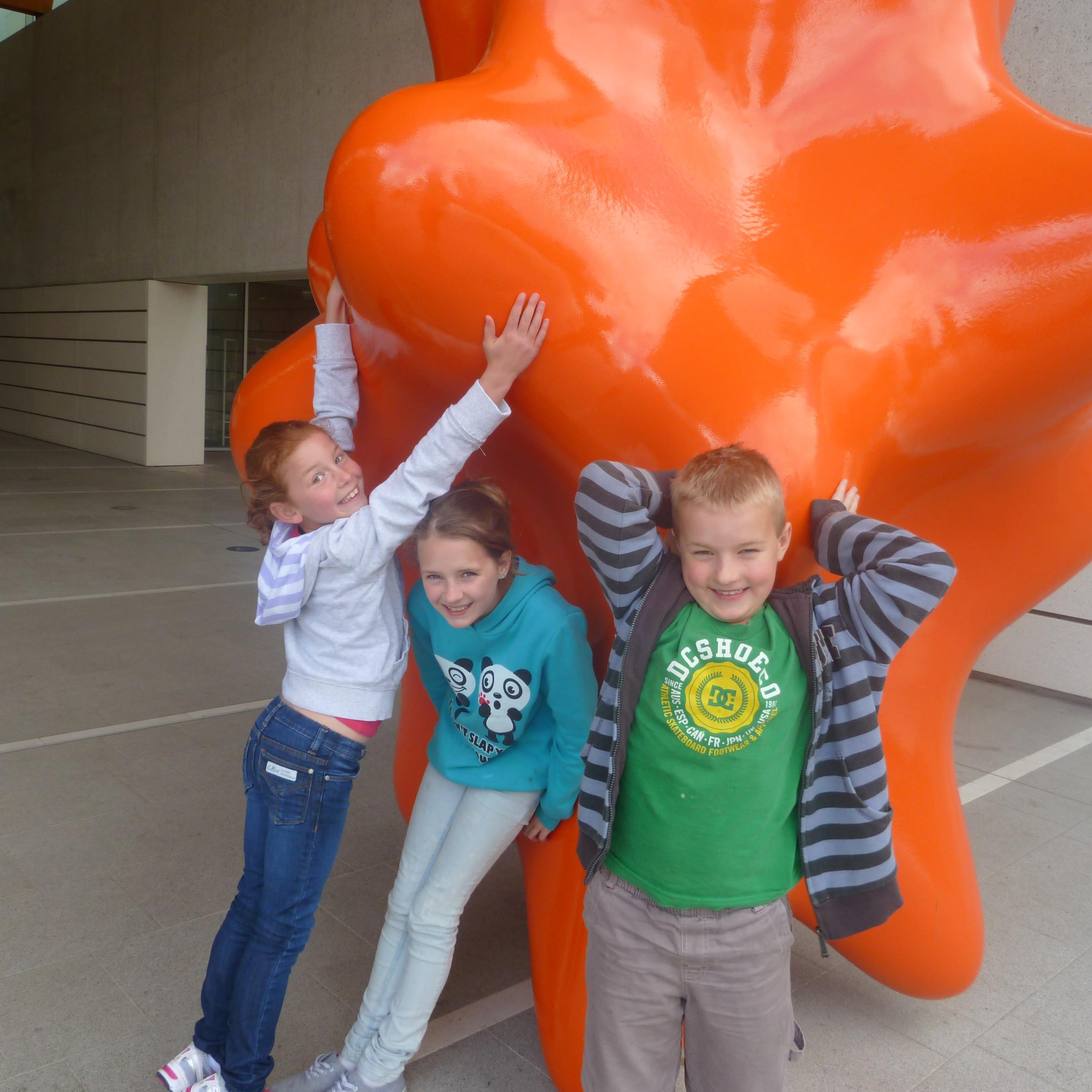}
        \end{minipage}
        \begin{minipage}{0.32\linewidth}
        \scriptsize{
        \textbf{SCST:} Three young children standing around an orange orange statue with three young girls in an orange building.\\
        \textbf{DiCO (Ours):} Three young children reaching up on a orange statue.
        }
    \end{minipage}
    
    \vspace{0.3cm}

    \begin{minipage}{0.165\linewidth}
        \includegraphics[width=0.97\linewidth]{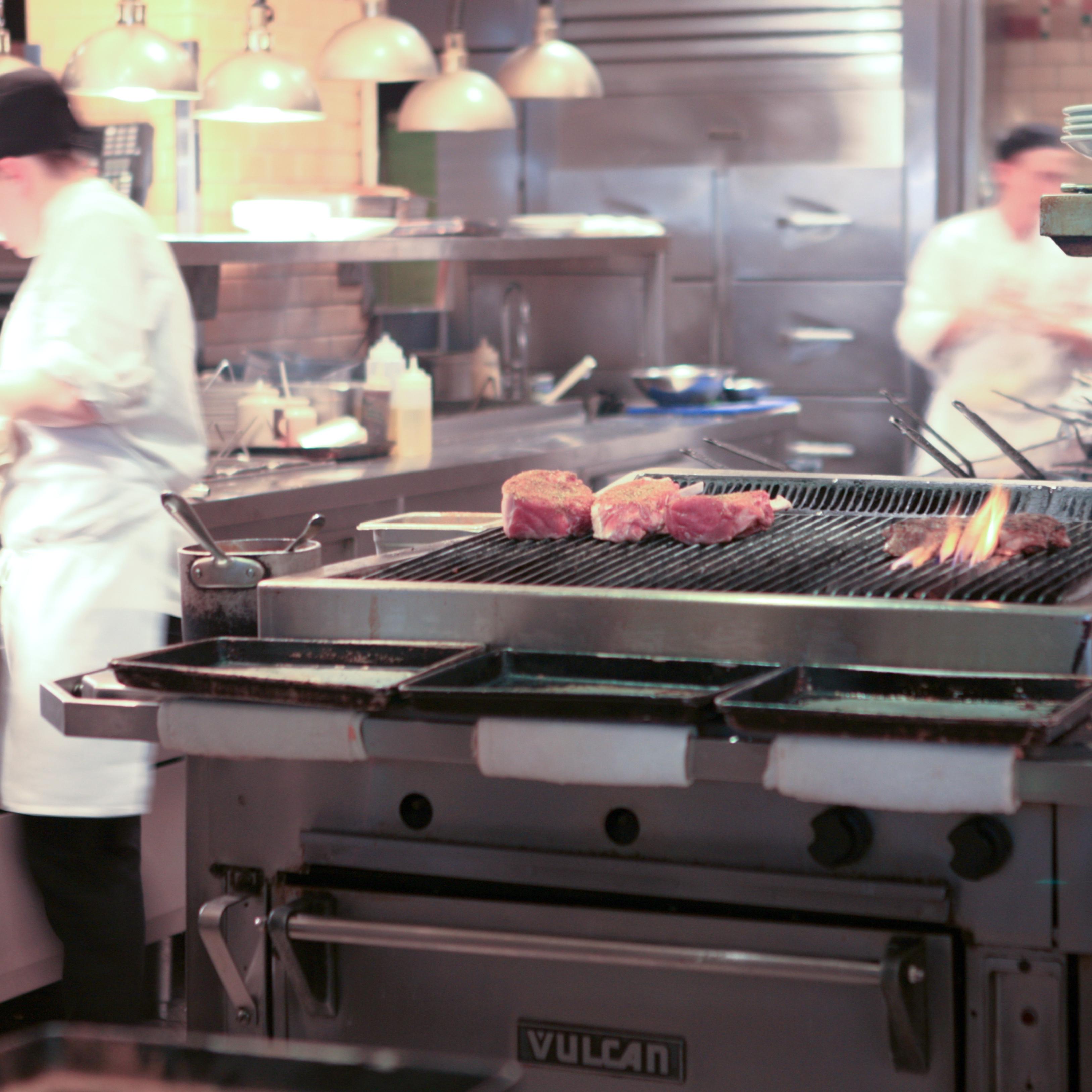}
        \end{minipage}
        \begin{minipage}{0.32\linewidth}
        \scriptsize{
        \textbf{SCST:} Two chefs working in an industrial industrial kitchen with a grill in a stainless steel oven with lots of meat on the counter. \\
        \textbf{DiCO (Ours):} Two chefs working in a commercial kitchen with a metal grill.
        }
    \end{minipage}
    \hspace{0.02cm}
    \begin{minipage}{0.165\linewidth}
        \includegraphics[width=0.97\linewidth]{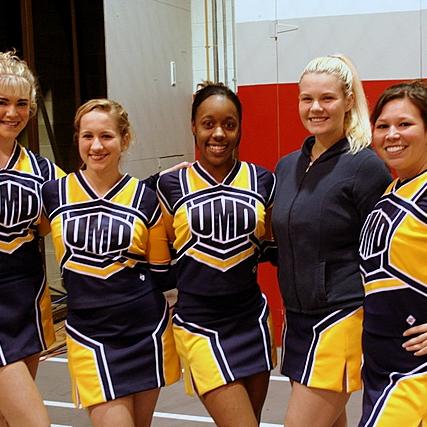}
        \end{minipage}
        \begin{minipage}{0.32\linewidth}
        \scriptsize{
        \textbf{SCST:} A group of young women dressed in yellow school uniforms posing for a picture in a school uniform. \\
        \textbf{DiCO (Ours):} A group of young women wearing yellow and blue school uniforms posing for a picture together.
        }
    \end{minipage}
    
    \vspace{0.3cm}

        \begin{minipage}{0.165\linewidth}
        \includegraphics[width=0.97\linewidth]{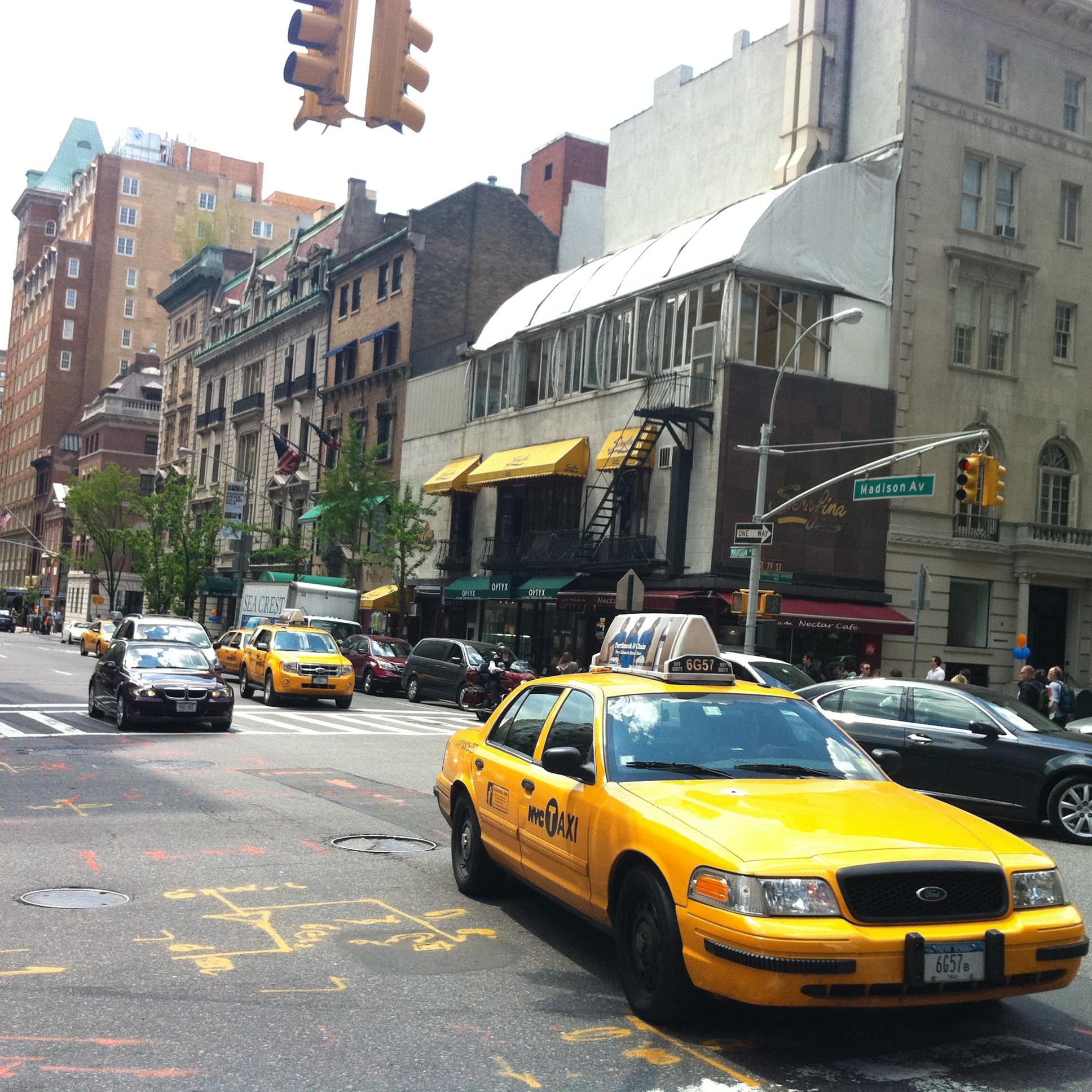}
        \end{minipage}
        \begin{minipage}{0.32\linewidth}
        \scriptsize{
        \textbf{SCST:} A yellow taxi cab driving down a busy city street with cars on a busy city street with buildings in the background.\\
        \textbf{DiCO (Ours):} A yellow taxi cab driving down a busy city street with other cars and buildings.
        }
    \end{minipage}
    \hspace{0.02cm}
    \begin{minipage}{0.165\linewidth}
        \includegraphics[width=0.97\linewidth]{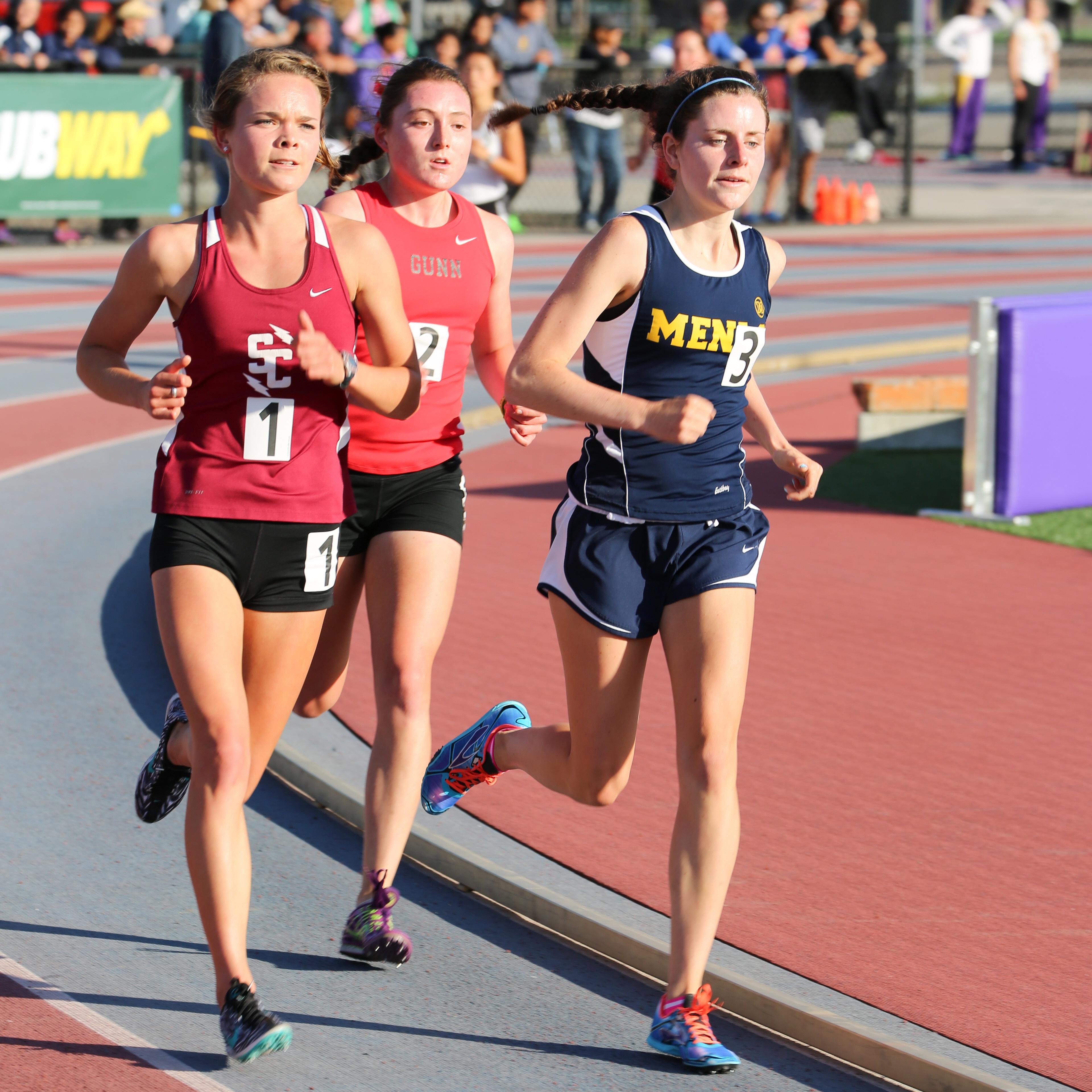}
        \end{minipage}
        \begin{minipage}{0.32\linewidth}
        \scriptsize{
        \textbf{SCST:} A group of young women running on a race race track with two girls running around a race track with a crowd in the background.\\
        \textbf{DiCO (Ours):} A group of young women running across a track at a competition.
        }
    \end{minipage}
    \vspace{-0.15cm}
    \caption{Qualitative results on sample images from TextCaps.}
    \label{fig:textcaps}
\end{figure}

\begin{figure}[t]
    \begin{minipage}{0.165\linewidth}
        \includegraphics[width=0.97\linewidth]{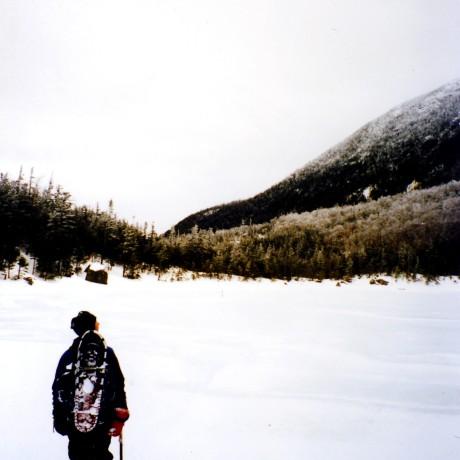}
        \end{minipage}
        \begin{minipage}{0.32\linewidth}
        \scriptsize{
        \textbf{SCST:} A person with a backpack walking through deep snow with a backpack walking through a forest with mountains in the background. \\
        \textbf{DiCO (Ours):} A person standing in the snow with a backpack near a large hill with a mountain in the background.
        }
    \end{minipage}
    \hspace{0.02cm}
    \begin{minipage}{0.165\linewidth}
        \includegraphics[width=0.97\linewidth]{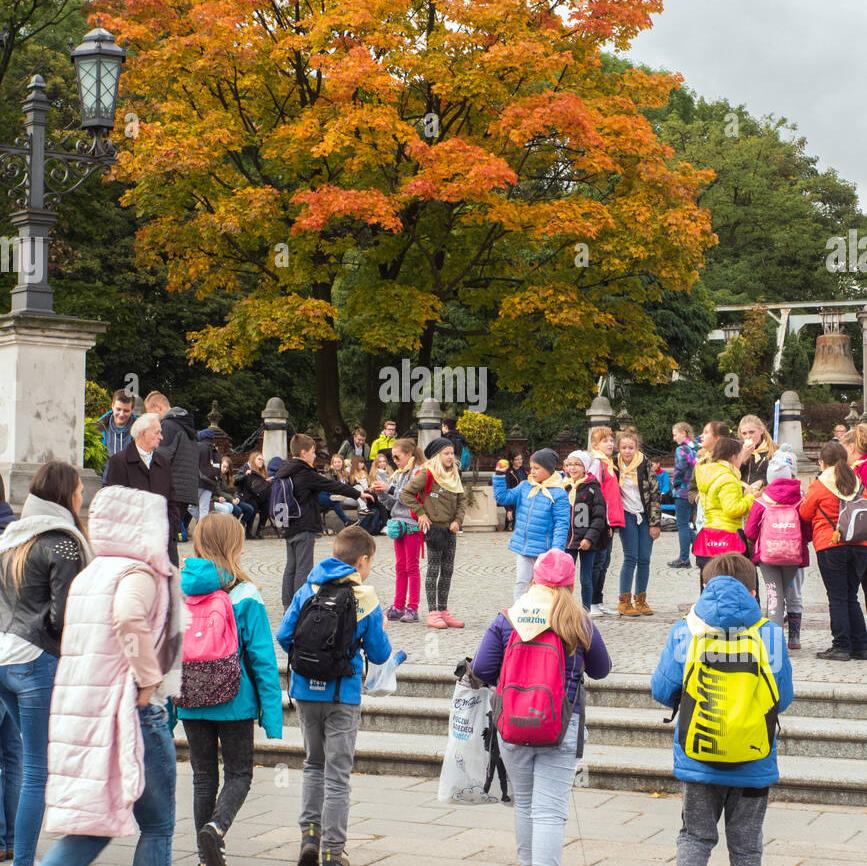}
        \end{minipage}
        \begin{minipage}{0.32\linewidth}
        \scriptsize{
        \textbf{SCST:} A group of people walking down a sidewalk with lots of people with backpacks walking around a path with trees in the background. \\
        \textbf{DiCO (Ours):} A crowd of people standing on a sidewalk next to a tree covered with colorful leaves.
        }
    \end{minipage}
    
    \vspace{0.3cm}

    \begin{minipage}{0.165\linewidth}
        \includegraphics[width=0.97\linewidth]{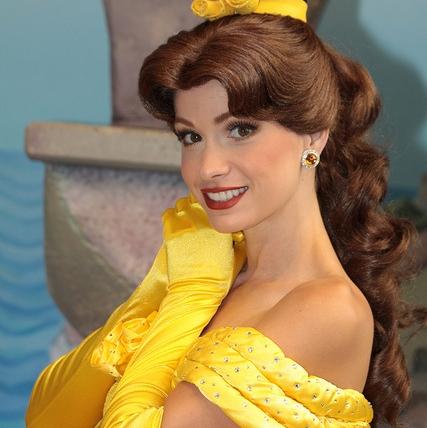}
        \end{minipage}
        \begin{minipage}{0.32\linewidth}
        \scriptsize{
        \textbf{SCST:} A beautiful young woman wearing a yellow costume posing for a picture wearing a yellow dress with her hand on her face.\\
        \textbf{DiCO (Ours):} A beautiful young woman wearing a yellow dress posing with her arm around her neck.
        }
    \end{minipage}
    \hspace{0.02cm}
    \begin{minipage}{0.165\linewidth}
        \includegraphics[width=0.97\linewidth]{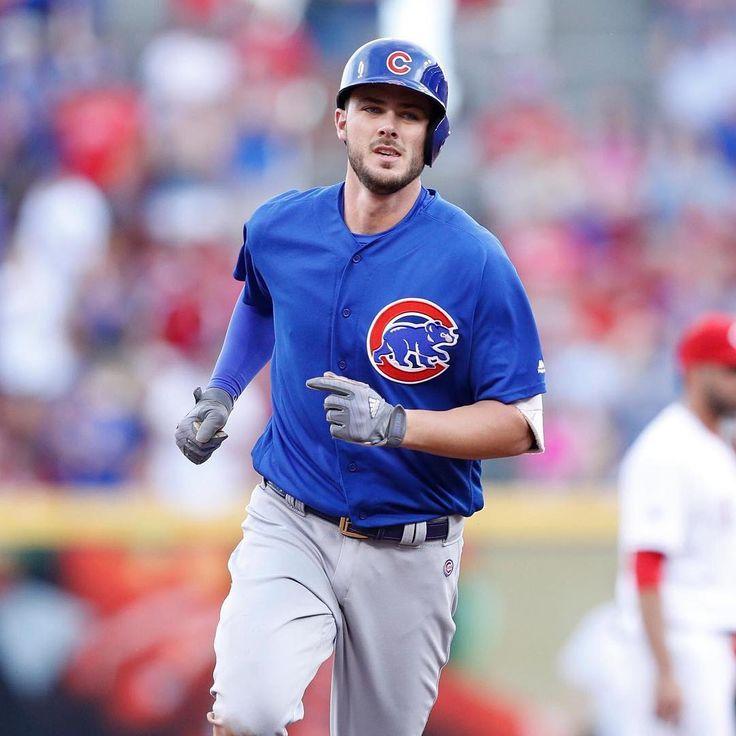}
        \end{minipage}
        \begin{minipage}{0.32\linewidth}
        \scriptsize{
        \textbf{SCST:} A professional baseball player running with a blue helmet in a blue uniform in a stadium with a crowd crowd in the background.\\
        \textbf{DiCO (Ours):} A professional baseball player in blue and white uniform running through a stadium.
        }
    \end{minipage}
    
    \vspace{0.3cm}

    \begin{minipage}{0.165\linewidth}
        \includegraphics[width=0.97\linewidth]{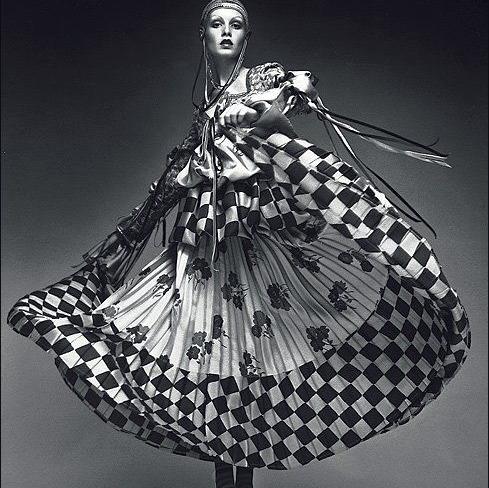}
        \end{minipage}
        \begin{minipage}{0.32\linewidth}
        \scriptsize{
        \textbf{SCST:} Black and white photograph of a woman dressed dressed in black and white dress with long black and white clothing on a dark surface. \\
        \textbf{DiCO (Ours):} A black and white photo of a woman dressed in black and white clothing.
        }
    \end{minipage}
    \hspace{0.02cm}
    \begin{minipage}{0.165\linewidth}
        \includegraphics[width=0.97\linewidth]{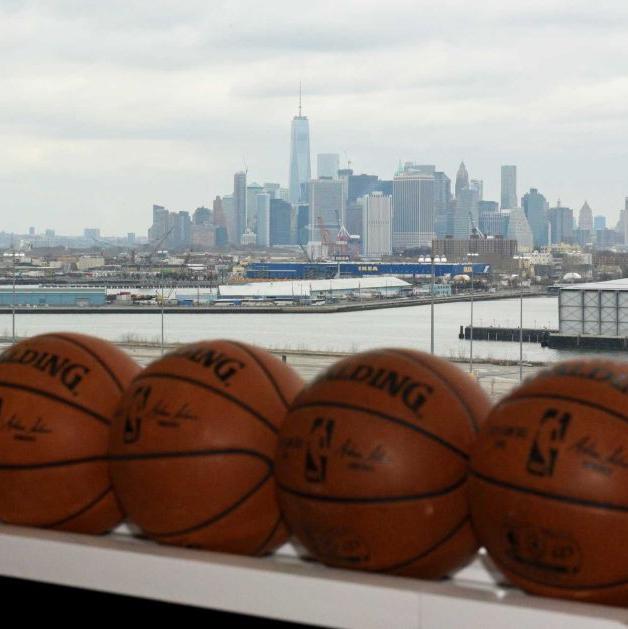}
        \end{minipage}
        \begin{minipage}{0.32\linewidth}
        \scriptsize{
        \textbf{SCST:} A row of basketball balls sitting next to a row of orange balls in front of a body of water with city in background.\\
        \textbf{DiCO (Ours):} A row of basketball balls in front of a view of a city skyline in the background.
        }
    \end{minipage}
    \vspace{-0.15cm}
    \caption{Qualitative results on sample images from CC3M.}
    \label{fig:cc3m}
\end{figure}

Finally, we report additional qualitative results to qualitatively validate the effectiveness of our training strategy. In particular, Fig.~\ref{fig:coco_qualitatives_supp1} and Fig.~\ref{fig:coco_qualitatives_supp2} show sample images from the COCO dataset and captions predicted by \ours in comparison to those generated by SCST, the model proposed in~\cite{cho2022fine}, and the large-scale model BLIP-2~\cite{li2023blip}. As it can be seen, \ours generates significantly more detailed captions than BLIP-2, while reducing repetitions typically present in SCST-generated sentences.
To qualitative validate the generalization capabilities to out-of-domain images, we report sample captions predicted by \ours and SCST using PAC-S as reward on nocaps~\cite{agrawal2019nocaps} (Fig.~\ref{fig:nocaps}), VizWiz~\cite{gurari2020captioning} (Fig.~\ref{fig:vizwiz}), TextCaps~\cite{sidorov2020textcaps} (Fig.~\ref{fig:textcaps}), and CC3M~\cite{sharma2018conceptual} (Fig.~\ref{fig:cc3m}).

\begin{figure}[t]
\centering
    \begin{minipage}{0.165\linewidth}
        \includegraphics[width=0.97\linewidth]{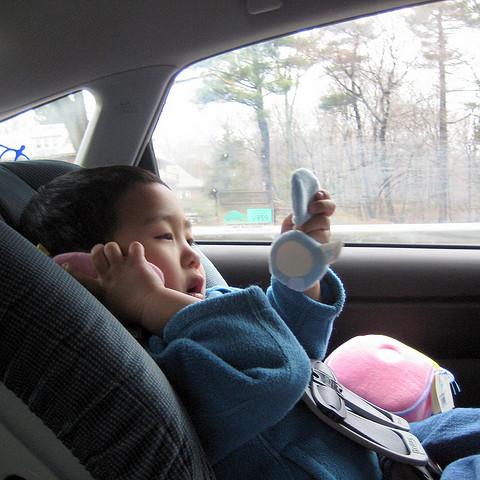}
    \end{minipage}
\begin{minipage}{0.31\linewidth}
    \scriptsize{
    \textbf{$\mathcal{M}^2$ Transf.~\cite{cornia2020meshed}:} A young boy sitting in the passenger seat of a car.\\
    \textbf{COS-Net~\cite{li2022comprehending}:} A little girl sitting in the back seat of a car holding a cell phone.\\
    \textbf{Transf. (SCST):} A child sitting in a car holding a cell phone.\\
    \textbf{Transf. (\ours):} A child sitting in the back seat of a car talking on a cell phone.
    }
\end{minipage}
\hspace{0.02cm}
    \begin{minipage}{0.165\linewidth}
        \includegraphics[width=0.97\linewidth]{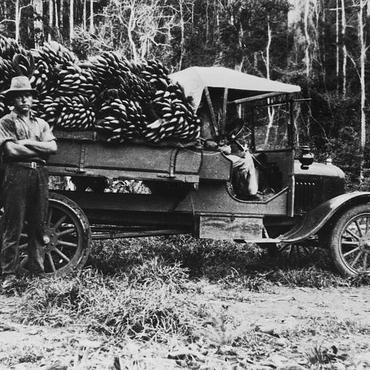}
    \end{minipage}
\begin{minipage}{0.31\linewidth}
    \scriptsize{
    \textbf{$\mathcal{M}^2$ Transf.~\cite{cornia2020meshed}:} A man sitting on the back of a truck with bananas.\\
    \textbf{COS-Net~\cite{li2022comprehending}:} An old truck with a man standing on the back of it.\\
    \textbf{Transf. (SCST):} A man standing next to a truck full of bananas.\\
    \textbf{Transf. (\ours):} A black and white photo of a man with a truck full of bananas.
    }
\end{minipage}   
\vspace{0.3cm}

\begin{minipage}{0.165\linewidth}
        \includegraphics[width=0.97\linewidth]{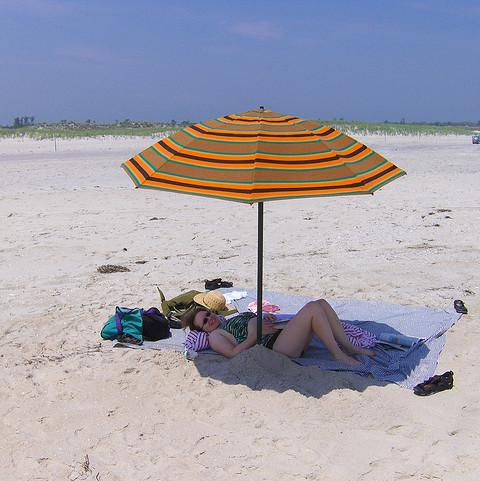}
    \end{minipage}
\begin{minipage}{0.31\linewidth}
    \scriptsize{
    \textbf{$\mathcal{M}^2$ Transf.~\cite{cornia2020meshed}:} A woman laying on the beach under an umbrella on a.\\
    \textbf{COS-Net~\cite{li2022comprehending}:} A woman laying on a beach under an umbrella.\\
    \textbf{Transf. (SCST):} A woman laying on a blanket under an umbrella on a.\\
    \textbf{Transf. (\ours):} A woman laying on a blanket under an umbrella on the beach.
    }
\end{minipage}
\hspace{0.02cm}
\begin{minipage}{0.165\linewidth}
        \includegraphics[width=0.97\linewidth]{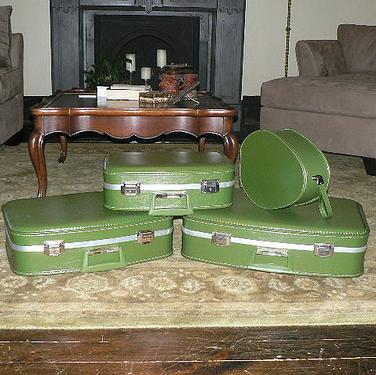}
    \end{minipage}
\begin{minipage}{0.31\linewidth}
    \scriptsize{
    \textbf{$\mathcal{M}^2$ Transf.~\cite{cornia2020meshed}:} A bunch of green suitcases stacked on top of a fireplace.\\
    \textbf{COS-Net~\cite{li2022comprehending}:} A group of green luggage sitting on a couch in a living room.\\
    \textbf{Transf. (SCST):} Three green suitcases stacked on top of each other.\\
    \textbf{Transf. (\ours):} Three green suitcases sitting on the floor in a living room.
    }
\end{minipage}      
\vspace{0.3cm}

\begin{minipage}{0.165\linewidth}
        \includegraphics[width=0.97\linewidth]{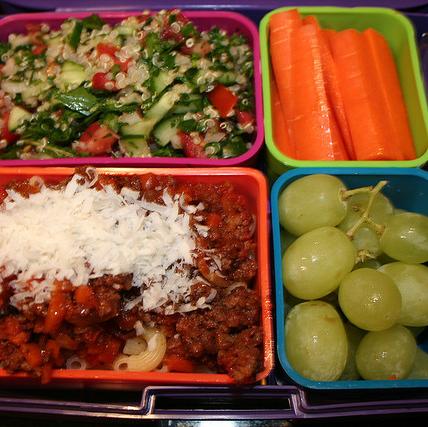}
    \end{minipage}
\begin{minipage}{0.31\linewidth}
    \scriptsize{
    \textbf{$\mathcal{M}^2$ Transf.~\cite{cornia2020meshed}:} A tray of food with rice and vegetables in a.\\
    \textbf{COS-Net~\cite{li2022comprehending}:} A group of plastic containers filled with food.\\
    \textbf{Transf. (SCST):} Four containers of food with carrots and a.\\
    \textbf{Transf. (\ours):} Four plastic containers filled with different types of food.
    }
\end{minipage}
\hspace{0.02cm}
\begin{minipage}{0.165\linewidth}
        \includegraphics[width=0.97\linewidth]{images/supplementary/fig_9/cropped_134863.jpeg}
    \end{minipage}
\begin{minipage}{0.31\linewidth}
    \scriptsize{
    \textbf{$\mathcal{M}^2$ Transf.~\cite{cornia2020meshed}:} An empty street at sunset with a green traffic light.\\
    \textbf{COS-Net~\cite{li2022comprehending}:} An intersection with traffic lights and a city street.\\
    \textbf{Transf. (SCST):} A city street with traffic lights and a.\\
    \textbf{Transf. (\ours):} A street with traffic lights and a building in the background.
    }
\end{minipage} 
\vspace{0.2cm}
\caption{Qualitative results on sample images from the COCO Karpathy test split using \ours with CIDEr reward. We compare our approach with a standard Transformer trained with SCST and CIDEr as reward, $\mathcal{M}^2$ Transformer~\cite{cornia2020meshed} and COS-Net~\cite{li2022comprehending}.}
\label{fig:coco_qualitatives_supp_cider}
\end{figure}

In Fig.~\ref{fig:coco_qualitatives_supp_cider}, we instead show some qualitative results when using CIDEr as reward. In this case, we compare \ours with standard image captioning models, including a vanilla Transformers trained with the same visual features used in our approach, COS-Net~\cite{li2022comprehending}, and $\mathcal{M}^2$ Transformer~\cite{cornia2020meshed}. All competitors have been trained with a standard XE+SCST training protocol. Also in this setting, \ours is able to generate high-quality captions compared to competitors, confirming that it can also be employed as a valid alternative to SCST for training standard image captioning models.

\section{Limitations}
\label{sec:limitations}
\begin{figure}[t]
    \begin{minipage}{0.165\linewidth}
        \includegraphics[width=0.97\linewidth]{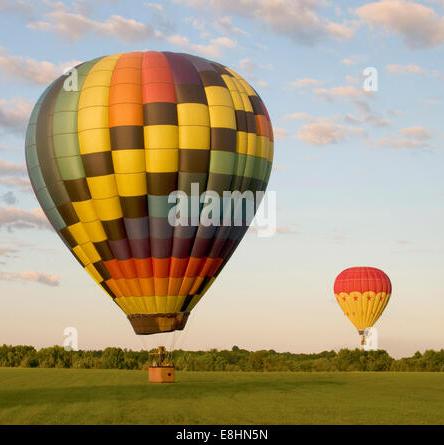}
        \end{minipage}
        \begin{minipage}{0.32\linewidth}
        \scriptsize{
        \textbf{SCST:} Two colorful colorful kites flying over a grassy field with a yellow yellow red yellow and yellow flags on clear day. \\
        \textbf{DiCO (Ours):} Two colorful kites flying through a field on a clear day.
        }
    \end{minipage}
    \hspace{0.02cm}
    \begin{minipage}{0.165\linewidth}
        \includegraphics[width=0.97\linewidth]{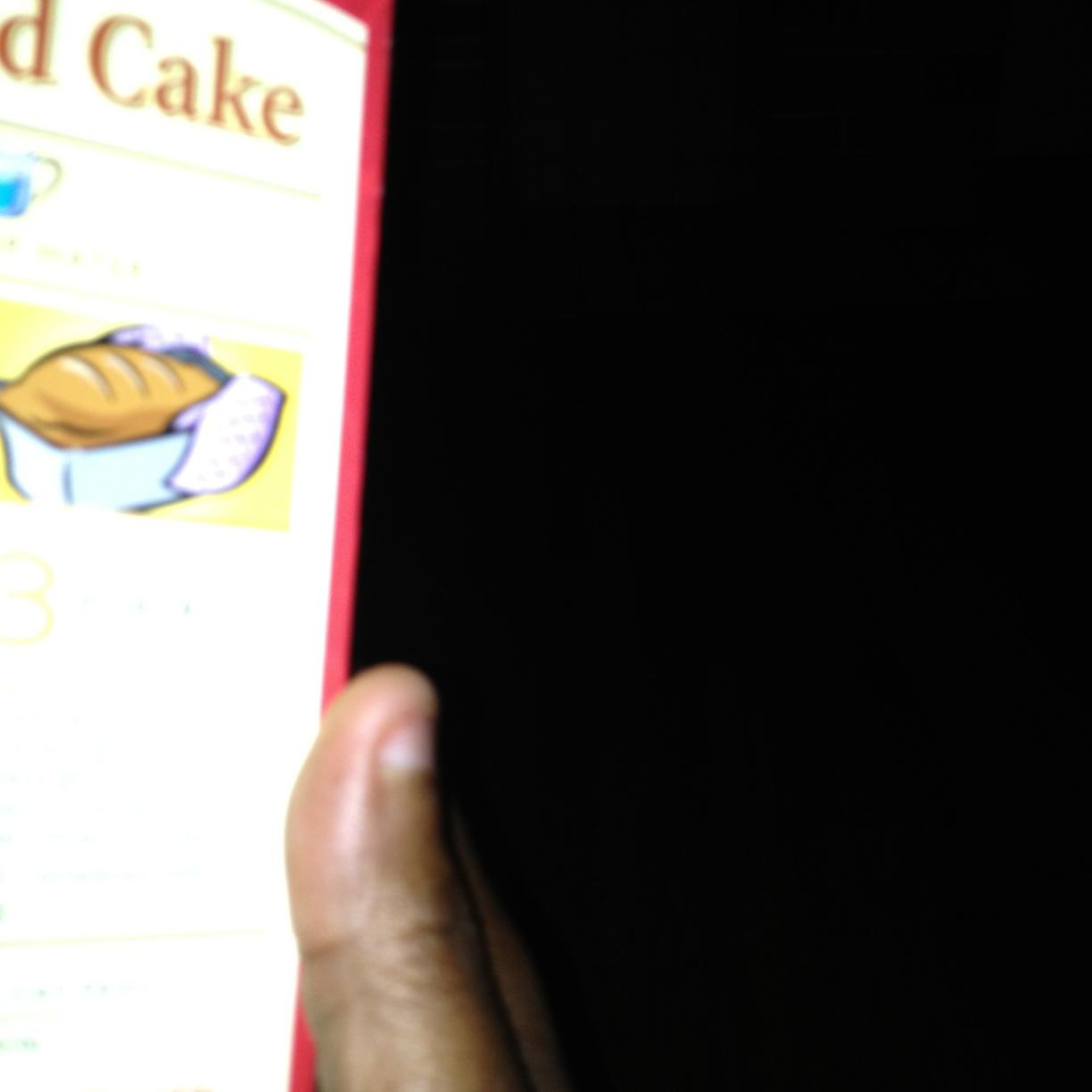}
        \end{minipage}
        \begin{minipage}{0.32\linewidth}
        \scriptsize{
        \textbf{SCST:} A hand holding up a piece of cake with a picture of a person's hand holding out a paper on a dark surface.\\
        \textbf{DiCO (Ours):} A person's hand holding up a piece of cake with a picture of a sign on it.
        }
    \end{minipage}
    
    \vspace{0.3cm}

    \begin{minipage}{0.165\linewidth}
        \includegraphics[width=0.97\linewidth]{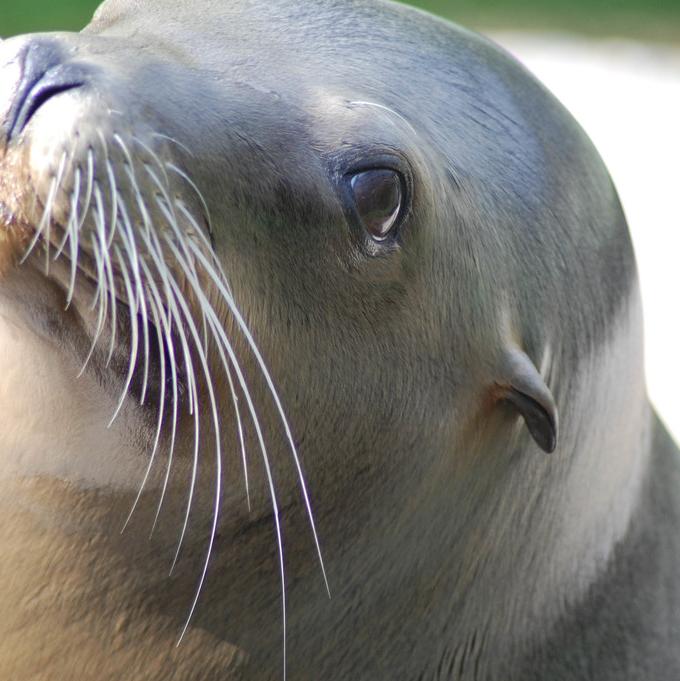}
        \end{minipage}
        \begin{minipage}{0.32\linewidth}
        \scriptsize{
        \textbf{SCST:} A close up picture of a penguin with its eyes open with its eyes open. \\
        \textbf{DiCO (Ours):} A close up picture of a very small brown stuffed animal.
        }
    \end{minipage}
    \hspace{0.02cm}
    \begin{minipage}{0.165\linewidth}
        \includegraphics[width=0.97\linewidth]{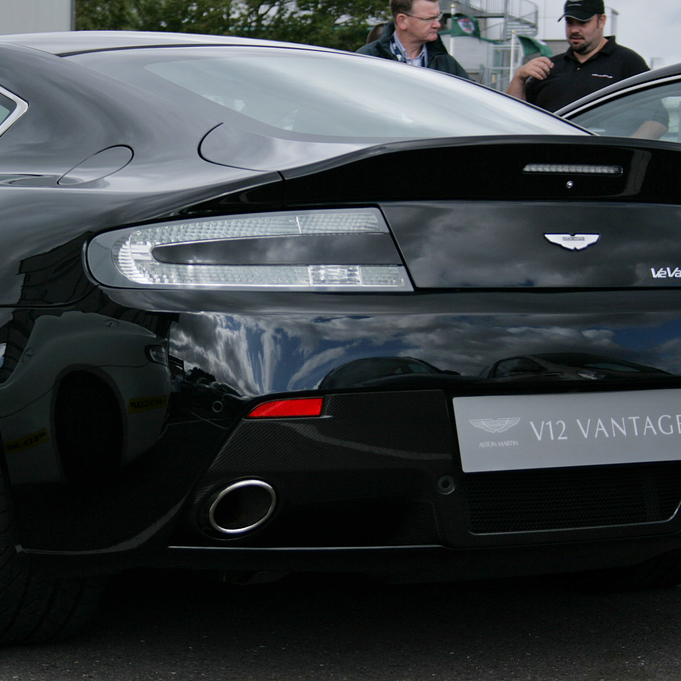}
        \end{minipage}
        \begin{minipage}{0.32\linewidth}
        \scriptsize{
        \textbf{SCST:} A black car with a black leather seat parked in a parking lot with people looking at something in the background. \\
        \textbf{DiCO (Ours):} A black silver car parked in a parking lot.
        }
    \end{minipage}    
    \vspace{0.3cm}

    \begin{minipage}{0.165\linewidth}
        \includegraphics[width=0.97\linewidth]{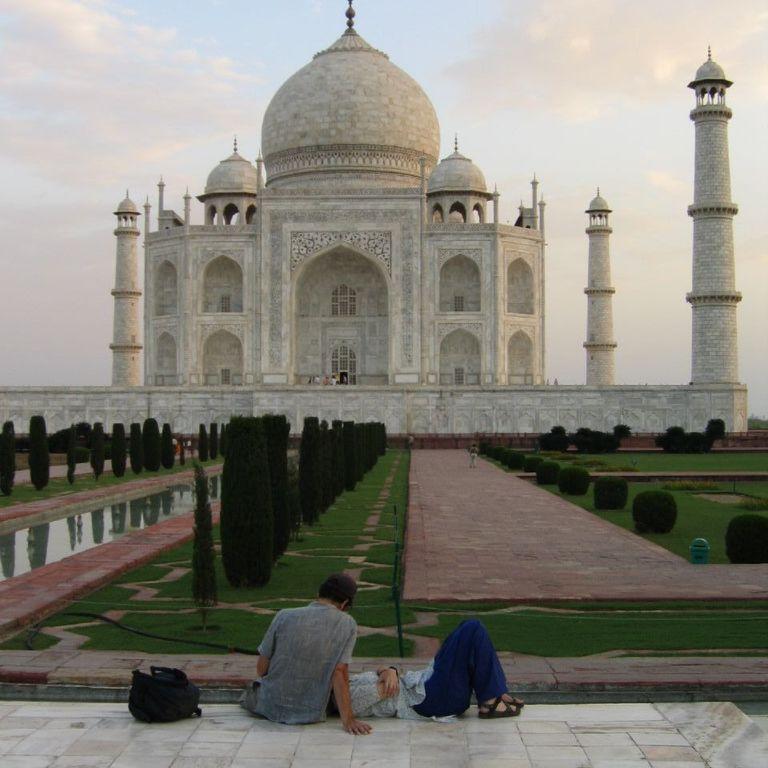}
        \end{minipage}
        \begin{minipage}{0.32\linewidth}
        \scriptsize{
        \textbf{SCST:} Two people sitting in front of a white monument with a monument in front of a white monument. \\
        \textbf{DiCO (Ours):} Two people sitting on a bench in front of a large white monument.
        }
    \end{minipage}
    \hspace{0.02cm}
    \begin{minipage}{0.165\linewidth}
        \includegraphics[width=0.97\linewidth]{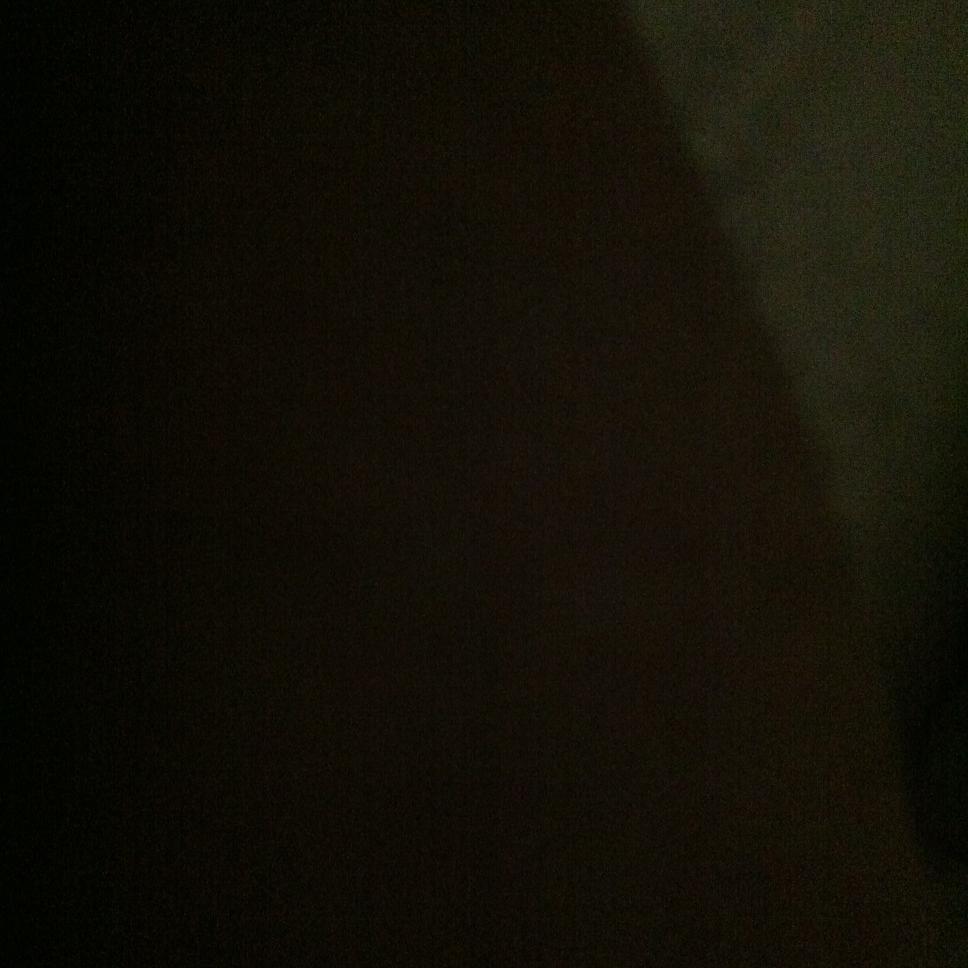}
        \end{minipage}
        \begin{minipage}{0.32\linewidth}
        \scriptsize{
        \textbf{SCST:} A close up view of a person's leg in a dark dark dark dark dark dark colored shadow against a white background.\\
        \textbf{DiCO (Ours):} The shadow of a person's feet in a dark room.
        }
    \end{minipage}
    \vspace{-0.15cm}
    \caption{Qualitative results showcasing samples where DiCO fails in comparison to the SCST training methodology.}
    \label{fig:limitations}
\end{figure}

As with all image captioning models, we acknowledge that our method might fail to provide informative captions in some rare contexts. To qualitatively evaluate the limitations of our approach, we report some failure cases in Fig.~\ref{fig:limitations}. As it can be seen, \ours may produce factual errors, \eg~mistaking balloons for \textit{kites} (first sample, first row) or a \textit{stuffed animal} for a seal (first sample, second row). Additionally, \ours may fail to recognize known entities, thus providing only a broad description of the scene (\eg~a \textit{white monument} rather than the Taj Mahal mausoleum, or a \textit{black silver car} rather than an Aston Martin). This can be conducted to the image-caption pairs contained in the COCO dataset, which lack open-world knowledge. Finally, when the main subject of the image is uncertain (second sample, third row), \ours may overlook the picture and generate captions based on its learned priors, resulting in hallucinations.

\end{document}